\documentclass[10pt,journal,compsoc]{IEEEtran}
\pdfoutput=1
\usepackage{afterpage}
\ifCLASSOPTIONcompsoc
  \usepackage[nocompress]{cite}
\else
  \usepackage{cite}
\fi

\usepackage[utf8]{inputenc} 
\usepackage[T1]{fontenc}    
\usepackage{hyperref}       
\usepackage{url}            
\usepackage{booktabs}       
\usepackage{amsfonts}       
\usepackage{nicefrac}       
\usepackage{microtype}      
\usepackage{algorithm}
\usepackage{eqparbox}
\usepackage{algorithmic}
\usepackage{amsmath}
\usepackage{multirow}
\usepackage{graphicx}
\usepackage{wrapfig}
\usepackage{threeparttable}
\usepackage{subfigure}
\usepackage{amsmath}
\usepackage{verbatim}
\usepackage{xcolor}
\usepackage{enumitem}

\newtheorem{definition}{Definition}

\begin{document}
\bstctlcite{IEEEexample:BSTcontrol}
\title{AlphaGAN: Fully Differentiable Architecture Search for Generative Adversarial Networks}

%

\author{Yuesong Tian$^{[0000-0001-6287-4174]}$ \quad
\IEEEcompsocitemizethanks{\IEEEcompsocthanksitem Yuesong Tian is with the College of Biomedical Engineering and Instrument Science, Zhejiang University, Hangzhou, China. \protect\\
Email: \texttt{tianys163@gmail.com}
\IEEEcompsocthanksitem Li Shen is with JD Explore Academy, Beijing, China. Li Shen is the corresponding author. 
E-mail: \texttt{mathshenli@gmail.com}
\IEEEcompsocthanksitem Li Shen is with Tencent AI Lab, shenzhen, China. Email: \texttt{lshen.lsh@gmail.com}
\IEEEcompsocthanksitem Zhifeng Li, and Wei Liu are with Tencent Data Platform, Shenzhen, China.\protect\\
E-mail: \texttt{michaelzfli@tencent.com}, \quad \texttt{wl2223@columbia.edu}
\IEEEcompsocthanksitem Guinan Su is with Microsoft, Beijing, China. \protect\\
Email: \texttt{guinansu33@gmail.com}

}
Li Shen$^{[0000-0001-5659-3464]}$ \quad 
Li Shen$^{[0000-0002-2283-4976]}$ \quad \\
Guinan Su$^{[0000-0003-0504-6676]}$ \quad
Zhifeng Li$^{[0000-0001-5902-5067]}$ \quad
Wei Liu$^{[0000-0002-3865-8145]}$
}
\IEEEtitleabstractindextext{%
\begin{abstract}\label{abstract-sec}
Generative Adversarial Networks (GANs) are formulated as minimax game problems that generative networks attempt to approach real data distributions by adversarial learning against discriminators which learn to distinguish generated samples from real ones, of which the intrinsic problem complexity poses challenges to performance and robustness. In this work, we aim to boost model learning from the perspective of network architectures, by incorporating recent progress on automated architecture search into GANs. Specially we propose a fully differentiable search framework, dubbed {\em alphaGAN}, where the searching process is formalized as solving a bi-level minimax optimization problem. The outer-level objective aims for seeking an optimal network architecture towards pure Nash Equilibrium conditioned on the network parameters of generators and discriminators  optimized with a traditional adversarial loss within inner level. The entire optimization performs a first-order approach by alternately minimizing the two-level objective in a fully differentiable manner that enables obtaining a suitable architecture efficiently from an enormous search space. Extensive experiments on CIFAR-10 and STL-10 datasets show that our algorithm can obtain high-performing architectures only with $3$-GPU hours on a single GPU in the search space comprised of approximate $2\times 10^{11}$ possible configurations. 
We further validate the method on the state-of-the-art network StyleGAN2, and push the score of Fréchet Inception Distance (FID) further, i.e., achieving 1.94 on CelebA, 2.86 on LSUN-church and 2.75 on FFHQ, with relative improvements $3\%\sim26\%$ over the baseline architecture. We also provide a comprehensive analysis of the behavior of the searching process and the properties of searched architectures, which would benefit further research on architectures for generative models. Codes and models are available at \url{https://github.com/yuesongtian/AlphaGAN}.

\end{abstract}

}

\maketitle

\section{Introduction}\label{introduction-sec}
Generative Adversarial Networks (GANs) \cite{goodfellow2014generative} have shown promising performance on a variety of generative tasks (e.g., image generation \cite{brock2018large, donahue2019large, karras2017progressive, karras2019style, karras2020analyzing}, image translation \cite{isola2017image, zhu2017unpaired, wang2018high, choi2018stargan, choi2020stargan, park2020contrastive}, dialogue generation \cite{li2017adversarial}, and image inpainting \cite{yu2018generative}), which are typically formulated as adversarial learning between a pair of networks, called generator and discriminator \cite{oliehoek2017gangs, salimans2016improved}. Pursuing high-performance generative networks is non-trivial as challenges may arise from every factor in the training process, from loss functions to network architectures. There is a rich history of research aiming to improve training stabilization and alleviate mode collapse by providing the generator with more meaningful and suitable supervision signals, such as improving generative adversarial losses (e.g., Wasserstein distance \cite{arjovsky2017wasserstein}, Least Squares loss \cite{mao2017least}, and hinge loss \cite{miyato2018spectral}), regularization methods (e.g., gradient penalty \cite{gulrajani2017improved, kodali2017convergence}), and self-supervised methods \cite{sinha2019small, zhang2019consistency, zhao2020improved, karras2020analyzing, zhao2020differentiable}.

Alongside the direction of modifying supervision signals, enhancing the architecture of generators has also proven to be significant for stabilizing training and improving generalization. \cite{radford2015unsupervised} exploits deep convolutional networks in both generator and discriminator, and a series of approaches \cite{miyato2018spectral, gulrajani2017improved, ye2019progressive,brock2018large} show that residual blocks \cite{he2016deep} are capable of facilitating the training of GANs. Recently, StyleGAN family \cite{karras2019style,karras2020analyzing} introduces mapping networks for transforming random latent codes, which are consequentially injected into each convolutional layers as style information. The architectures have shown excellent performance on large datasets including CelebA \cite{liu2015faceattributes}, LSUN-church \cite{yu15lsun}, and Flickr-Faces-HQ \cite{karras2019style}). However, such a manual architecture design typically requires many efforts and domain-specific knowledge from human experts, which is even challenging 
for GANs due to the minimax formulation that it intrinsically has \cite{goodfellow2014generative}. Recent progress of architecture search on a set of supervised learning tasks \cite{zoph2016neural,liu2018darts,zoph2018learning, brock2017smash, liu2018darts} has shown that remarkable achievements can be reached by automating the architecture search process.

In this paper, we focus on improving GAN from the perspective of neural architecture search (NAS). In most of NAS methods, the search process is typically supervised on a validation set through the evaluation measure which is a good surrogate to the final objective of system performance, e.g., substituting accuracy with cross entropy loss, while it is difficult to inspect the on-the-fly quality of both generators and discriminators (except reaching the point of Nash Equilibrium \cite{nash1950equilibrium}) for three reasons. First, the training of GANs is a non-trivial nonconvex-nonconvave bi-level optimization problem \cite{jin2020local}, which is difficult and unstable to optimize and monitor. Second, the loss values of GANs cannot reflect the training status of GANs \cite{brock2018large, grnarova2019domain}. Third, the computation of commonly adopted metrics IS and FID are time-consuming and not differentiable, impeding the development of efficient NAS methods on GANs. 

Inspired from Game theory, which targets at finding pure Nash Equilibrium between pairs of generators and discriminators \cite{salimans2016improved, heusel2017gans, grnarova2019domain}, we propose a fully differentiable architecture search framework for GANs, dubbed {\em alphaGAN}, in which a differential evaluation metric is introduced for guiding architecture search towards pure Nash Equilibrium. We formulate the search process of alphaGAN as a bi-level minimax optimization problem, and solve it efficiently via stochastic gradient-type methods. Specifically, the outer-level objective aims to optimize the generator architecture parameters towards pure Nash Equilibrium, whereas the inner-level constraint targets at optimizing the weight parameters conditioned on the architecture currently searched. The formulation of alphaGAN is a generic form, which is task-agnostic and suitable for any generation tasks in a minimax formulation.

The most related work is AdversarialNAS \cite{gao2019adversarialnas}, which is the first gradient-based NAS-GAN method, searching the architecture of GANs under the loss of GANs. alphaGAN differs from AdversarialNAS in three main aspects.

\begin{itemize}
\item[$\bullet$]The metrics of search are different. alphaGAN aims to search generator architectures towards Nash Equilibrium which is  achieved through the evaluation metric duality gap.  AdversarialNAS searches the architectures by directly using the adversarial loss. As pointed out in \cite{brock2018large, grnarova2019domain, lin2019gradient}, the values of adversarial loss are intractable to describe ``how well GAN has been trained“ due to its specific minimax structure. However, duality-gap as a generic evaluation metric for minimax problems, is capable of measuring the distance of current model to pure Nash Equilibrium, thus is more appropriate for guiding the search.

\item[$\bullet$]Search algorithms are also different. The weight parameters and the architecture parameters of alphaGAN are optimized in the inner level and the outer level, respectively. We expect to find the architectures towards pure Nash Equilibrium on the validation set, in which the weight parameters (including both of generators and discriminators) associated with the architecture are learned on the training set. AdversarialNAS follows the algorithm in the original GAN \cite{goodfellow2014generative}, in which the optimization of generators and discriminators performs alternately. Unlike only optimizing weight parameters of generators (or discriminators) at every iteration in the original GAN, AdversarialNAS also optimizes architecture parameters besides weight parameters.

\item[$\bullet$]The performance of alphaGAN is better. On the datasets CIFAR-10, STL-10, CelebA, LSUN-church and FFHQ, alphaGAN outperforms AdversarialNAS with less computational cost.  alphaGAN also reaches state-of-the-art performance on STL-10, CelebA, LSUN-church, and FFHQ, compared to both manually designed architectures and automatically searched architectures.
\end{itemize}

In addition, we conduct extensive empirical studies on both conventional vanilla GANs \cite{miyato2018spectral} and  state-of-the-art backbone StyleGAN2 \cite{karras2019style}. The experiments show that alphaGAN is capable of discovering high-performance architectures while being much faster than modern GAN architecture search methods, i.e., the obtained architectures can achieve superior performance on CIFAR-10 and state-of-the-art results on STL-10, with much smaller parameter sizes. Comprehensive studies are presented to investigate the effect of search configuration on model performance, which we expect to use for understanding the method. The experiments on StyleGAN2 further demonstrate the effectiveness of the method, i.e., the optimized architecture can boost performance to achieve state-of-the-art results on multiple challenging datasets including CelebA \cite{liu2015faceattributes}, LSUN-church \cite{yu15lsun} and FFHQ \cite{karras2019style}.

The main contributions of the work are summarized:
\begin{itemize}[leftmargin=*]
\item[1)]  We propose a novel architecture search framework for generative adversarial networks, which is mathematically formulated as a bi-level minimax problem. In the inner-level problem, a GAN is trained with the searched architecture over the training dataset. In the outer-level problem, a differentiable evaluation metric is utilized to guide the search process towards pure Nash Equilibrium over the validation dataset. 

\item[2)]We solve the search optimization in a fully differentiable and efficient manner, e.g., the method can obtain an optimal architecture from a larger search space in 3 GPU hours on the CIFAR-10 dataset, 8 times faster than the counterpart method AdversarialNAS \cite{gao2019adversarialnas}. 

\item[3)]The searched architectures are high-performing, which achieve state-of-the-art results on CIFAR-10 and STL-10 over modern GAN search methods. We extend the search space on the StyleGAN2 backbone \cite{karras2020analyzing} by incorporating its characteristic. To the best of our knowledge, alphaGAN is the first NAS-GAN framework deploying and working well on large datasets, which demonstrates the effectiveness of alphaGAN is not confined to certain GAN topology or small datasets. 

\end{itemize}
\section{Related work}\label{related-work-sec}

\textbf{Generative Adversarial Networks (GANs).} GANs are proposed in \cite{goodfellow2014generative}, composed of two networks, generator and discriminator. Discriminator aims to distinguish between generated samples and real ones, while generator aims to generate samples to fool the discriminator. In other words, the two networks are trained in an adversarial manner, leading to instability of training process \cite{salimans2016improved}. 
A family of methods aim to address the issue by enhancing supervision, from the perspectives of introducing novel generative adversarial functions \cite{arjovsky2017wasserstein, mao2017least, miyato2018spectral}, regularization \cite{gulrajani2017improved, kodali2017convergence}, or self-supervised mechanisms \cite{sinha2019small, zhang2019consistency, zhao2020improved, karras2020training, zhao2020differentiable}.

As another direction of improving GANs, exploring high-performance networks has proven to be useful by many works. DCGAN \cite{radford2015unsupervised} introduces convolutional neural networks (CNNs) to the generator and discriminator. Inspired from the dominant prosperity of ResNet \cite{he2016deep}, \cite{miyato2018spectral, zhang2018self, zhang2018self} exploits residual blocks in the architecture of the generator and discriminator, enabling the generation of images with a large resolution (e.g., larger than 64x64) with the representation ability brought by residual blocks. On the basis of the stack of residual blocks, BigGAN \cite{brock2018large} further modifies the architecture of the generator via introducing the information input in the side of the generator, which is viewed as the input of Class Conditional Batch Normalization (CCBN) in BigGAN. StyleGAN \cite{karras2019style} also enables the information input in the side of the generator, exploiting the information to the side as the input of the internal AdaIN \cite{huang2017arbitrary}. Instead of employing residual blocks, StyleGAN2 \cite{karras2020analyzing} employs ``skip" topology in the generator. However, the above methods require the efforts of human experts. In this paper, we hope to promote the architecture of the generator via AutoML. 

\textbf{Neural Architecture Search (NAS).} NAS aims to automatically design the architecture of CNN under the given search space in computer vision tasks, such as image classification \cite{zoph2016neural} \cite{zoph2018learning} \cite{liu2017hierarchical} \cite{liu2018progressive} \cite{real2019regularized} and object detection \cite{ghiasi2019fpn} \cite{chen2019detnas} \cite{peng2019efficient}. NAS can be divided into three types according to the search algorithm, Reinforcement learning based (RL-based), evolution-based, and gradient-based. RL-based NAS \cite{zoph2016neural} \cite{zoph2018learning} exploits accuracy as reward signal and trains a controller to sample the optimal structure. Evolution-based NAS \cite{real2019regularized} exploits evolutionary algorithm to search the architecture. Both RL-based NAS and evolution-based NAS cost more than 2000 GPU-hours. Gradient-based NAS \cite{liu2018darts} relaxes the discrete choice of structure to the continuous search space and obtains the optimal architecture via solving a bi-level optimization problem. Gradient-based NAS can directly search the architecture through gradient descent because it is differentiable. Compared with RL-based NAS and evolution-based NAS, gradient-based NAS is time-efficient, costing less than 100 GPU-hours.

\begin{table*}
  \caption{The summary of NAS-GAN methods.}
  \label{Summary_NAS-GAN}
  \vspace{-0.3cm}
  \centering
  \resizebox{0.8\textwidth}{!}{
  \begin{tabular}{llll}
    \toprule
    Method & Type & Evaluation metric & Task \\
    \midrule
    AutoGAN(\cite{gong2019autogan}) & \multirow{4}*{RL based} & IS & \multirow{3}*{Conventional GANs} \\
    AGAN(\cite{wang2019agan}) &  & IS & \\
    E$^2$GAN(\cite{tian2020off}) &  & IS + FID & \\
    TPSR(\cite{lee2020journey}) &  & PSNR LPIPS & Super Resolution \\
    AdversarialNAS(\cite{gao2019adversarialnas}) & gradient based & Generative Adversarial Functions & Conventional GANs \\
    \midrule
    alphaGAN & gradient based & Duality-gap & Conventional GANs/StyleGAN2 \\
    \bottomrule
    \vspace{-0.5cm}
    \end{tabular}}
     \vspace{-0.4cm}
\end{table*}

Currently, previous works \cite{gong2019autogan, wang2019agan, tian2020off} employ NAS to automatically search the architecture of the generator with a reinforcement learning paradigm, rewarded by Inception Score \cite{salimans2016improved} or Fréchet Inception Distance \cite{heusel2017gans}, which are task-dependent and non-differential metrics. AutoGAN \cite{gong2019autogan} exploits Inception score as the reward signal and trains a controller to sample the optimal architecture of the generator. Analogously, AGAN \cite{wang2019agan} searches the architecture of the generator via reinforcement learning under a larger search space.  However, RL-based search requires enormous computation resources. E2GAN \cite{tian2020off} exploits off-policy reinforcement learning to efficiently search the architecture of the generator, rewarded by both IS and FID, of which the computation is also time-consuming. On the other hand,  Gao et al. \cite{gao2019adversarialnas} proposed the first gradient-based NAS-GAN framework, in which traditional minimax loss functions are exploited to optimize architecture and network parameters both. The recent NAS-GAN methods are summarized in Tab. \ref{Summary_NAS-GAN}.

In particular,  AdversarialNAS \cite{gao2019adversarialnas} is the most related work to our proposed alphaGAN, which directly exploits the minimax objective function of GANs to optimize both the architecture parameters and the weight parameters. However, many previous works \cite{brock2018large, grnarova2019domain, lin2019gradient} have pointed out that the minimax objective function of GANs cannot reflect the status of GANs. Thus, a more suitable evaluation metric to guide the search of the architecture of the generator is essential. 

\begin{figure}[t]
  \centering
  \includegraphics[width=0.5\textwidth]{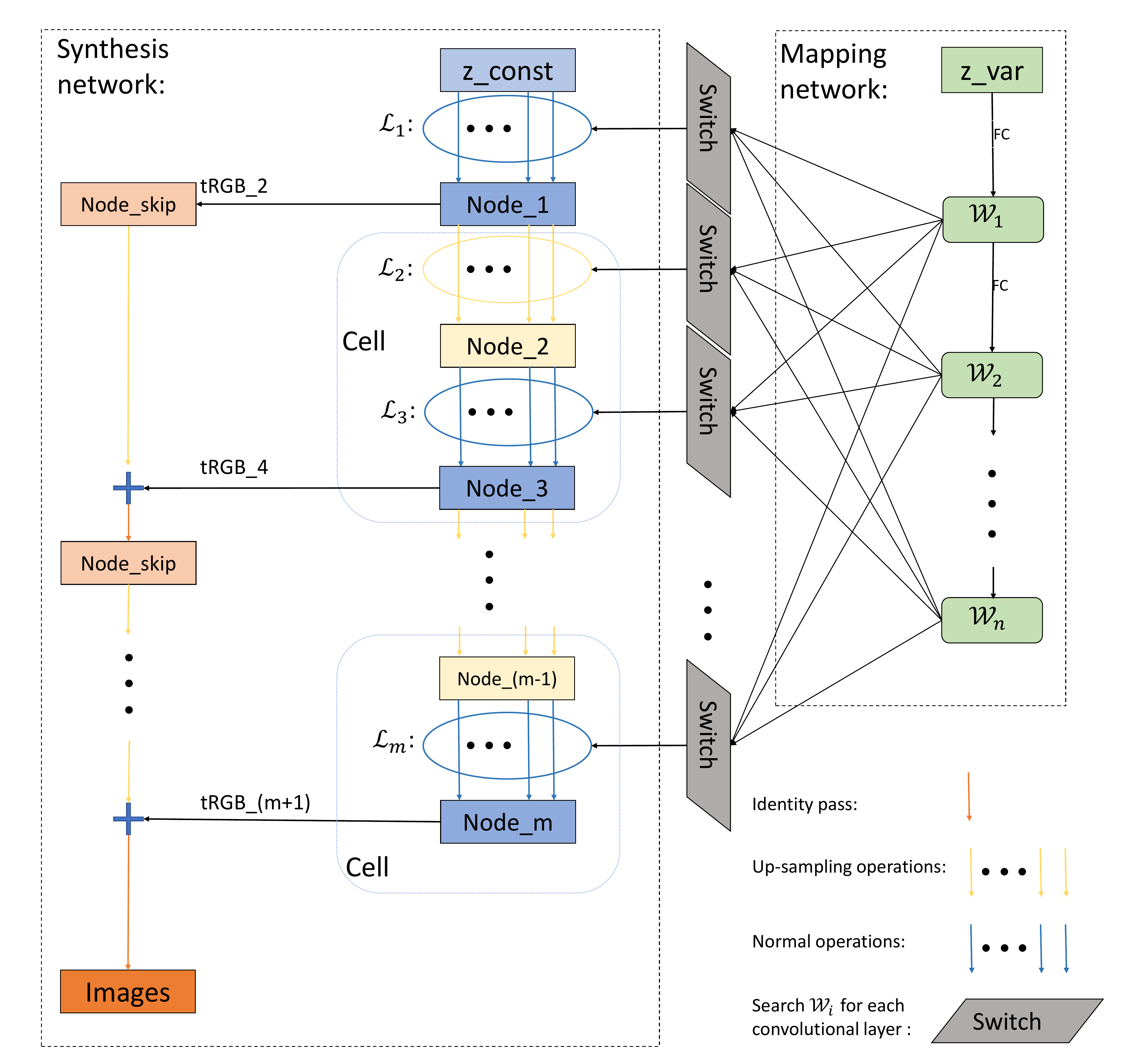}
  \vspace{-0.6cm}
  \caption{Searching the operations and the intermediate latent $\left\{\mathcal{W}_{i}\right\}_{i=1,...,n}$ of StyleGAN2. ``tRGB\_2", ``tRGB\_4", and ``tRGB\_(m+1)" denote the skip connection layers in the original StyleGAN2 \cite{karras2020analyzing}, which are conv\_1x1 operations. The number (e.g., 2) after the underscore in ``tRGB\_2" denotes that ``tRGB\_2" receives the latent $\mathcal{W}_{i}$ identical to the latent received by the convolutional layer $\mathcal{L}_{2}$.}
  \label{The depiction of StyleGAN2}
  \vspace{-0.5cm}
\end{figure}

\section{Preliminaries}\label{preliminary-section}

Minimax Games have regained a lot of attraction \cite{osborne1994course, du2013minimax} since they are popularized in machine learning, such as generative adversarial networks (GAN) \cite{salimans2016improved}, reinforcement learning \cite{ho2016generative, pinto2017robust}, etc.
Given the function ${\rm Adv}\!:\mathbb{X} \times \mathbb{Y} \to \mathbb{R}$, we consider a minimax game and its dual form: 
\begin{gather*}
\min_{G}\! \max_{D} {\rm Adv}(G,D) \!=\! \min_{G}\big\{\max_{D} {\rm Adv}(G,D)\big\}, \\
\max_{D}\! \min_{G} {\rm Adv}(G,D) \!=\! \max_{D}\big\{\min_{G} {\rm Adv}(G,D)\big\}.
\end{gather*}
Nash Equilibrium \cite{nash1950equilibrium} of a minimax game can be used to characterize the best decisions of two players $G$ and $D$ for the above minmax game, where no player can benefit by changing strategy while the other players keep unchanged.
\begin{definition}\label{pure-equilibrium}
$(\overline{G},\overline{D})$ is called Nash Equilibrium of game $\min_{G}\max_{D}{\rm Adv}(G,D)$ if
\begin{align}\label{pure-equilibrium-eq1}
    {\rm Adv}(\overline{G},D)\leq{\rm Adv}(\overline{G},\overline{D})\leq{\rm Adv}(G,\overline{D})
\end{align}
\end{definition}\label{mixed nash equilibrium}
holds for any $(G,D)$ in $\mathbb{X} \times \mathbb{Y}$, where $\overline{G} = \arg\min_{G} {\rm Adv}(G,D)$ and $\overline{D} = \arg\max_{D} {\rm Adv}(G,D)$. When a minimax game equals to its dual problem, $(\overline{G},\overline{D})$ is Nash Equilibrium of the game. Hence, the gap between the minimax problem and its dual form can be used to measure the degree of approaching Nash Equilibrium \cite{grnarova2019domain}.

Generative Adversarial Network (GAN) proposed in \cite{goodfellow2014generative} is mathematically defined as a minimax game problem with a binary cross-entropy loss of competing between the distributions of real and synthetic images generated by the GAN model.  
Despite remarkable progress achieved by GANs, training high-performance models is still challenging for many generative tasks due to its fragility to almost every factor in the training process. Architectures of GANs have proven useful for stabilizing training and improving generalization \cite{miyato2018spectral, gulrajani2017improved, ye2019progressive,brock2018large}, and we hope to discover architectures by automating the design process with limited computational resource in a principled differentiable manner.

\section{GAN Architecture Search as Fully Differential Optimization}\label{alphaGAN-section}

 
\subsection{Formulation of differentiable architecture search}\label{alphaGAN-formulation-subsection}
Differentiable Architecture Search was first proposed in \cite{liu2018darts}, where the problem is formulated as a bi-level optimization:
\begin{align}\label{darts-formulation}
    & \min_{\alpha} \mathcal{L}_{val}(\alpha, \omega), \quad s.t. \quad\omega = \arg\min_{\omega} \mathcal{L}_{train}(\alpha, \omega),
\end{align}
where $\alpha$ and $\omega$ denote the optimized variables of architectures and network parameters, respectively. DARTS aims to seek the optimal architecture that performs best on the validation set with the network parameters trained on the training set. The search process is supervised by the cross-entropy loss, a good surrogate for the metric accuracy.

However, deploying such a framework for searching architectures of GANs is non-trivial. The training of GANs corresponds to the optimization of a minimax problem (as shown above), which learns an optimal generator trying to fool the additional discriminator. However, generator evaluation is independent of discriminators but based on some extra metrics (e.g., Inception Score \cite{salimans2016improved} and FID \cite{heusel2017gans}), which are typically discrete and task-dependent.

{ \bf Evaluation function.} 
Using a suitable and differential evaluation metric function to inspect the on-the-fly quality of both the generator and discriminator is necessary for a GAN architecture search framework. Due to the intrinsic minimax property of GANs, the training process of GANs can be viewed as a zero-sum game as in \cite{salimans2016improved,goodfellow2014generative}. The zero-sum game includes two players competing in an adversarial manner. The universal objective of training GANs consequentially can be regarded as reaching the pure equilibrium in Definition \ref{pure-equilibrium}. 
Hence, we adopt the primal-dual gap function \cite{grnarova2019domain, peng2020dggan} for evaluating the generalization of vanilla GANs. Given a pair of $G$ and $D$, the duality-gap function is defined as 
\begin{align}\label{evaluation-function}
     \mathcal{V}(G,D) 
    &= {\rm Adv}(G,\overline{D}) - {\rm Adv}(\overline{G},D) \\
    &:=  \max_{D}{\rm Adv}(G,D)-\min_{G}{\rm Adv}(G,D). \nonumber
\end{align}
The evaluation metric $\mathcal{V}(G,D)$ is non-negative and $V(G,D)=0$ can only be achieved when Nash Equilibrium in Definition \eqref{pure-equilibrium} holds. $\mathcal{V}(G,D)$ consists of solving the global optimum $\overline{G}$ and $\overline{D}$. Function \eqref{evaluation-function} provides a quantified measure of describing ``how close is the current GAN to the pure equilibrium", which can be used for assessing the training status.

For neural architecture search, the objective is to find the architecture working best on the validation set, where the associated weight parameters are supposed to be optimal on the training set. From this perspective, the architecture search for GANs can be formulated as a specific bi-level optimization problem:
\begin{align}\label{abstract-formulation}
    & \min_{\alpha}\left\{\mathcal{V}(G,D): (G,D) := \arg\min_{G} \max_{D} {\rm Adv}\, (G,D)\right\},
\end{align}
where $\mathcal{V}(G,D)$ performs on the validation dataset and supervises seeking the optimal generator architecture as an outer-level problem, and the inner-level optimization on $ {\rm Adv}\, (G,D)$ aims to learn suitable network parameters (including both the generator and discriminator) for GANs based on the current architecture and the training dataset. 


In this work, we exploit the hinge loss from \cite{miyato2018spectral,zhang2018self} or non-saturating loss form \cite{goodfellow2014generative}.

{\bf AlphaGAN formulation.}
By integrating the generative adversarial function (hinge loss or non-saturating loss) and evaluation function \eqref{evaluation-function} into the bi-level optimization \eqref{abstract-formulation}, we can obtain the final objective for the framework as follows,
\begin{align}
    & \min_{\alpha} \mathcal{V}_{val}(G,D) = {\rm Adv}(G,\overline{D}) - {\rm Adv}(\overline{G},D) \label{alphaGAN-formulation-a}\\
    & \;s.t.\quad \omega\in\arg\min_{\omega_{G}} \max_{\omega_{D}} {\rm Adv}_{train}(G,D) \label{alphaGAN-formulation-b}
\end{align}
where architecture parameters $\alpha$ are optimized on the validation dataset, conditioned on that weight parameters $\omega$ are optimal on the training dataset. According to the formulation \eqref{alphaGAN-formulation-a} \eqref{alphaGAN-formulation-b} of alphaGAN, weight parameters $\omega$ are optimized in the inner level and architecture parameters $\alpha$ are optimizaed in the outer level, respectively.

Generator $G$ and discriminator $D$ are parameterized with variables $(\alpha_{G},\omega_{G})$ and $(\omega_{D})$, respectively,  $\overline{D} = \arg\max_{D}{\rm Adv}_{val}(G,D)$, and $\overline{G} = \arg\min_{G}{\rm Adv}_{val}(G,D)$. 
The search process contains three parts of parameters, weight parameters $\omega=(\omega_{G},\omega_{D})$, test-weight parameters $\overline{\omega}=(\omega_{\overline{G}},\omega_{\overline{D}})$, and architecture parameters $\alpha=(\alpha_{G})$. The architecture of the discriminator can be optimized in this framework, however we are mainly concerned with the architecture of the generator for two reasons. First, in traditional GAN methods, a discriminator is used to supervise the training of the generator, and will be omitted during inference. Second, we find that the impact of searching discriminator for seeking better generator architectures is marginal and even hampers the process in practice (more details can be found in Section \ref{search D?}).

\subsection{Algorithm and Optimization}\label{alphaGAN-optimization-subsection}


In this section, we will give a detailed description of the training algorithm and optimization process of alphaGAN. We first describe the entire network structure of the generator and the discriminator, the search space of the generator, and the continuous relaxation of architectural parameters.


\begin{figure}[t!]
  \centering
  \subfigure[\texttt{Generator}]{\includegraphics[angle=90, scale=0.5]{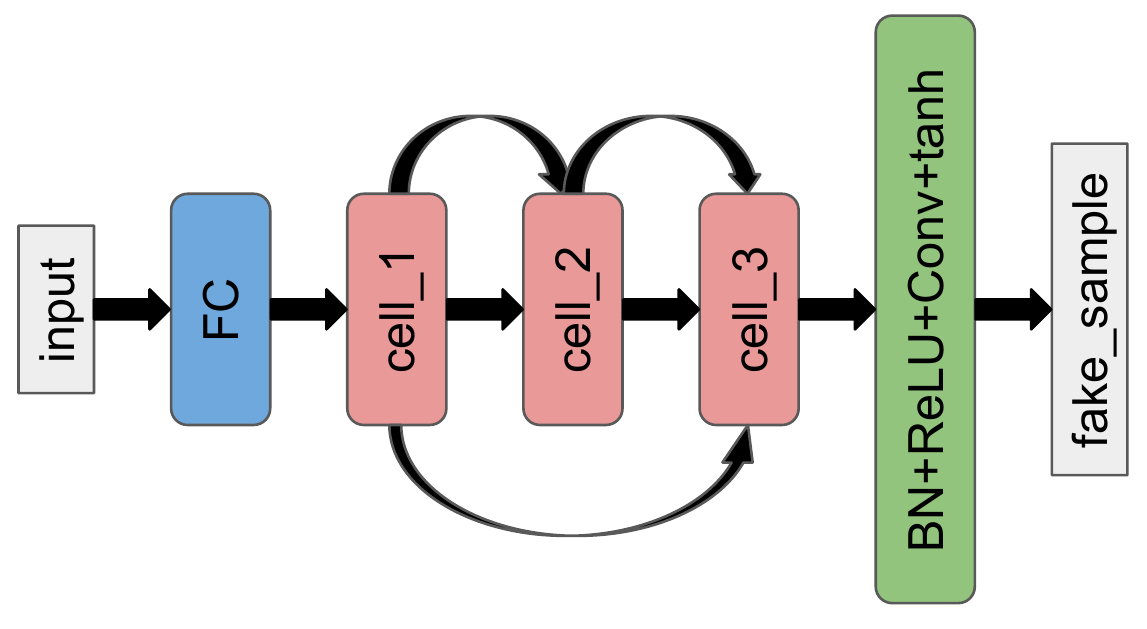}}\hspace{0.5cm}
  \subfigure[\texttt{Cell}]{\includegraphics[scale=0.4]{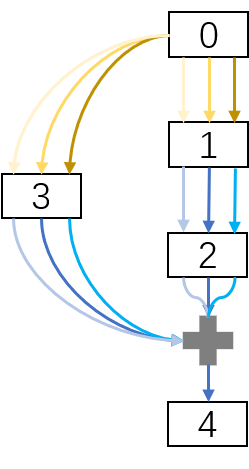}}
  \vspace{-0.3cm}
  \caption{The topology of the generator and the cell in conventional GANs.}
  \label{The structure of G}
\end{figure}

\textbf{Base Backbone of $G$ and $D$.} We deploy alphaGAN on both conventional GANs and StyleGAN2. As for conventional GANs, the illumination of the entire structure is shown in Fig. \ref{The structure of G}. The entire structure is identical to those in Auto-GAN \cite{gong2019autogan} and SN-GAN \cite{miyato2018spectral}, composed of several cells, as shown in Fig. \ref{The structure of G}. Each cell, regarded as a directed acyclic graph, is comprised of the nodes representing intermediate feature maps and the edges connecting pairs of nodes via different operations. 

Regarding StyleGAN2, the illustration of the entire structure of it is shown in Fig. \ref{The depiction of StyleGAN2}. Alongside the operations to be searched, we additionally search the intermediate latent fed into the convolutional layers, elaborated in Sec. \ref{Search space of StyleGAN2}.

We apply a fixed network architecture for the discriminator, based on the conventional design as \cite{miyato2018spectral} or the design as StyleGAN2 \cite{karras2020analyzing}.

\renewcommand\algorithmiccomment[1]{%
  \hfill\# \ \eqparbox{}{#1}%
}

\begin{algorithm}[t!]
\caption{Searching the architecture of alphaGAN}
\label{Solving-alphaGAN}
{\bf Parameters:} Initialize weight parameters ($\omega^{1}_{G}$,$\omega^{1}_{D}$). Initialize generator architecture parameters $\alpha^{1}_{G}$. Initialize base learning rate $\eta$, momentum parameter $\beta_1$, and exponential moving average parameter $\beta_2$ for Adam optimizer. 
\begin{algorithmic}[1] 
\FOR {$k=1,2,\cdots, K$}
\STATE Set $(\omega^{k,1}_{G}, \omega^{k,1}_{D}) = (\omega^k_{G}, \omega^k_{D})$ and set $\alpha^{k,1}_{G}= \alpha^k_{G}$;
\FOR [``weight\_part"] {$t=1,2,\cdots, T$}  
\STATE Sample real data $\{x^{(l)}\}_{l=1}^{m}\sim\mathbb{P}_{r}$ from the training set and noise $\{z^{(l)}\}_{l=1}^{m}\sim\mathbb{P}_{z};$ Estimate the gradient of {\rm Adv} loss in Eq.~\eqref{alphaGAN-formulation-b} with $\{x^{(l)},z^{(l)}\}$ at $(\omega^{k,t}_{G},\omega^{k,t}_{D})$, dubbed $\nabla{\rm Adv}(\omega^{k,t}_{G},\omega^{k,t}_{D});$
\STATE $\omega^{k,t+1}_{D}= {\rm Adam}\big(\nabla_{\omega_{D}}{\rm Adv}(  \omega^{k,t}_{G},\omega^{k,t}_{D}),\omega^{k,t}_{D},\eta,\beta_{1},\beta_{2}\big);$
\STATE $\omega^{k,t+1}_{G}= {\rm Adam}\big(\nabla_{\omega_{G}}{\rm Adv}(\omega^{k,t}_{G},\omega^{k,t}_{D}),\omega^{k,t}_{G},\eta,\beta_{1},\beta_{2}\big);$
\ENDFOR 
\STATE Set $(\omega^{k+1}_{G},\omega^{k+1}_{D})=(\omega^{k,T}_{G},\omega^{k,T}_{D})$;
\STATE Receive architecture searching parameter $\alpha_{G}^{k}$ and network weight parameters ($\omega^{k+1}_{G}$,$\omega^{k+1}_{D}$); Estimate neural architecture parameters $({\omega}^{k+1}_{\overline{G}},{\omega}^{k+1}_{\overline{D}})$ of $(\overline{G}, \overline{D})$ via Algorithm \ref{Solving-GD};  \COMMENT{``test\_weight\_part"}
\FOR [``arch\_part"] {$s=1,2,\cdots,S$}
\STATE Sample real data $\{x^{(l)}\}_{l=1}^{m}\sim\mathbb{P}_{r}$ from the validation set and latent variables $\{z^{(l)}\}_{l=1}^{m}\sim\mathbb{P}_{z}$. Estimate the gradient of the $\mathcal{V}$ loss in Eq.~\eqref{evaluation-function} with $\{x^{(l)},z^{(l)}\}$ at $(\alpha^{k,s})$, dubbed $\nabla{\mathcal{V}}(\alpha_{G}^{k,s});$\\
\STATE $\alpha_{G}^{k,s+1}= {\rm Adam}(\nabla_{\alpha_{G}}\mathcal{V}(\alpha_{G}^{k,s}),\alpha_{G}^{k,s},\eta,\beta_{1},\beta_{2});$
\ENDFOR
\STATE Set $\alpha_{G}^{k+1}$=$\alpha_{G}^{k,S};$
\ENDFOR
\STATE Return $\alpha_{G}$=$\alpha_{G}^{K}.$
\end{algorithmic}
\end{algorithm}

\begin{algorithm}
\caption{Solving $\overline{G}$ and $\overline{D}$}
\label{Solving-GD}
{\bf Parameters:} Receive architecture searching parameter $\alpha_{G}$ and weight parameter ($\omega_{G}$,$\omega_{D}$). Initialize weight parameter  $(\omega_{\overline{G}}^{1},\omega_{\overline{D}}^{1})=(\omega_{G},\omega_{D})$ for $(\overline{G},\overline{D})$. Initialize base learning rate $\eta$, momentum parameter $\beta_1$, and EMA parameter $\beta_2$ for Adam optimizer. 
\begin{algorithmic}[1] 
\FOR {$r=1,2,\cdots, R$}
    \STATE Sample real data $\{x^{(l)}\}_{l=1}^{m}\sim\mathbb{P}_{r}$ from the validation dataset and noise $\{z^{(l)}\}_{l=1}^{m}\sim\mathbb{P}_{z};$ Estimate the gradient of the {\rm Adv} loss in Eq.~\eqref{alphaGAN-formulation-a} with $\{x^{(l)},z^{(l)}\}$ at $(\omega_{G},\omega_{\overline{D}}^{r})$, dubbed $\nabla{\rm Adv}(\omega_{G},\omega_{\overline{D}}^{r});$
    \STATE $\omega_{\overline{D}}^{r+1}= {\rm Adam}\big(\nabla_{\omega_{\overline{D}}}{\rm Adv}(  \omega_{G},\omega_{\overline{D}}^{r}),\omega_{\overline{D}}^{r},\eta,\beta_{1},\beta_{2}\big);$
\ENDFOR
\FOR {$r=1,2,\cdots, R$}
    \STATE Sample noise $\{z^{(l)}\}_{l=1}^{m}\sim\mathbb{P}_{z};$ Estimate the gradient of the {\rm Adv} loss in Eq.~\eqref{alphaGAN-formulation-a} with $\{z^{(l)}\}$ at point $(\omega_{\overline{G}}^{r},\omega_{D})$, dubbed $\nabla{\rm Adv}(\omega_{\overline{G}}^{r},\omega_{D})$;
    \STATE $\omega_{\overline{G}}^{r+1}= {\rm Adam}\big(\nabla_{\omega_{\overline{G}}}{\rm Adv}(\omega_{\overline{G}}^{r},\omega_{D}),\omega_{\overline{G}}^{r},\eta,\beta_{1},\beta_{2}\big);$
\ENDFOR 
    \STATE Return $(\omega_{\overline{G}},\omega_{\overline{D}})=(\omega_{\overline{G}}^{R},\omega_{\overline{D}}^{R}).$
\end{algorithmic}
\end{algorithm}

\textbf{Search space of $G$.} The search space is compounded from two types of operations, i.e., normal operations and up-sampling operations. 
As for conventional GANs, the pool of normal operations, denoted as $\mathcal{O}_nc$, is comprised of \{conv\_1x1,\quad conv\_3x3,\quad conv\_5x5,\quad sep\_conv\_3x3,\quad  sep\_conv\_5x5,\quad sep\_conv\_7x7\}. The pool of up-sampling operations, denoted as $\mathcal{O}_uc$, is comprised of \{
deconv,\quad nearest,\quad bilinear\}, where ``deconv'' denotes the ConvTransposed\_2x2.
operation. alphaGAN allows {$(6^3\times3^2)^3\times3^3\approx2\times10^{11}$} possible configurations for the conventional generator architecture, which is larger than $\sim10^5$ of AutoGAN \cite{gong2019autogan}.

Regarding StyleGAN2, the pool of normal operations, denoted as $\mathcal{O}_{ns}$, is comprised of \{conv\_1x1,\quad conv\_3x3,\quad conv\_5x5\}. The pool of up-sampling operations, denoted as $\mathcal{O}_{us}$, is comprised of \{
deconv,\quad nearest\_conv,\quad bilinear\_conv\}, where ``nearest\_conv'' denotes the nearest interpolation followed by conv\_3x3 and ``bilinear\_conv" denotes the bilinear interpolation followed by conv\_3x3, respectively. We will elaborate on why we design that search space for StyleGAN2 in Sec. \ref{Scalability}. alphaGAN allows up to $8^{18}\times3^{17}\approx2.3\times10^{24}$ possible configurations for the generator architecture.

\textbf{Continuous relaxation.} The discrete selection of operations is approximated by using a soft decision with a mutually exclusive function, following \cite{liu2018darts}. Formally, let $o \in \mathcal{O}_n$ denote some normal operations on node $i$, and $\alpha^o_{i,j}$ represent the architectural parameter with respect to the operation between node $i$ and its adjacent node $j$, respectively. Then the node output induced by the input node $i$ can be calculated by
\begin{align}
    O_{i,j}(x)=\sum_{o\in \mathcal{O}_n}\frac{\exp\big(\alpha_{i,j}^o\big)}{\sum_{o'\in \mathcal{O}_n}\exp\big(\alpha_{i,j}^{o'}\big)}o(x),
\end{align}
and the final output is summed over all of its preceding nodes, i.e., $x^j = \sum_{i \in Pr(j)}O_{i,j}(x^i)$. The selection on up-sampling operations follows the same procedure.

{\bf Solving alphaGAN.} We apply an alternating minimization method to solve alphaGAN \eqref{alphaGAN-formulation-a}-\eqref{alphaGAN-formulation-b} with respect to variables $\big((\omega_G,\omega_D), (\omega_{\overline{G}},\omega_{\overline{D}}), \alpha_{G}\big)$ in Algorithm \ref{Solving-alphaGAN}, which is fully differentiable. Algorithm \ref{Solving-alphaGAN} is composed of three parts. The first part (line 3-8), called ``weight\_part", aims to optimize weight parameters $\omega$ on the training dataset via Adam optimizer \cite{kingma2014adam}. The second part (line 9), called ``test-weight\_part", aims to optimize the weight parameters $(\omega_{\overline{G}},\omega_{\overline{D}})$, and the third part (line 10-12), called 'arch\_part', aims to optimize architecture parameters $\alpha_{G}$ by minimizing the duality gap. Both 'test-weight\_part' and 'arch\_part' are optimized over the validation dataset via Adam optimizer. Algorithm \ref{Solving-GD} illuminates the detailed process of computing $\overline{G}$ and $\overline{D}$ by updating weight parameters $(\omega_{\overline{G}},\omega_{\overline{D}})$ with last searched generator network architecture parameters $\alpha_G$ and related network weight parameters $(\omega_G,\omega_D)$. In summary, the variables $\big((\omega_G,\omega_D), (\omega_{\overline{G}},\omega_{\overline{D}}), \alpha_{G}\big)$ are optimized in an alternating fashion.
\section{Evaluation on Conventional GANs}\label{Conventional GANs}
In this section, we conduct extensive experiments on the topology of conventional GANs like \cite{miyato2018spectral, gong2019autogan, arjovsky2017wasserstein}, comprised of residual blocks, to explore the optimal configurations of alphaGAN. We want to emphasize that some configurations are significant for the effectiveness and the versatility of alphaGAN, e.g., exploiting the discriminator with the fixed architecture, and updating the weight parameters $\omega_{G}$ and $\omega_{D}$ while approximating $\overline{G}$ and $\overline{D}$, respectively.


During searching, we use a minibatch size of 64 for both generators and discriminators, and the channel number of 256 for generators and 128 for discriminators. When re-training the network, we exploit the identical discriminator in AutoGAN. We use a minibatch size of 128 for generators and 64 for discriminators. The channel number is set to 256 for generators and 128 for discriminators, respectively. As the architectures of discriminators are not optimized variables, an identical architecture is used for searching and re-training (the configuration is the same as in \cite{gong2019autogan}). These configurations are utilized by default except we state otherwise. When testing, 50,000 images are generated with random noise, and IS \cite{salimans2016improved} and FID \cite{heusel2017gans} are used to evaluate the performance of generators. GPU we use is Tesla P40. More details of the experimental setup and empirical studies about the rest of the proposed method can be found in the appendix.


\newcommand{\tabincell}[2]{\begin{tabular}{@{}#1@{}}#2\end{tabular}}
\begin{table*}
  \caption{Comparison with state-of-the-art GANs on CIFAR-10. $\dagger$ denotes the results reproduced by us, with the structure released by AutoGAN and trained under the same setting as AutoGAN.}
  \label{Effectivenesss on CIFAR-10}
  \centering
  \vspace{-0.3cm}
  \resizebox{\textwidth}{!}{
  \begin{tabular}{c|cccc|cc|cc}
    \toprule
    \multirow{2}*{Architecture} & \multicolumn{4}{|c|}{Search} & \multicolumn{2}{c|}{Re-train} & \multirow{2}*{\tabincell{c}{IS\\ ($\uparrow$ is better)}} & \multirow{2}*{\tabincell{c}{FID\\ ($\downarrow$ is better)}} \\
    \cline{2-7}
     & \tabincell{c}{Dataset} & \tabincell{c}{Cost\\ (GPU-hours)} & \tabincell{c}{Search\\ space} & \tabincell{c}{Search\\ method} & \tabincell{c}{Params\\ (M)} & \tabincell{c}{FLOPs\\ (G)} &  &  \\
    \midrule
    \midrule
    SN-GAN(\cite{miyato2018spectral}) & - & - & - & manual & - & - & $8.22\pm0.05$ & $21.7\pm0.01$\\
    Progressive GAN(\cite{karras2017progressive}) & - & - & - & manual &  &  & 8.80$\pm$0.05 & -\\
    WGAN-GP, ResNet(\cite{gulrajani2017improved}) & - & - & - & manual &  &  & $7.86\pm0.07$ & -\\
    StyleGAN2 + DiffAugment(\cite{zhao2020differentiable}) & - & - & - & manual & - & - & - & $9.89$ \\
    ADA(\cite{karras2020training}) & - & - & - & manual & - & - & $\textbf{9.83}\pm\textbf{0.04}$ & $\textbf{2.92}\pm\textbf{0.05}$ \\
    \midrule
    \midrule
    AutoGAN(\cite{gong2019autogan}) & CIFAR-10 &  & $\sim10^5$ & RL & $5.192$ & $1.77$ & $8.55\pm0.1$ & $12.42$ \\
    \hline
    AutoGAN$\dagger$ & CIFAR-10 & 82 & $\sim10^5$ & RL & $5.192$ & $1.77$ & $8.38\pm0.08$ & $13.95$ \\
    \hline
    AGAN(\cite{wang2019agan}) & CIFAR-10 & $28800$ & $\sim20000$ & RL & - & - & $8.29\pm0.09$ & $30.5$ \\
    \hline
    E$^2$GAN(\cite{tian2020off}) & CIFAR-10 & $7$ & $\sim10^5$ & RL & - & - & $8.51\pm0.13$ & $11.26$ \\
    \hline
    Random search(\cite{li2019random}) & CIFAR-10 & $40$ & $\sim2\times10^{11}$ & Random & $2.701$ & $1.11$ & $8.46\pm0.09$ & $15.43$ \\
    \hline
    AdversarialNAS(\cite{gao2019adversarialnas}) & CIFAR-10 & $24$ & $\sim10^{38}$ & gradient & $8.85$ & $2.68$ & $8.74\pm0.07$ & $10.87$ \\
    \hline
    alphaGAN$_{(l)}$ & CIFAR-10 & $22$ & $\sim2\times10^{11}$ & gradient & $8.618$ & $2.78$ & $8.88\pm0.12$ & $11.34$\\
    \hline
    alphaGAN$_{(s)}$ & CIFAR-10 & $3$ & $\sim2\times10^{11}$ & gradient & $2.953$ & $1.32$ & $\textbf{8.98}\pm\textbf{0.09}$ & $\textbf{10.35}$\\
    \bottomrule
  \end{tabular}}
  \vspace{-0.6cm}
\end{table*}

\begin{table}
  \renewcommand\arraystretch{1.1}
  \caption{Results on STL-10. The structures of alphaGAN$_{(l)}$ and alphaGAN$_{(s)}$ are searched on CIFAR-10 and re-trained on STL-10. $\dagger$ denotes the reproduced results, with the architectural configurations released by the original papers.}
  \vspace{-0.3cm}
  \resizebox{0.5\textwidth}{!}{
  \label{Transfer ability on STL-10}
  \centering
  \begin{tabular}{lllll}
    \toprule
    \tabincell{c}{Architecture} & \tabincell{c}{Params\\ (M)} & \tabincell{c}{FLOPs\\ (G)} & IS & FID \\
    \midrule
    SN-GAN(\cite{miyato2018spectral}) & - & - & $9.10\pm0.04$ & $40.1\pm0.5$\\
    ProbGAN(\cite{he2019probgan}) & - & - & $8.87\pm0.095$ & $46.74$\\
    Improving MMD GAN(\cite{wang2018improving}) & - & - & $\textbf{9.36}$ & $\textbf{36.67}$\\
    \midrule
    Auto-GAN(\cite{gong2019autogan})     & $5.853$ & $3.98$ & $9.16\pm0.12$ & $31.01$     \\
    Auto-GAN$\dagger$     & $5.853$ & $3.98$ & $9.38\pm0.08$ & $27.69$ \\  
    AGAN(\cite{wang2019agan})  & - & - & $9.23\pm0.08$ & $52.7$ \\
    E$^2$GAN(\cite{tian2020off}) & - & - & $9.51\pm0.09$ & $25.35$ \\
    AdversarialNAS(\cite{gao2019adversarialnas}) & $13.26$ & $6.03$ & $9.63\pm0.19$ & $26.98$\\
    alphaGAN$_{(l)}$     & $9.279$ & $6.26$ & $9.53\pm0.12$ & $24.52$\\
    alphaGAN$_{(s)}$     & $3.613$ & $2.97$ & $\textbf{10.12}\pm\textbf{0.13}$ & $\textbf{22.43}$\\ 
    \bottomrule
  \end{tabular}}
  \vspace{-0.4cm}
\end{table}


\subsection{Searching on CIFAR-10}\label{effectiveness of alphaGAN}
 We first compare our method with recent automated GAN methods. During the searching process, the entire dataset is randomly split into two sets for training and validation, respectively, each of which contains 25,000 images. For a fair comparison, we report the performance of best run (over 3 runs) for reproduced baselines and ours in Table \ref{Effectivenesss on CIFAR-10}, and provide the performance of several representative works with manually designed architectures for reference. As there is inevitable perturbation on searching optimal architectures due to stochastic initialization and optimization \cite{arber2019understanding}, we provide a detailed analysis and discussion about the searching process in Sec. \ref{analysis}. And the statistical properties of architectures searched by alphaGAN are in the appendix.
 
 Performances of alphaGAN with two search configurations are shown in Tab. \ref{Effectivenesss on CIFAR-10} by adjusting step sizes $T$, $S$, and $R$ for updating the ``weight\_part", ``arch\_part", and ``test-weight\_part" in Algorithms \ref{Solving-alphaGAN} and \ref{Solving-GD}, respectively, where alphaGAN$_{(l)}$ represents passing through every epoch on the training and validation sets for each loop, i.e., $T=390$ and $R=390$. Whereas alphaGAN$_{(s)}$ represents using smaller interval steps with $T=20$ and $R=20$. alphaGAN$_{(l)}$ and alphaGAN$_{(s)}$ share the same step size of ``arch\_part", i.e., $S=20$.

The results show that our method performs well in the two configurations and achieves superior results, compared with AdversarialNAS \cite{gao2019adversarialnas}. Compared to automated baselines, alphaGAN has shown a substantial advantage in searching efficiency. Particularly, alphaGAN$_{(s)}$ can attain the best trade-off between efficiency and performance, and it can achieve comparable results by searching in a large search space (significantly larger than RL-based baselines) in a considerably efficient manner (i.e., only 3 GPU hours compared to the baselines with tens to thousands of GPU hours). Moreover, the architecture obtained by alphaGAN$_{(s)}$ is light-weight and computationally efficient, which reaches a good trade-off between performance and computation complexity. We also conduct the experiments of searching on STL-10 (shown in Tab. \ref{Effectivenesss on CIFAR-10}) and observe consistent phenomena, demonstrating that the effectiveness of our method is not confined to the CIFAR-10 dataset.

\subsection{Transferability on STL-10}\label{transfer ability}


To validate the transferability of the architectures obtained by alphaGAN, we directly train models by exploiting the obtained architectures on STL-10. The results are shown in Tab. \ref{Effectivenesss on CIFAR-10}.  Both alphaGAN$_{(l)}$ and alphaGAN$_{(s)}$ show remarkable superiority in performance over the baselines with either automated or manually designed architectures. It reveals the benefit that the architecture searched by alphaGAN can be effectively exploited across datasets. It is surprising that alphaGAN$_{(s)}$ is best-behaved, which achieves the best performance in both IS and FID scores. It also shows that compared to an increase in the model complexity, appropriate selection and composition of operations can contribute to model performance in a more efficient manner which is consistent with the primary motivation of NAS.   


\subsection{Ablation Study}\label{Better understand}
We conduct ablation experiments on CIFAR-10 to explore the optimal configurations for alphaGAN, including the studies with the questions: the effect of searching the discriminator architecture, the manner of obtaining the optimal generator $\overline{G}$, the effect of the channels in search, and the effect of step sizes $S$ of ``arch\_part".

\begin{table}[h]
  \renewcommand\arraystretch{1.1}
  \caption{Ablation studies on CIFAR-10.}
  \label{Ablation study summary}
  \vspace{-0.3cm}
  \centering
  \resizebox{0.5\textwidth}{!}{
  \begin{tabular}{llllll}
    \toprule
    Type & Search D? & \multicolumn{2}{c}{Obtain $\overline{G}$} & IS & FID \\
    \cmidrule{3-4}
    & & Update $\alpha_{G}$ & Update $\omega_{G}$ \\
    \midrule
    \multirow{5}*{alphaGAN$_{(l)}$} & $\checkmark$ & $\times$ & $\checkmark$ & $8.51\pm0.09$ & $18.07$\\
     & $\times$ & $\times$ & $\checkmark$ & $8.88\pm0.12$ & $11.34$\\
    \cmidrule{2-6}
     & $\times$ & $\times$ & $\checkmark$ & $8.88\pm0.12$ & $11.34$\\
     & $\times$ & $\checkmark$ & $\times$ & $7.06\pm0.06$ & $43.99$\\
     & $\times$ & $\checkmark$ & $\checkmark$ & $8.43\pm0.11$ & $13.91$\\
    \midrule
    \multirow{5}*{alphaGAN$_{(s)}$} & $\checkmark$ & $\times$ & $\checkmark$ & $8.70\pm0.11$ & $15.56$\\
     & $\times$ & $\times$ & $\checkmark$ & $8.98\pm0.09$ & $10.35$\\
    \cmidrule{2-6}
     & $\times$ & $\times$ & $\checkmark$ & $8.98\pm0.09$ & $10.35$\\
     & $\times$ & $\checkmark$ & $\times$ & $8.45\pm0.09$ & $15.47$\\
     & $\times$ & $\checkmark$ & $\checkmark$ & $8.18\pm0.11$ & $18.85$\\
    \bottomrule
  \end{tabular}
  }
\vspace{-0.3cm}
\end{table}

\textbf{Search D's architecture or not? }\label{search D?}
A problem may arise from alphaGAN: If searching discriminator structures can facilitate the searching and training of generators? The results in Tab. \ref{Ablation study summary} show that searching the discriminator cannot help the search of the optimal generator. We also conduct the trial by training GANs with the obtained architectures by searching G and D, while the final performance is inferior to the setting of retraining with a given discriminator configuration. Simultaneously searching architectures of both G and D potentially increases the effect of inferior discriminators which may hamper the search of optimal generators conditioned on strong discriminators. In this regard, solely learning generators' architectures may be a better choice.

\textbf{How to obtain $\overline{G}$? }
In the definition of duality gap, $\overline{G}$ and $\overline{D}$ denote the optima of G and D, respectively. As both of the architecture and network parameters are variables for $\overline{G}$, we do the experiments of investigating the effect of updating $\omega_{G}$ and $\alpha_{G}$ for attaining $\overline{G}$. The results in Table \ref{Ablation study summary} show that updating $\omega_{G}$ solely achieves the best performance. Approximating $\overline{G}$ with $\omega_{G}$ update solely means that the architectures of G and $\overline{G}$ are identical, and hence optimizing architecture parameters $\alpha_{G}$ in \eqref{alphaGAN-formulation-a} can be viewed as the compensation for the gap brought by the weight parameters of $\omega_{G}$ and $\omega_{\overline{G}}$.

\subsection{Analysis}\label{analysis}
We have seen that alphaGAN can find high-performing architectures, and would like to provide a further analysis of the proposed algorithm in this section. 

\begin{figure}
  \centering
  \subfigure[\texttt{IS in search}]{\includegraphics[width=4.2cm]{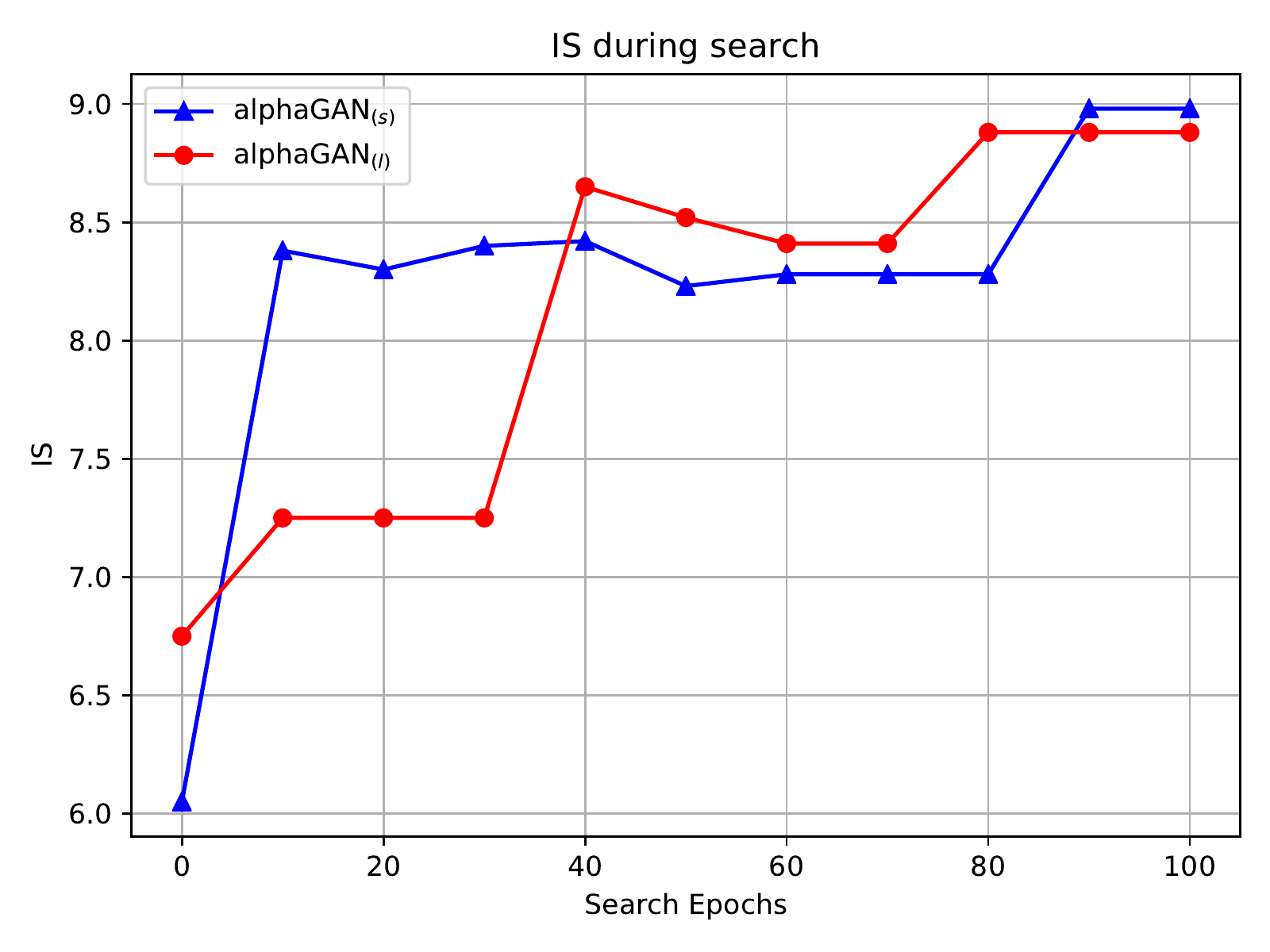}}
  \subfigure[\texttt{FID in search}]{\includegraphics[width=4.2cm]{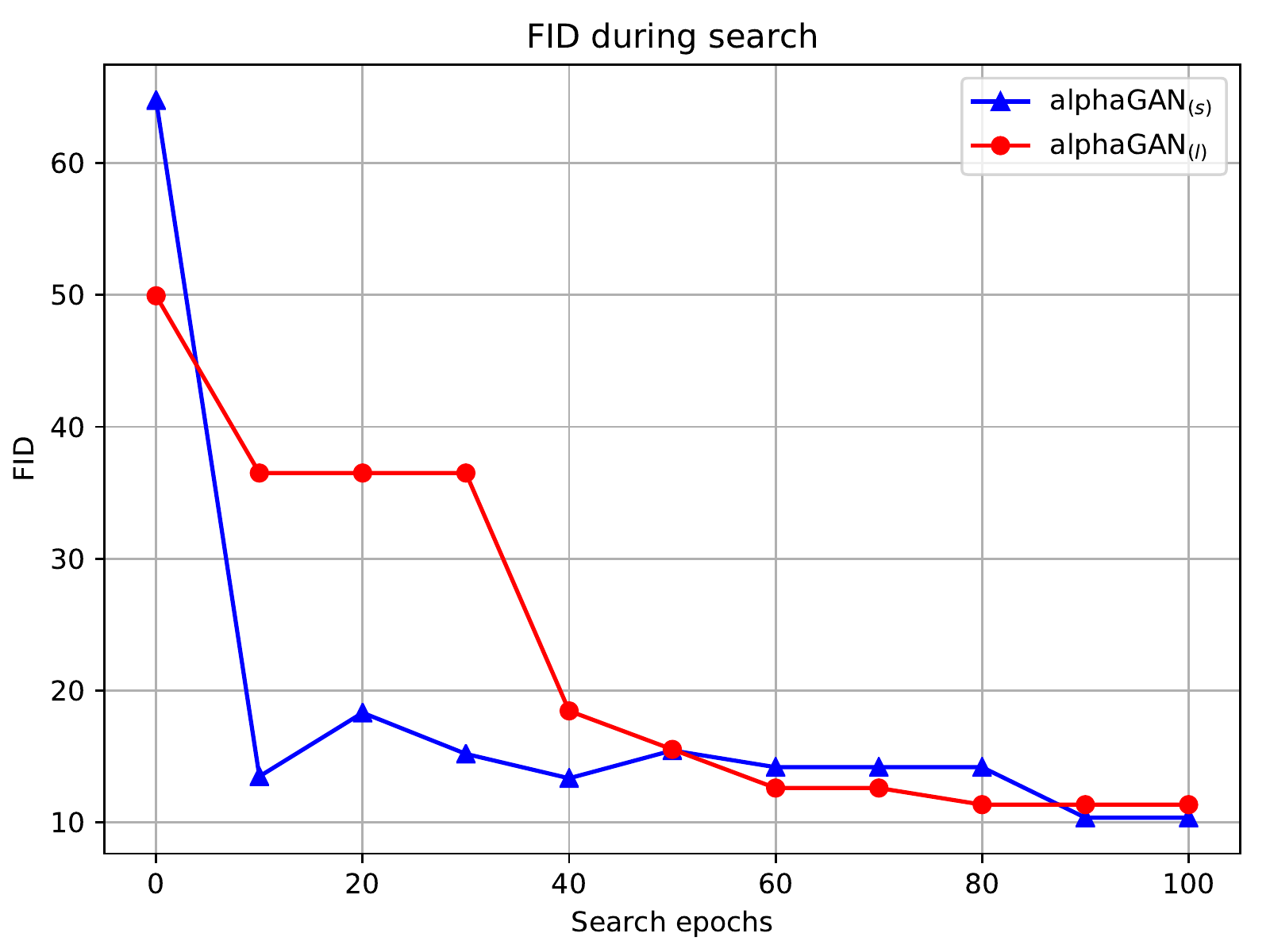}}
  \vspace{-0.3cm}
  \caption{Tracking architectures during searching. alphaGAN$_{(s)}$ is denoted by blue color with plus marker and alphaGAN$_{(l)}$ is denoted by red color with triangle marker.}
  \vspace{-0.4cm}
  \label{performance during search}
\end{figure}

\begin{figure}
  \centering
  \subfigure[\texttt{alphaGAN$_{(l)}$}]{\includegraphics[width=4.2cm]{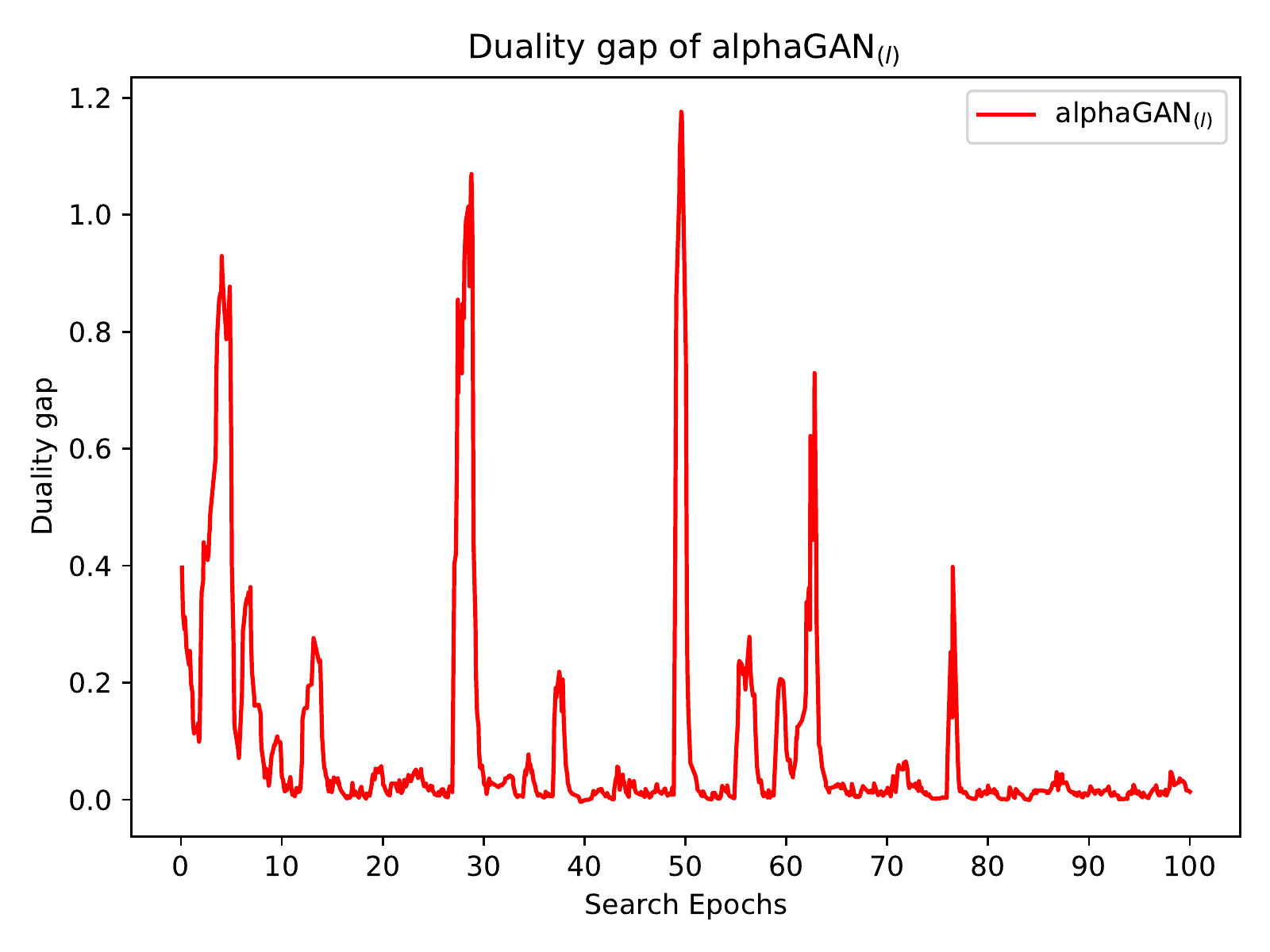}}
  \subfigure[\texttt{alphaGAN$_{(s)}$}]{\includegraphics[width=4.2cm]{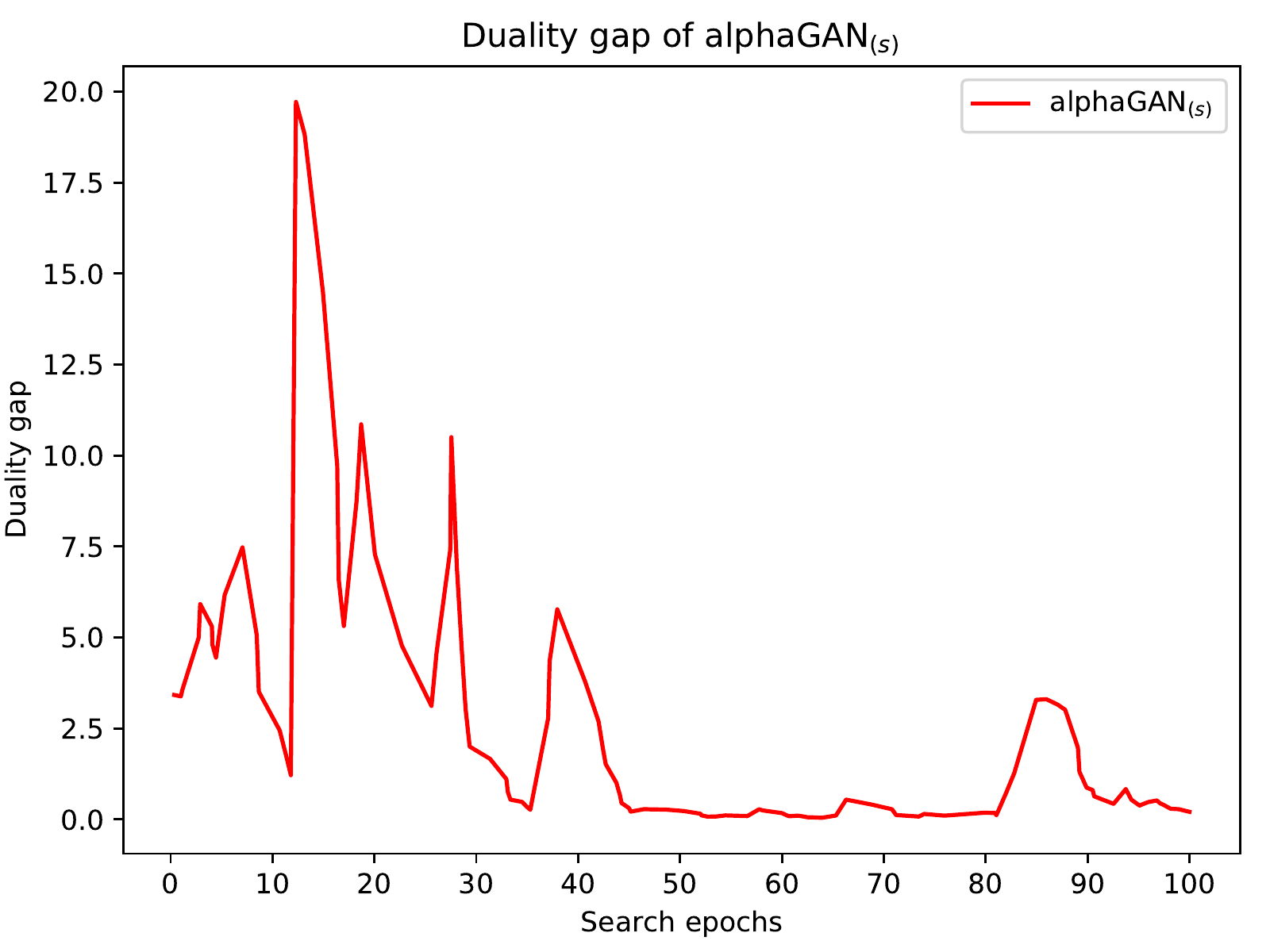}}
  \vspace{-0.3cm}
  \caption{Duality gap of alphaGAN$_{(l)}$ and alphaGAN$_{(s)}$ during searching.}
  \vspace{-0.4cm}
  \label{Duality gap curve of conventional GANs}
\end{figure}

\subsubsection{Architectures on Searching}
To understand the search process of alphaGAN, we track the intermediate architectures of alphaGAN$_{(s)}$ and alphaGAN$_{(l)}$ during searching, and re-train them on CIFAR-10 (in Fig. \ref{performance during search}). We observe a clear trend that the architectures are learned towards high performance during searching though slight oscillation may happen. Especially, alphaGAN$_{(l)}$ realizes gradual improvement in performance during the process, while alphaGAN$_{(s)}$ displays a faster convergence on the early stage of the process and can achieve comparable results, indicating that solving the inner-level optimization problem by virtue of rough approximations (as using more steps can always achieve a closer approximation to the optimum) can significantly benefit the efficiency of solving the bi-level problem without a sacrifice in accuracy.

We also plot the curve of duality gap during search, shown in Fig. \ref{Duality gap curve of conventional GANs}. Both curves show the descent tendency of duality gap. Given the performance trajectory in Fig. \ref{performance during search}, the search of alphaGAN discovers better architectures with the descent of duality gap, verifying our motivation to search the architecture of the generator towards Nash Equilibrium under the metric of duality gap. Observing the duality gap curve, we can find that the curve of alphaGAN$_{(l)}$ possesses more spikes, compared with alphaGAN$_{(s)}$, because ``arch\_part" size $S$ is much smaller than ``weight\_part" size $T$ and ``test\_weight\_part" size $R$ (i.e., $20$ vs $390$).

\begin{figure}[t]
  \centering
  \includegraphics[width=0.48\textwidth]{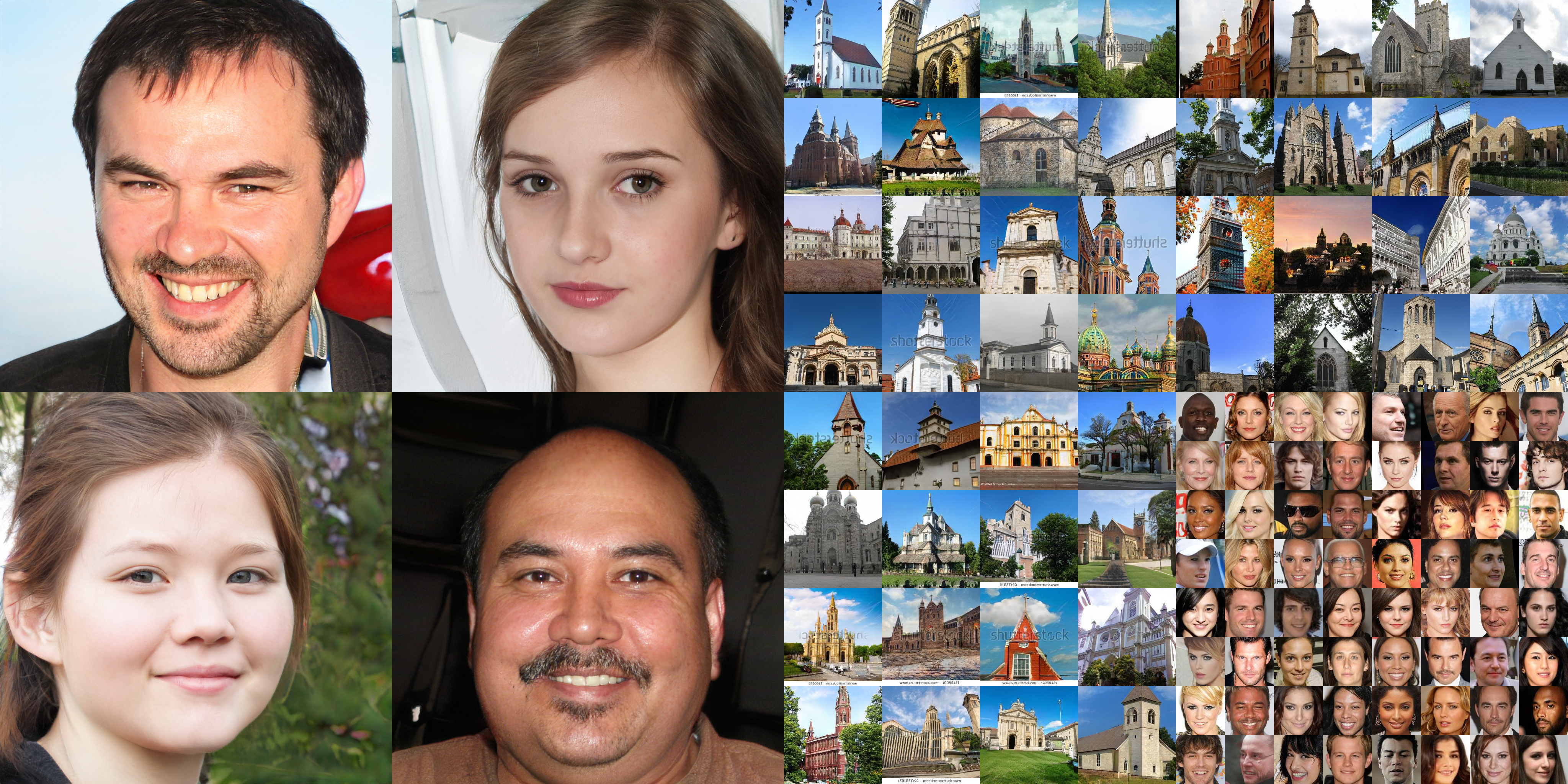}
  \vspace{-0.5cm}
  \caption{The generated images of the models with best FID on FFHQ, LSUN-church, and CelebA. We select the images with cherry picking.}
  \label{generated images of StyleGAN2}
  \vspace{-0.5cm}
\end{figure}

\section{Search the architecture of StyleGAN2}\label{Extension}
 
To validate the scalability and the versatility of the framework of alphaGAN, we exploit alphaGAN to search the architectures of the generator in state of the art StyleGAN2, according to the searching configurations drawn from the studies on CIFAR-10 in Sec. \ref{Conventional GANs}, e.g., exploiting the discriminator with the fixed architecture, and updating the weight parameters $\omega_{G}$ and $\omega_{D}$ while approximating $\overline{G}$ and $\overline{D}$. We want to emphasize that it is non-trivial to automatically search the architectures of GANs with a much deeper and wider network, which possesses the unique structure (e.g., the synthesis network and the mapping network of StyleGAN2). 

The difficulty comes from three main aspects. First, searching the architecture of heavy-weight StyleGAN2 on large datasets requires an efficient optimization process, whereas alphaGAN$_{(s)}$ can efficiently obtain the architecture via smaller step sizes, i.e., $T$, $S$, $R$. Second, designing the search space customized for StyleGAN2 needs to take the intrinsic property into account. We design a search space for StyleGAN2, in which not only operations but also the intermediate latent $\mathcal{W}_i$ fed into the convolutional layers is included, elaborated in Sec. \ref{Search space of StyleGAN2}. Third, considering the excellent performance of StyleGAN2, further promoting the performance via optimizing the architecture is non-trivial.

We inherit the structure of the discriminator (e.g., 'Residual' structure) in StyleGAN2 during searching and re-training. Moreover, tricks employed in StyleGAN2 are also adopted when re-training the obtained architecture of the generator, such as style mixing regularization and latent truncation. We totally conduct experiments on three datasets, i.e., FFHQ \cite{karras2019style} (1024x1024), LSUN-church \cite{yu15lsun} (256x256), and CelebA \cite{liu2015faceattributes} (128x128). Experiment details can be found in the supplementary material.


\subsection{Macro/Micro Search Space}\label{macro/micro}
As illustrated in the previous NAS works \cite{pham2018efficient}, the search space of NAS methods can be divided into two types, macro search space and micro search space. Macro search space denotes that the entire network possesses an unique architecture, and micro search space denotes that the entire network is comprised of several cells with the identical architecture. Compared with macro search space, micro search space allows for the architectures with less complexity but better transferability (i.e., not confined to the number of down-sampling operations or up-sampling operations). 

Previous NAS-GAN works \cite{gong2019autogan, wang2019agan, gao2019adversarialnas, tian2020off}, including alphaGAN for conventional GANs, are conducted under macro search space. Regarding StyleGAN2, to explore the impact of macro search space and micro search space on the performance of the obtained architectures, we experiment with both, elaborated in Sec. \ref{results of alphaGAN on StyleGAN2} and Tab. \ref{Scalability}.

\begin{figure}[h]
  \centering
  \subfigure[\texttt{$\mathcal{W}_{8}$}]{\includegraphics[width=2.1cm]{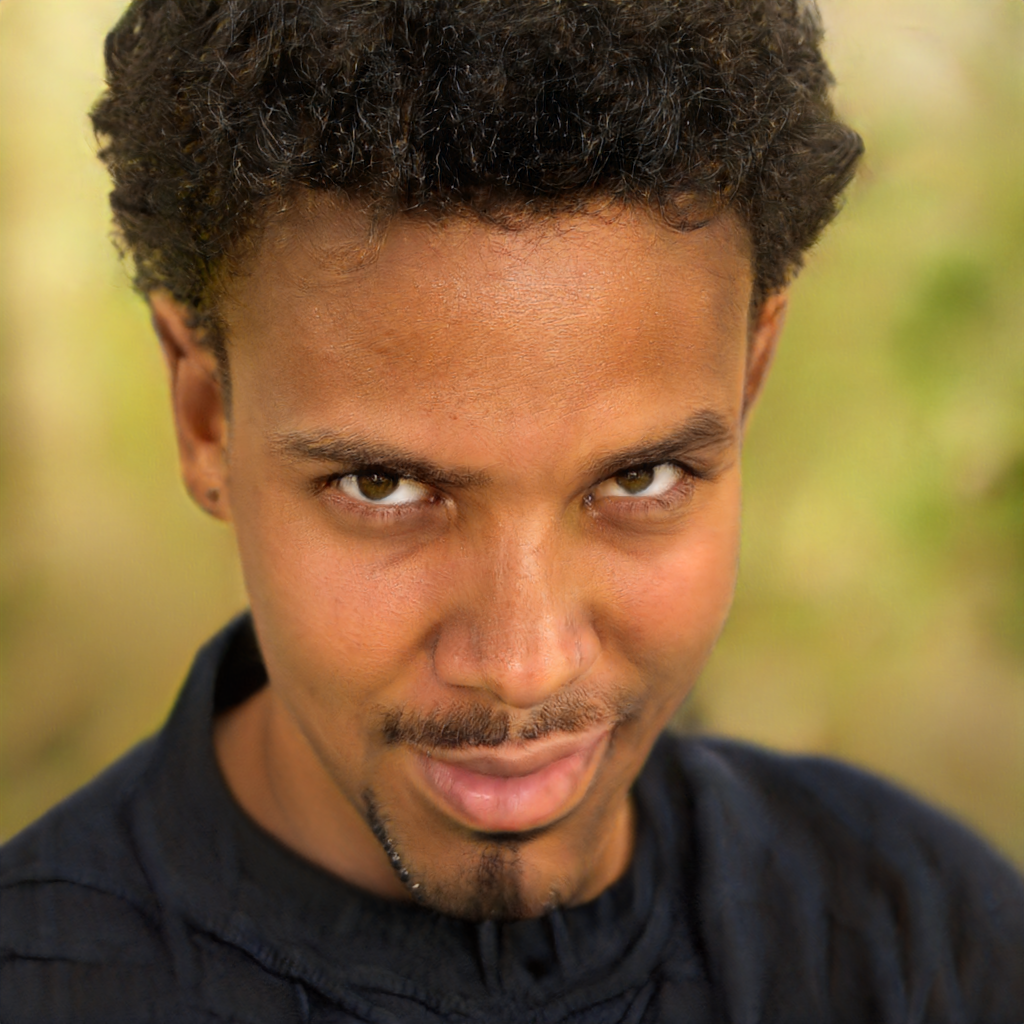}}
  \subfigure[\texttt{$\mathcal{W}_{7}$}]{\includegraphics[width=2.1cm]{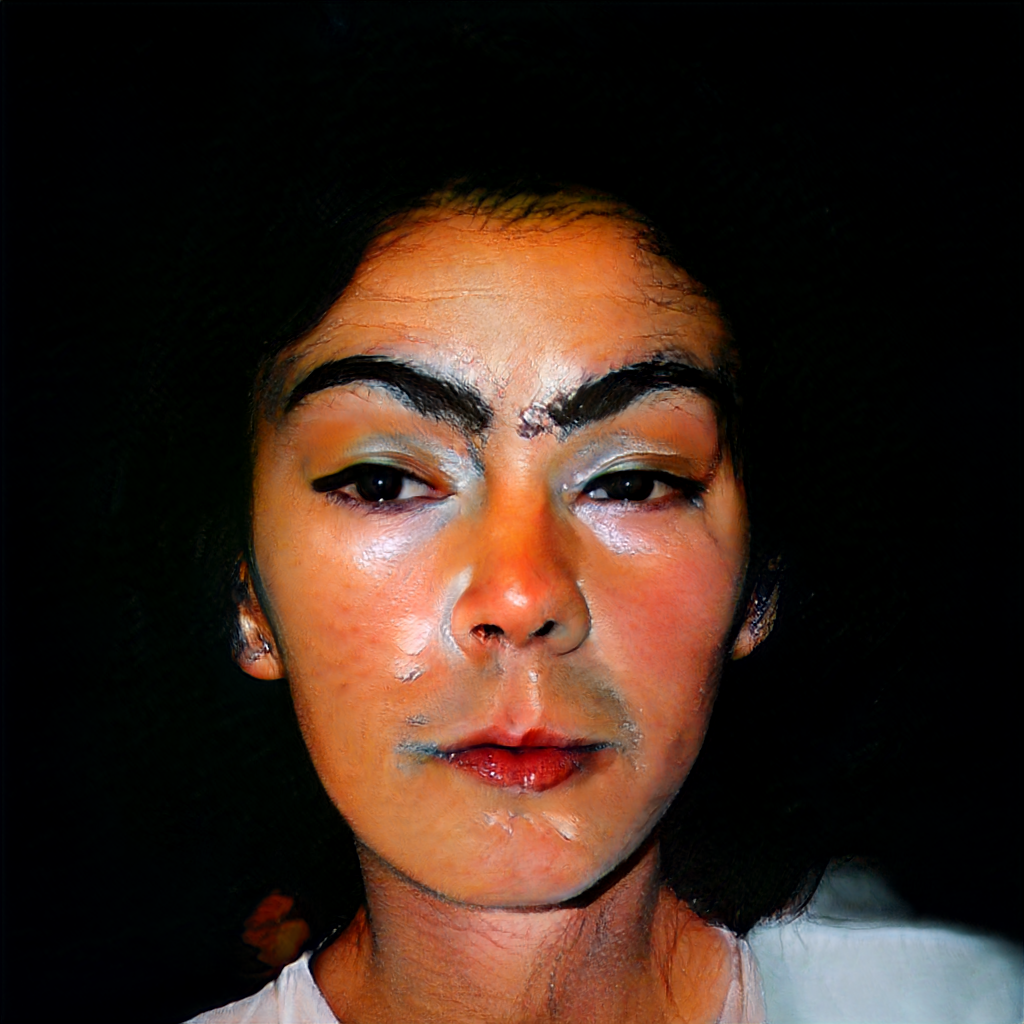}}
  \subfigure[\texttt{$\mathcal{W}_{6}$}]{\includegraphics[width=2.1cm]{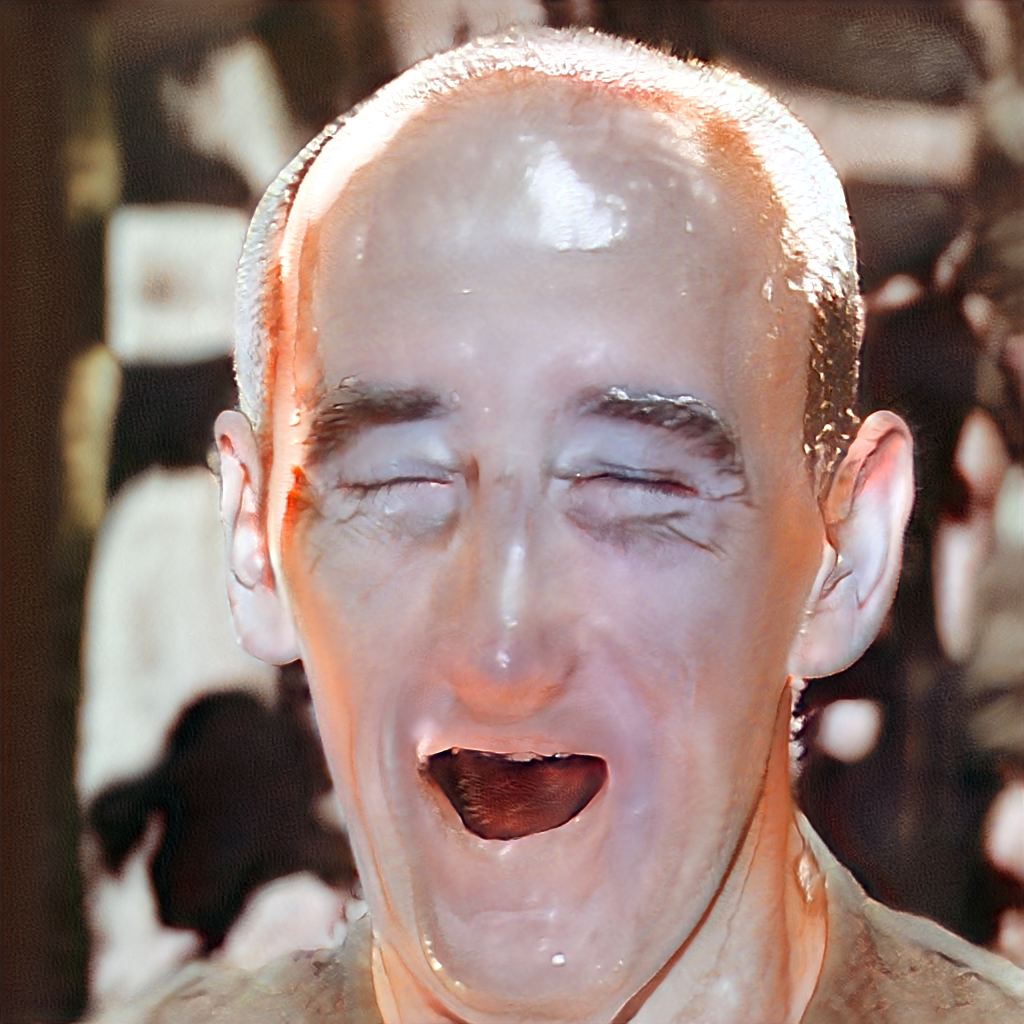}}
  \subfigure[\texttt{$\mathcal{W}_{5}$}]{\includegraphics[width=2.1cm]{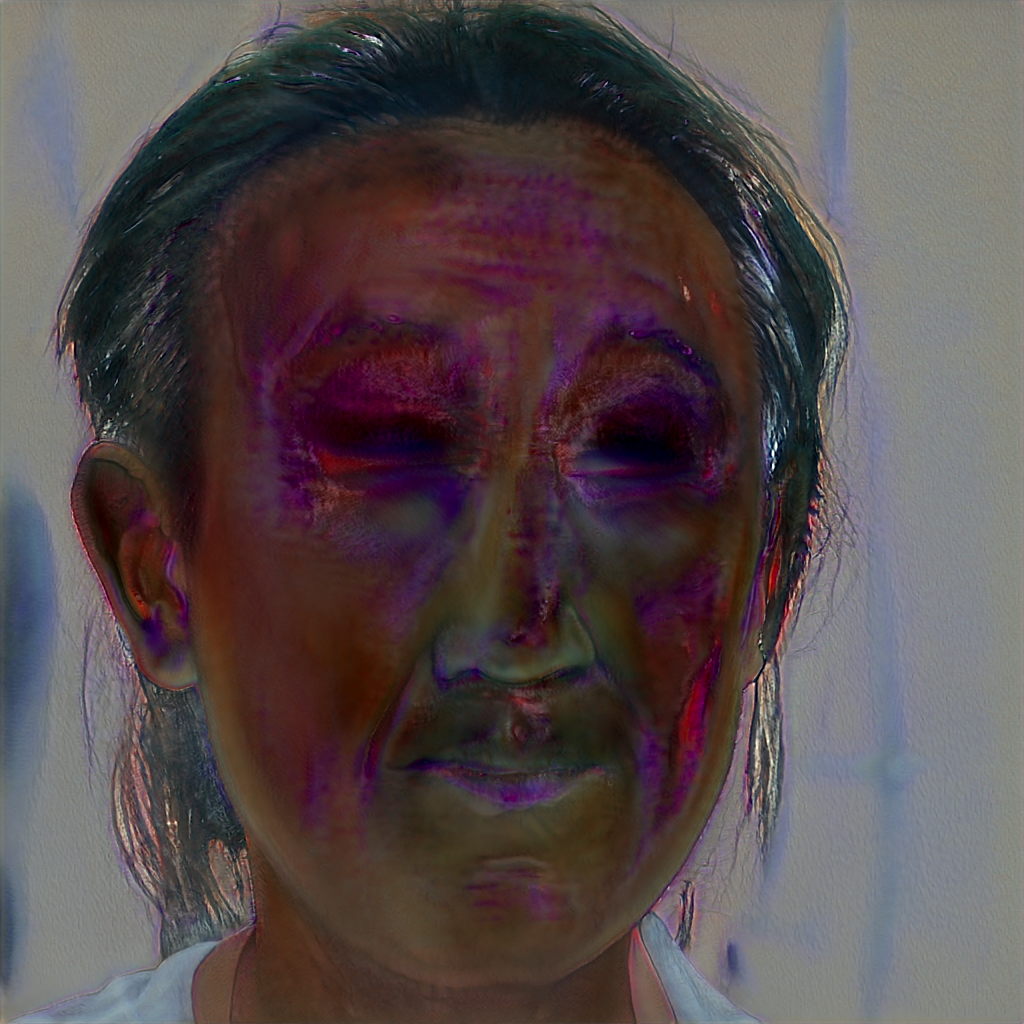}}
  \subfigure[\texttt{$\mathcal{W}_{4}$}]{\includegraphics[width=2.1cm]{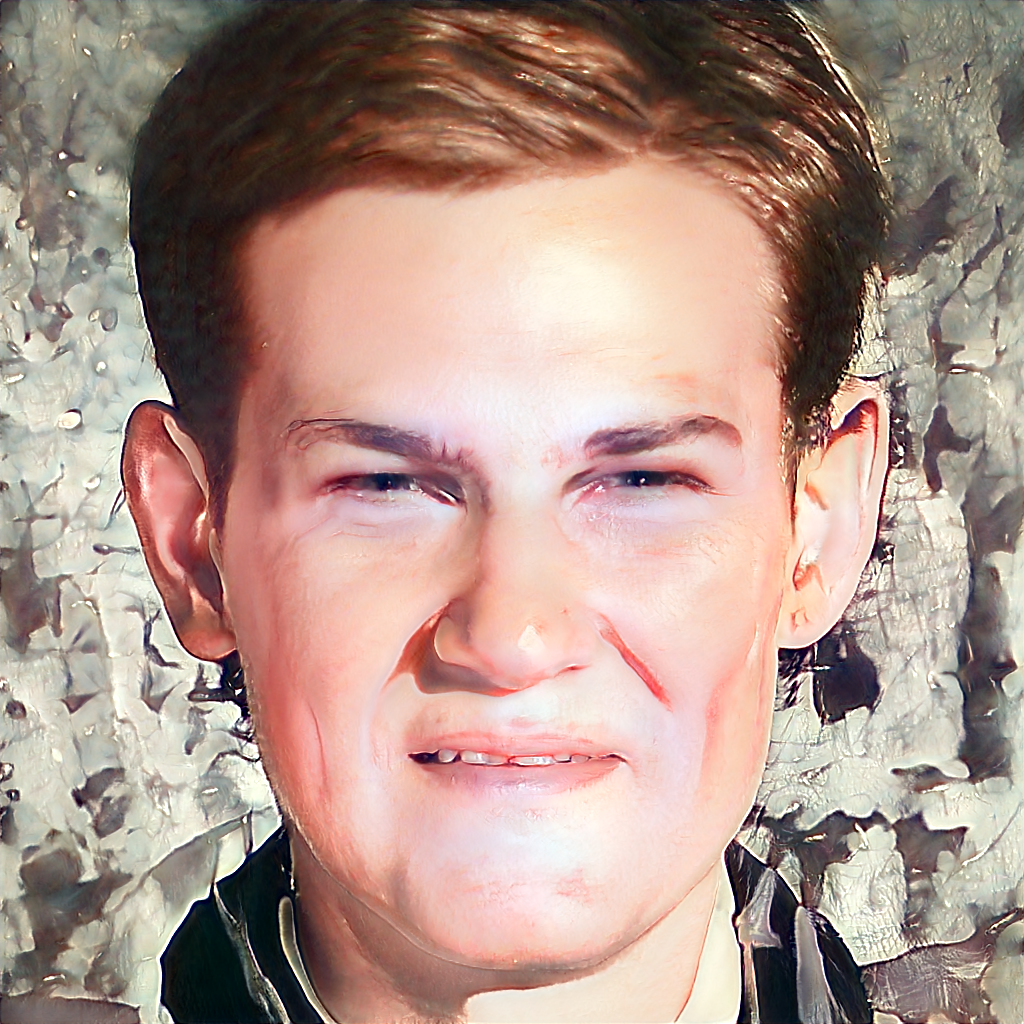}}
  \subfigure[\texttt{$\mathcal{W}_{3}$}]{\includegraphics[width=2.1cm]{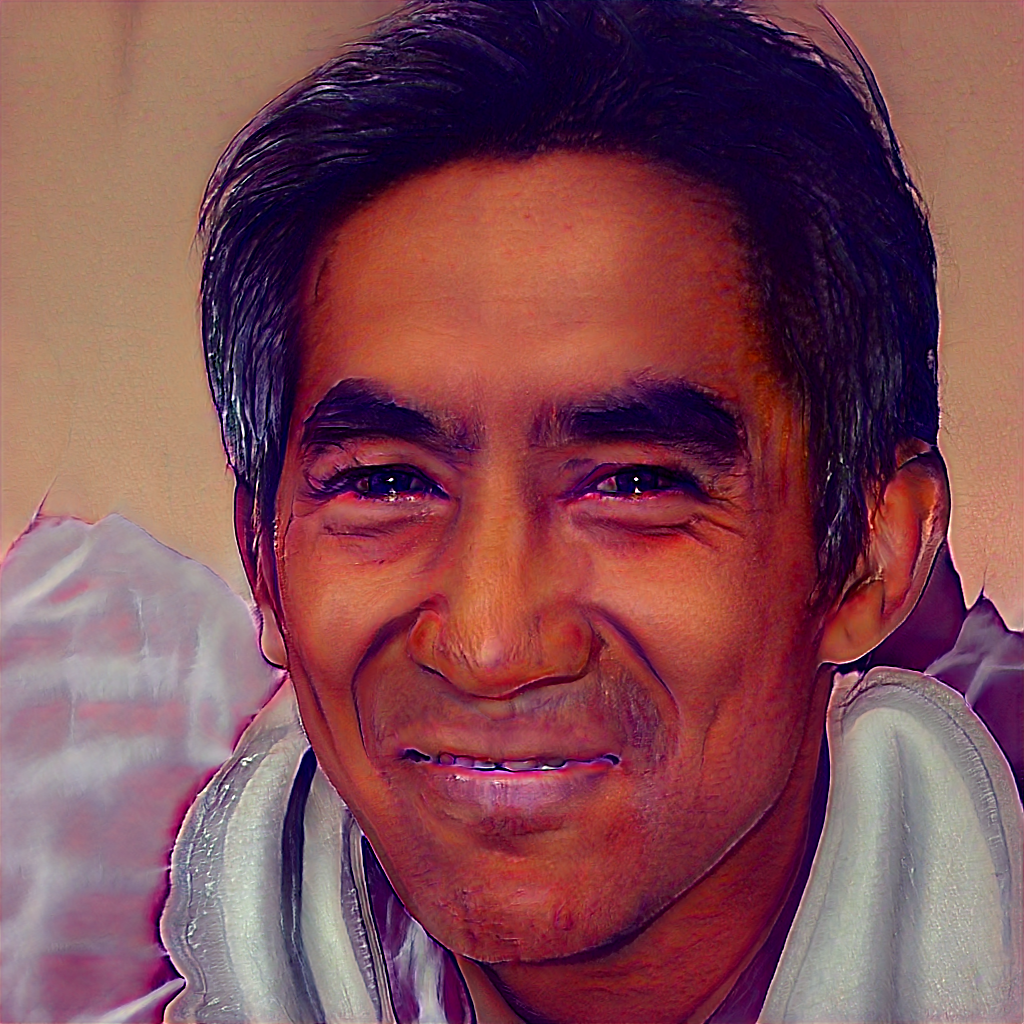}}
  \subfigure[\texttt{$\mathcal{W}_{2}$}]{\includegraphics[width=2.1cm]{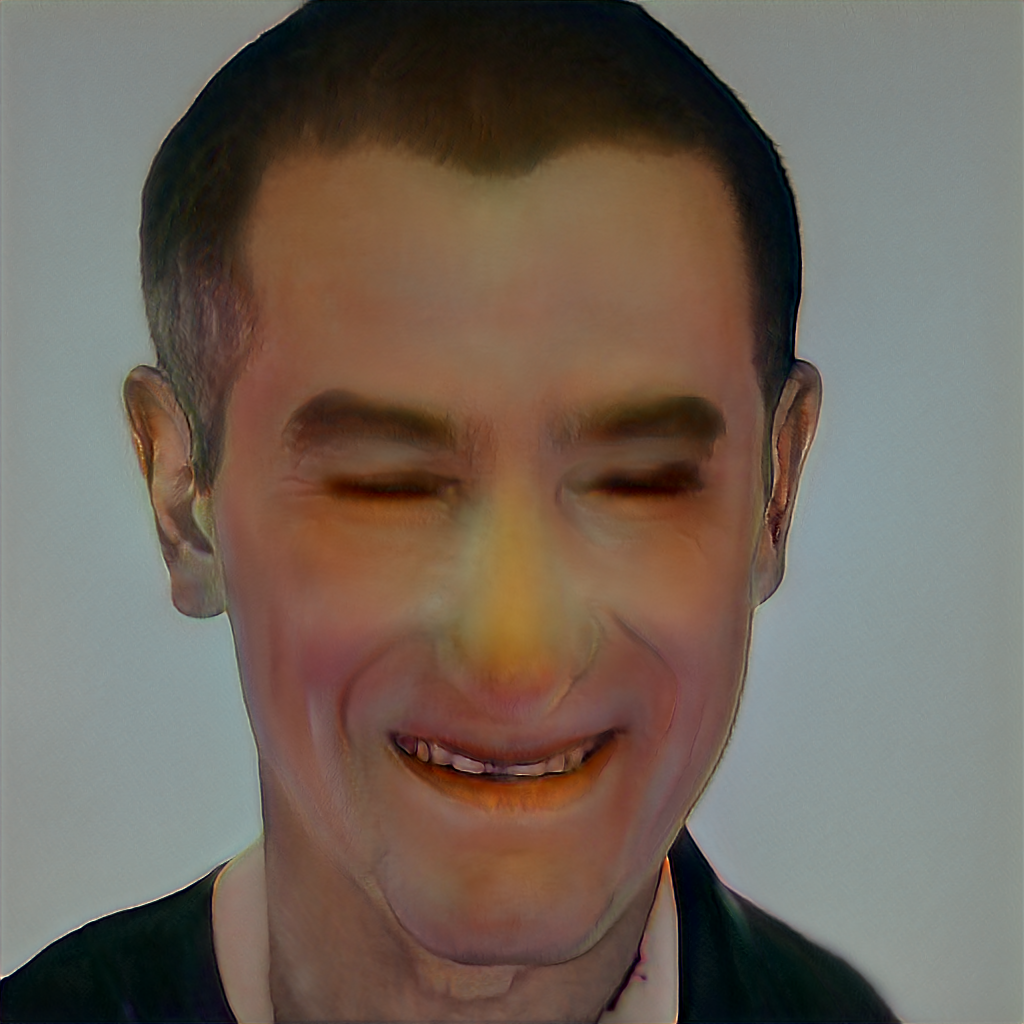}}
  \subfigure[\texttt{$\mathcal{W}_{1}$}]{\includegraphics[width=2.1cm]{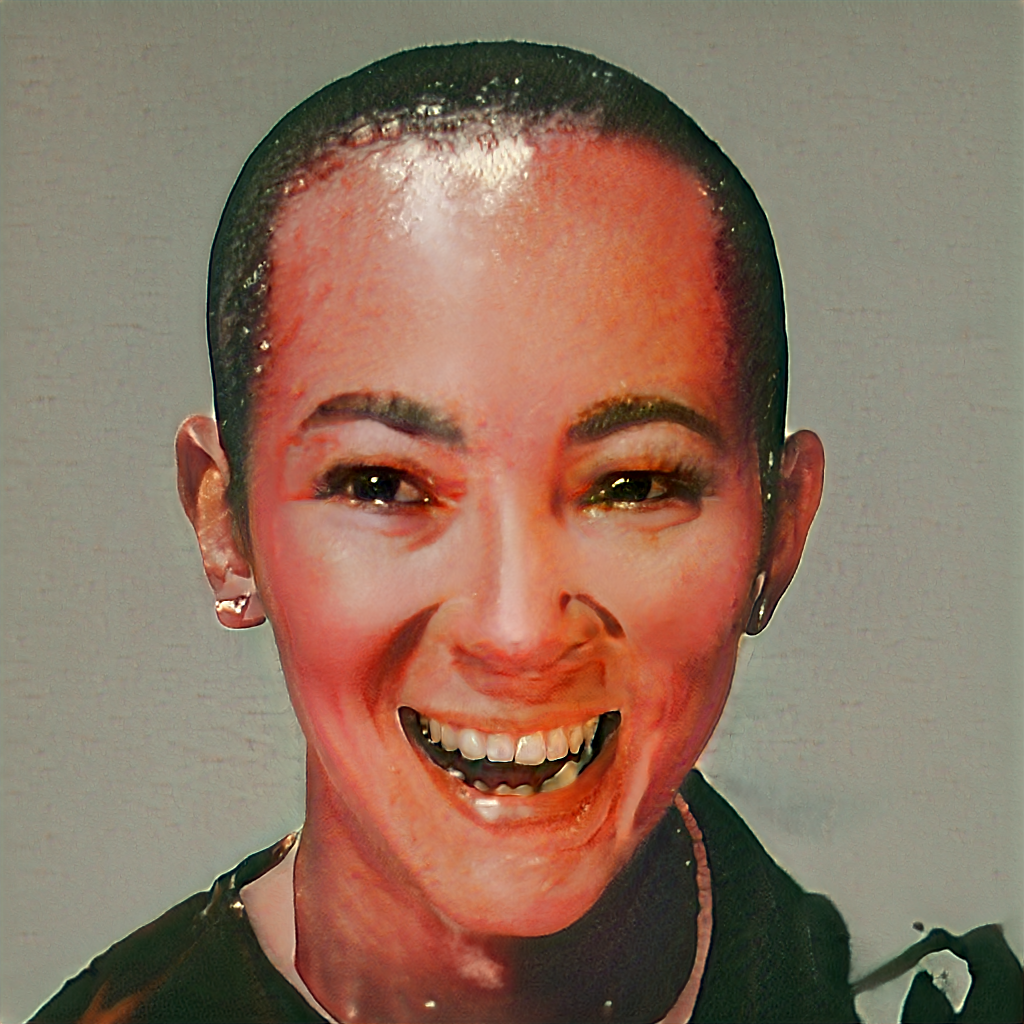}}
  \vspace{-0.3cm}
  \caption{The generated images with different intermediate latent $\mathcal{W}_{i=1,2,3,4,5,6,7,8}$ under the same latent z\_const and z\_var. $\mathcal{W}_{8}$ is the output of the last fully-connected layer in the mapping network. We exploit the pretrained model (trained with 25000K images) released by the authors of StyleGAN2 \cite{karras2020analyzing}.}
  \label{mapping different depths of w.}
  \vspace{-0.5cm}
\end{figure}

\begin{table*}[t]
  \caption{The results of exploiting alphaGAN to search the architecture of StyleGAN2. $\dagger$ denotes the reproduced results based on the official released code.}
  \label{Scalability}
  \vspace{-0.3cm}
  \centering
  \begin{tabular}{c|ccc|ccc|c}
    \toprule
    \multirow{2}*{Paradigm} & \multicolumn{3}{c|}{Search} & \multicolumn{3}{c|}{Re-train} & \multirow{2}*{FID} \\
    \cline{2-7}
    & Dataset & \tabincell{c}{Cost\\ (GPU-hours)} & Search space & Dataset & kimg & \tabincell{c}{Model size\\ (M)} \\
    \midrule
    \midrule
    COCO-GAN(\cite{lin2019coco}) & - & - & - & CelebA & - & - & $5.74$ \\
    MSG-StyleGAN(\cite{karnewar2020msg}) & - & - & - & LSUN-church & $24000$ & - & $5.2$ \\
    \hline
    \multirow{2}*{StyleGAN2(\cite{karras2020analyzing})} & - & - & - & church & $48000$ & $30.03$ & $3.86$ \\
     & - & - & - & FFHQ & $25000$ & $30.37$ & $2.84$ \\
    \hline
    \multirow{3}*{StyleGAN2$\dagger$} & - & - & - & CelebA & $15000$ & $29.33$ & $2.17$ \\
     & - & - & - & church & $15000$ & $30.03$ & $3.35$ \\
     & - & - & - & FFHQ & $20000$ & $30.37$ & $3.50$ \\
    \midrule
    \midrule
    \multirow{4}*{alphaGAN$_{(s)}$ + micro} & CelebA & $1.5$ & $9$ & CelebA & $15000$ & $29.33$ & $2.07$ \\
     & church & $2$ & $9$ & church & $15000$ & $52.32$ & $3.03$ \\
     & FFHQ & $6.5$ & $9$ & FFHQ & $20000$ & $52.73$ & $2.77$ \\
     & CelebA & $1.5$ & $9$ & FFHQ & $20000$ & $30.37$ & $2.84$ \\
    \hline
    \multirow{3}*{alphaGAN$_{(s)}$ + micro + dynamic $\left\{\mathcal{W}_{i}\right\}$} & CelebA & $2.5$ & $\sim6.2\times10^{11}$ & CelebA & $15000$ & $29.33$ & $\textbf{1.94}$ \\
     & church & $3$ & $\sim4.0\times10^{13}$ & church & $15000$ & $52.32$ & $\textbf{2.86}$ \\
     & FFHQ & $12$ & $\sim1.6\times10^{17}$ & FFHQ & $20000$ & $52.73$ & $\textbf{2.75}$ \\
    \hline
    \multirow{3}*{alphaGAN$_{(s)}$ + macro} & CelebA & $1.5$ & $\sim1.8\times10^5$ & CelebA & $15000$ & $41.39$ & $2.12$ \\
     & church & $1.5$ & $\sim1.6\times10^6$ & church & $15000$ & $41.44$ & $3.05$ \\
     & FFHQ & $6$ & $\sim1.3\times10^9$ & FFHQ & $20000$ & $31.68$ & $2.95$ \\
    \hline
    \multirow{3}*{alphaGAN$_{(s)}$ + macro + dynamic $\left\{\mathcal{W}_{i}\right\}$} & CelebA & $2$ & $\sim1.2\times10^{16}$ & CelebA & $15000$ & $35.10$ & $1.99$ \\
     & church & $3$ & $\sim7.0\times10^{18}$ & church & $15000$ & $35.67$ & $3.65$ \\
     & FFHQ & $12$ & $\sim2.3\times10^{24}$ & FFHQ & $20000$ & $30.10$ & $3.32$ \\
    \bottomrule
  \end{tabular}
\end{table*}

\begin{figure*}[t]
  \vspace{-0.5cm}
  \centering
  \subfigure[\texttt{CelebA}]{\includegraphics[width=0.3\textwidth]{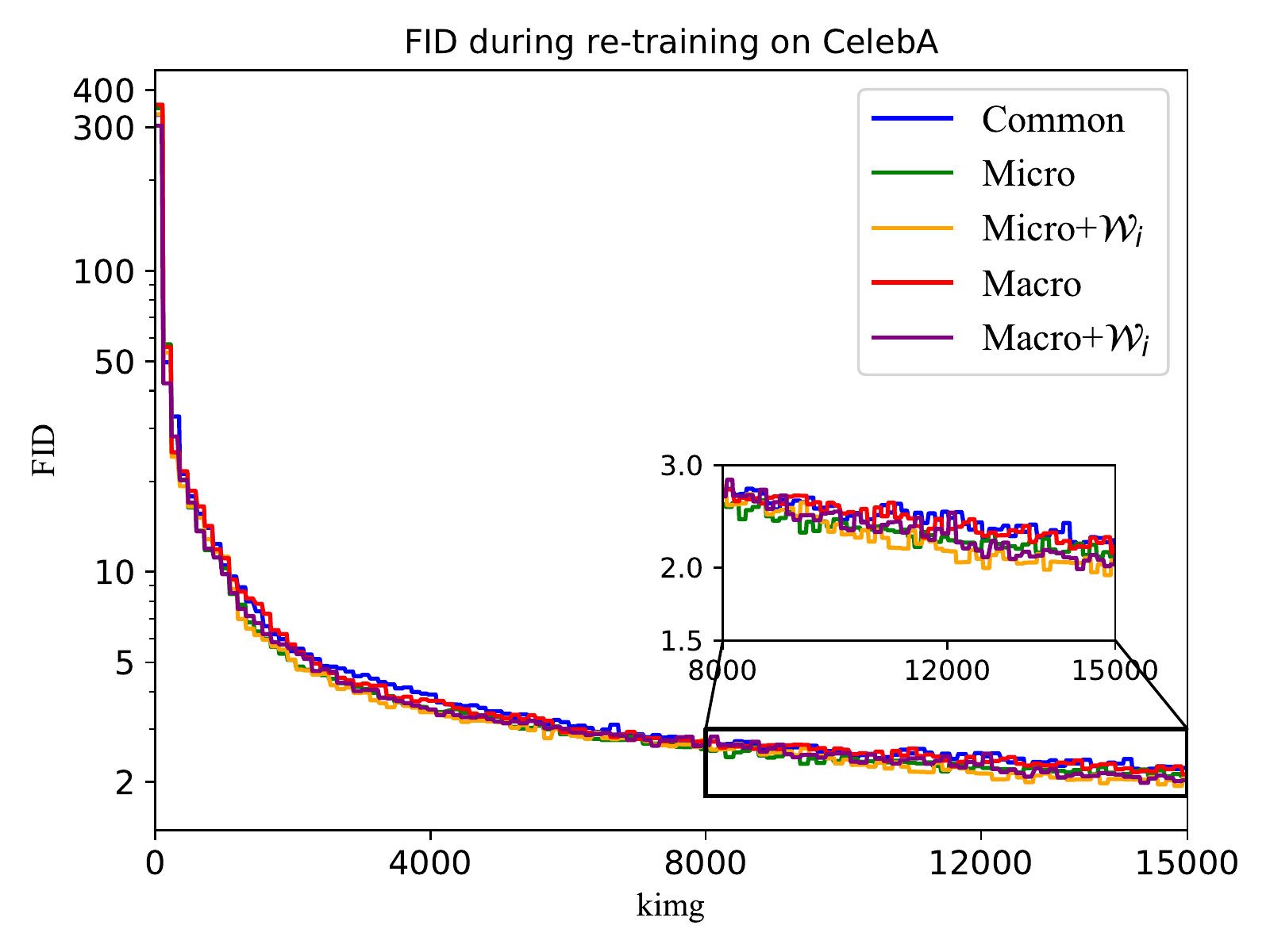}}
  \subfigure[\texttt{LSUN-church}]{\includegraphics[width=0.3\textwidth]{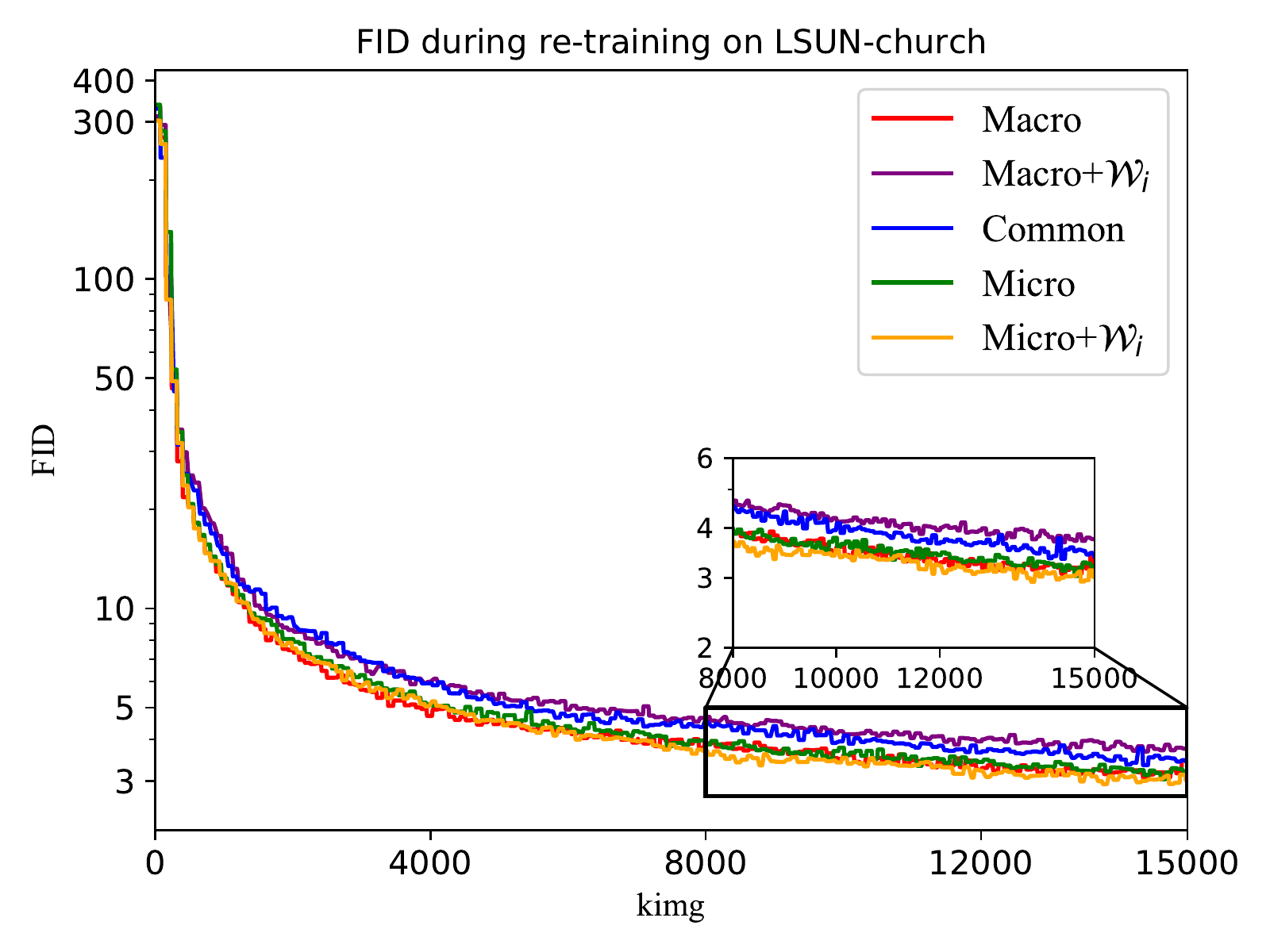}}
  \subfigure[\texttt{FFHQ}]{\includegraphics[width=0.3\textwidth]{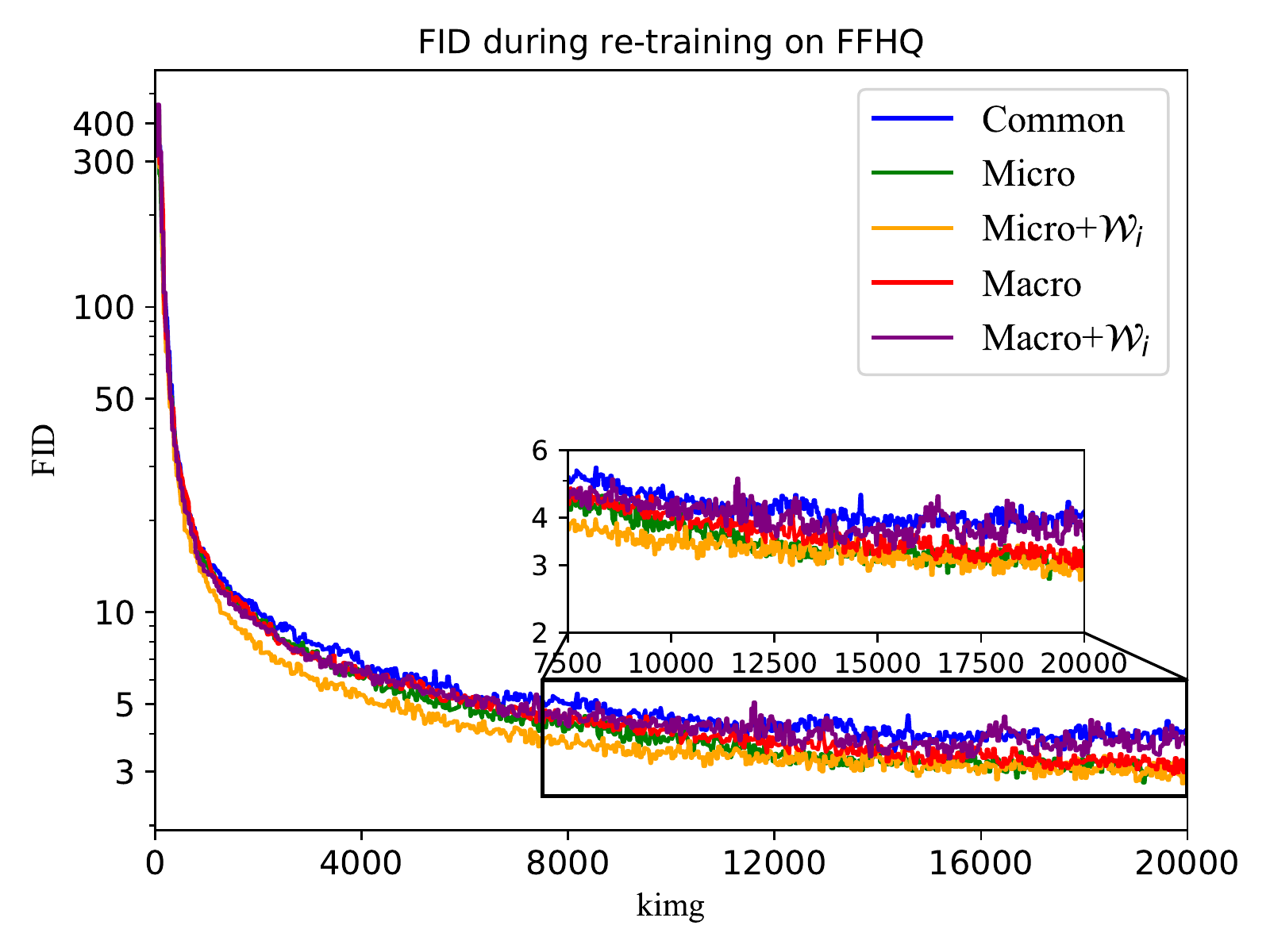}}
  \vspace{-0.4cm}
  \caption{FID curve of the searched architecture and the original architecture during re-training. We present the curves of re-training. ``Common architecture" denotes the original architecture exploited by StyleGAN2.}
  \vspace{-0.4cm}
  \label{Re-training curve}
\end{figure*}

\subsection{Details of the Search Space}\label{Search space of StyleGAN2}
\textbf{The pool of operations.} As mentioned in Sec. \ref{alphaGAN-optimization-subsection}, the pool of operations is modified while searching the architecture of StyleGAN2. The pool of normal operations $\mathcal{O}_{ns}$ is comprised of \{conv\_1x1, conv\_3x3, conv\_5x5\}, in which depth-wise separable convolution operations are removed because StyleGAN2 directly modulates the convolution kernels and depth-wise separable convolution indeed contains two convolution operations, i.e., depth-wise convolution and point-wise convolution. It is uncertain to modulate the depth-wise convolution or the point-wise convolution. Thus for simplicity, we remove the depth-wise separable convolution operations from $\mathcal{O}_{nc}$. 

As for up-sampling operations, instead of ``nearest" and ``bilinear", the pool of up-sampling operations $\mathcal{O}_{us}$ employs ``nearest + conv\_3x3" and ``bilinear + conv\_3x3". We upgrade common interpolation operations for two reasons. First, with conv\_3x3 followed, the style information can be added into ``learnable interpolation" operations like ``deconv", thus suitable for StyleGAN2. Second, with a completely equal number of learnable parameters, ``learnable interpolation" operations can be comparable with ``deconv", thus treated equally during searching. In the following, the two operations are denoted as ``nearest\_conv" and ``bilinear\_conv".

\textbf{Dynamic $\left\{\mathcal{W}_{i}\right\}_{i=1,...,n}$.} In the original implementation of StyleGAN2, all the convolutional layers $\left\{L_{j}\right\}_{j=1,...,m}$ and the skip connection layers $\left\{{\rm tRGB\_k}\right\}_{k=2,...,m+1}$ in the synthesis network receive the intermediate latent $\mathcal{W}_{n}$ of the last fully-connected layer in the mapping network as the source of style information, in which the other internal latent $\left\{\mathcal{W}_{i}\right\}_{i=1,...,n-1}$ does not correspond to meaningful style information. We try to feed the intermediate latent $\left\{\mathcal{W}_{i}\right\}_{i=1,...,n-1}$ except for $\mathcal{W}_{n}$ into the layers of the synthesis network, as shown in Fig. \ref{mapping different depths of w.}. Under the identical z\_const and z\_var, we can find that compared with $\mathcal{W}_{n}$, the synthesis network cannot exploit the intermediate latent $\left\{\mathcal{W}_{i}\right\}_{i=1,...,n-1}$ as the style information to precisely fit the distribution of real images, which indicates that the transformation process of the mapping network is not smooth. 

Thus, instead of uniformly feeding $\mathcal{W}_{n}$, we exploit alphaGAN to automatically select the intermediate latent $\left\{\mathcal{W}_{i}\right\}_{i=1,...,n}$ for each convolutional layer $\left\{L_{j}\right\}_{j=1,...,m}$ and skip connection layer $\left\{{\rm tRGB\_k}\right\}_{k=2,...,m+1}$ in the synthesis network, as shown in Fig. \ref{The depiction of StyleGAN2}, named ``dynamic $\left\{\mathcal{W}_{i}\right\}$". Our key motivation to search $\left\{\mathcal{W}_{i}\right\}_{i=1,...,n}$ is to enable each convolutional layer $\left\{L_{j}\right\}_{j=1,...,m}$ to select the most suitable intermediate latent $\mathcal{W}_{i}$ and make the process of mapping the initial latent z\_var to $\mathcal{W}_{n}$ smooth and meaningful. ``Dynamic $\left\{\mathcal{W}_{i}\right\}$" can also be viewed as a regularization method, which regularizes the mapping network to make the mapping process from z\_var to $\mathcal{W}_{n}$ smooth and meaningful.

\begin{table*}[t!]
  \caption{The architectures obtained via alphaGAN$_{(s)}$ with the ``micro + dynamic $\left\{\mathcal{W}_{i}\right\}$" paradigm. $*$ denotes that the latent $\mathcal{W}_{i}$ searched for the layer ``tRGB\_(m+1)" in Fig. \ref{The depiction of StyleGAN2}. $\mathcal{L}_{i}$ in the bracket is the according index of the layer in Fig. \ref{The depiction of StyleGAN2}.}
  \label{Searched_arch_style}
  \vspace{-0.3cm}
  \centering
  \resizebox{\textwidth}{!}{
  \begin{tabular}{c|c|ccccccccc}
    \toprule
    \multirow{2}*{Dataset} & \multirow{2}*{Searched operation/$\mathcal{W}_{i}$} & \multicolumn{9}{c}{Input resolution} \\
    \cline{3-11}
     &  & 4x4 & 8x8 & 16x16 & 32x32 & 64x64 & 128x128 & 256x256 & 512x512 & 1024x1024 \\
    \midrule
    \midrule
    \multirow{4}*{CelebA} & $\mathcal{W}_i$ for the normal operation & $\mathcal{W}_{2}$ ($\mathcal{L}_{1}$) & $\mathcal{W}_{4}$ ($\mathcal{L}_{3}$) & $\mathcal{W}_{1}$ ($\mathcal{L}_{5}$) & $\mathcal{W}_{1}$ ($\mathcal{L}_{7}$) & $\mathcal{W}_{1}$ ($\mathcal{L}_{9}$) & $\mathcal{W}_{8}$ ($\mathcal{L}_{11}$), ${\mathcal{W}_{5}}^{*}$ & - & - & - \\ 
     & Normal operation & \multicolumn{6}{c}{conv\_3x3} & - & - & - \\
     & $\mathcal{W}_i$ for the up-sampling operation & $\mathcal{W}_{3}$ ($\mathcal{L}_2$) & $\mathcal{W}_{1}$ ($\mathcal{L}_{4}$) & $\mathcal{W}_{4}$ ($\mathcal{L}_{6}$) & $\mathcal{W}_{6}$ ($\mathcal{L}_{8}$) & $\mathcal{W}_{3}$ ($\mathcal{L}_{10}$) & - & - & - & - \\
     & Up-sampling operation & \multicolumn{5}{c}{nearest\_conv} & - & - & - & - \\
    \hline
    \multirow{4}*{church} & $\mathcal{W}_i$ for the normal operation & $\mathcal{W}_{7}$ ($\mathcal{L}_{1}$) & $\mathcal{W}_{8}$ ($\mathcal{L}_{3}$) & $\mathcal{W}_{8}$ ($\mathcal{L}_{5}$) & $\mathcal{W}_{1}$ ($\mathcal{L}_{7}$) & $\mathcal{W}_{1}$ ($\mathcal{L}_{9}$) & $\mathcal{W}_{2}$ ($\mathcal{L}_{11}$) & $\mathcal{W}_{4}$ ($\mathcal{L}_{13}$), ${\mathcal{W}_{3}}^*$ & - & - \\ 
     & Normal operation & \multicolumn{7}{c}{conv\_5x5} & - & - \\
     & $\mathcal{W}_i$ for the up-sampling operation & $\mathcal{W}_{6}$ ($\mathcal{L}_{2}$) & $\mathcal{W}_{4}$ ($\mathcal{L}_{4}$) & $\mathcal{W}_{4}$ ($\mathcal{L}_{6}$) & $\mathcal{W}_{2}$ ($\mathcal{L}_{8}$) & $\mathcal{W}_{4}$ ($\mathcal{L}_{10}$) & $\mathcal{W}_{2}$ ($\mathcal{L}_{12}$) & - & - & - \\
     & Up-sampling operation & \multicolumn{6}{c}{bilinear\_conv} & - & - & - \\
    \hline
    \multirow{4}*{FFHQ} & $\mathcal{W}_i$ for the normal operation & $\mathcal{W}_{3}$ ($\mathcal{L}_{1}$) & $\mathcal{W}_{7}$ ($\mathcal{L}_{3}$) & $\mathcal{W}_{4}$ ($\mathcal{L}_{5}$) & $\mathcal{W}_{3}$ ($\mathcal{L}_{7}$) & $\mathcal{W}_{4}$ ($\mathcal{L}_{9}$) & $\mathcal{W}_{3}$ ($\mathcal{L}_{11}$) & $\mathcal{W}_{2}$ ($\mathcal{L}_{13}$) & $\mathcal{W}_{4}$ ($\mathcal{L}_{15}$) & $\mathcal{W}_{4}$ ($\mathcal{L}_{17}$), ${\mathcal{W}_{2}}^{*}$ \\  
     & Normal operation & \multicolumn{9}{c}{conv\_5x5} \\
     & $\mathcal{W}_i$ for the up-sampling operation & $\mathcal{W}_{3}$ ($\mathcal{L}_{2}$) & $\mathcal{W}_{6}$ ($\mathcal{L}_{4}$) & $\mathcal{W}_{8}$ ($\mathcal{L}_{6}$) & $\mathcal{W}_{6}$ ($\mathcal{L}_{8}$) & $\mathcal{W}_{6}$ ($\mathcal{L}_{10}$) & $\mathcal{W}_{8}$ ($\mathcal{L}_{12}$) & $\mathcal{W}_{3}$ ($\mathcal{L}_{14}$) & $\mathcal{W}_{6}$ ($\mathcal{L}_{16}$) & - \\
     & Up-sampling operations & \multicolumn{8}{c}{nearest\_conv} & - \\
    \bottomrule
  \end{tabular}}
  \vspace{-0.3cm}
\end{table*}

\subsection{Results}\label{results of alphaGAN on StyleGAN2}
The results of the architecture obtained via alphaGAN and original StyleGAN2 are shown in Tab. \ref{Scalability}. The generated images of the models with the best FID are presented in Fig. \ref{generated images of StyleGAN2}. As illustrated above, there are totally four paradigms of the search settings, i.e., ``micro", ``micro + dynamic $\left\{\mathcal{W}_{i}\right\}$", ``macro", and ``macro + dynamic $\left\{\mathcal{W}_{i}\right\}$".

\begin{figure*}[t]
  \vspace{-0.2cm}
  \centering
  \!\!\!\!\!\!\!\!\!\!
  \subfigure[\texttt{CelebA}]{\includegraphics[width=0.3\textwidth]{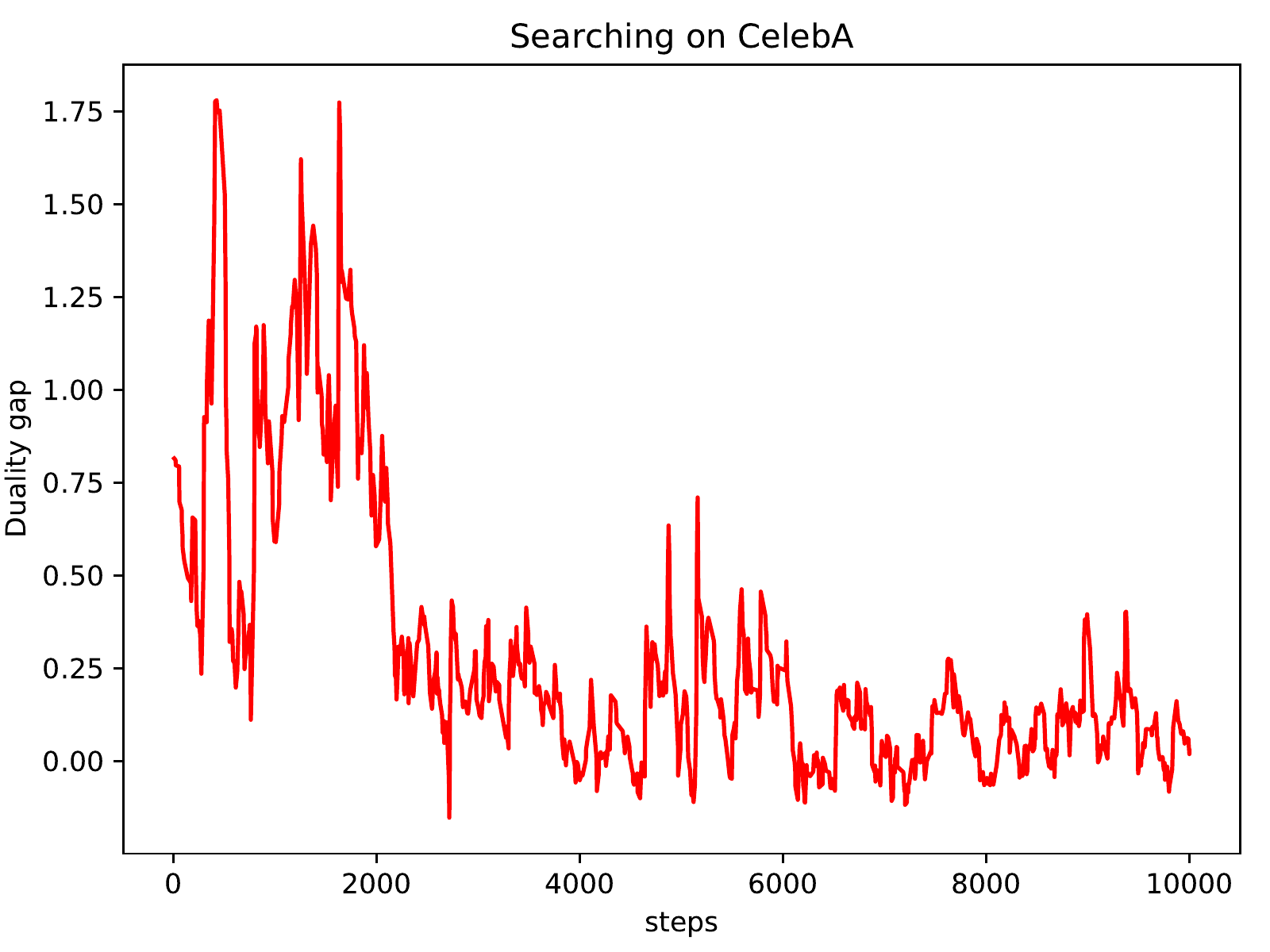}}
  \!\!\!\!
  \subfigure[\texttt{LSUN-church}]{\includegraphics[width=0.3\textwidth]{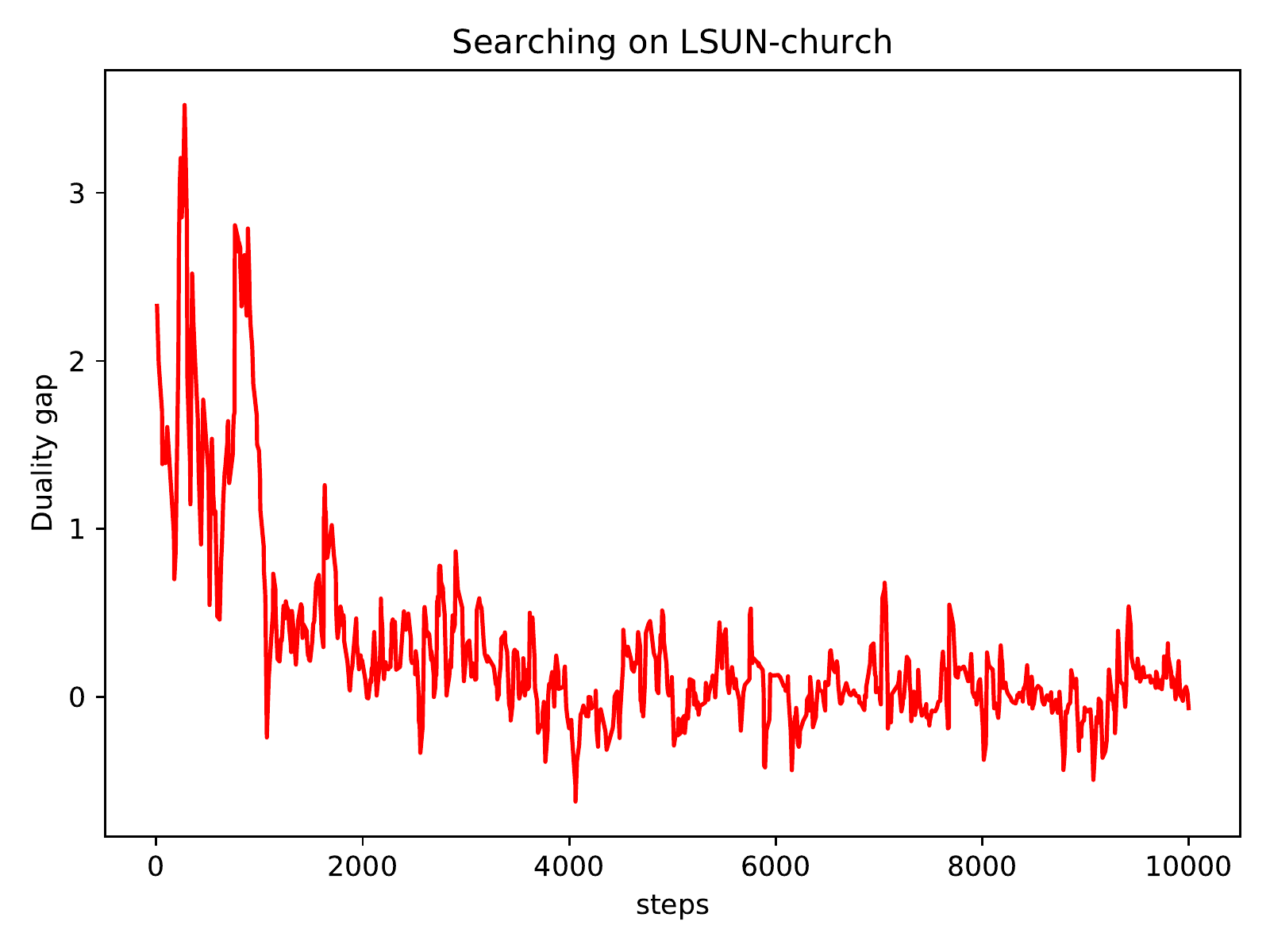}}\!\!\!
  \subfigure[\texttt{FFHQ}]{\includegraphics[width=0.3\textwidth]{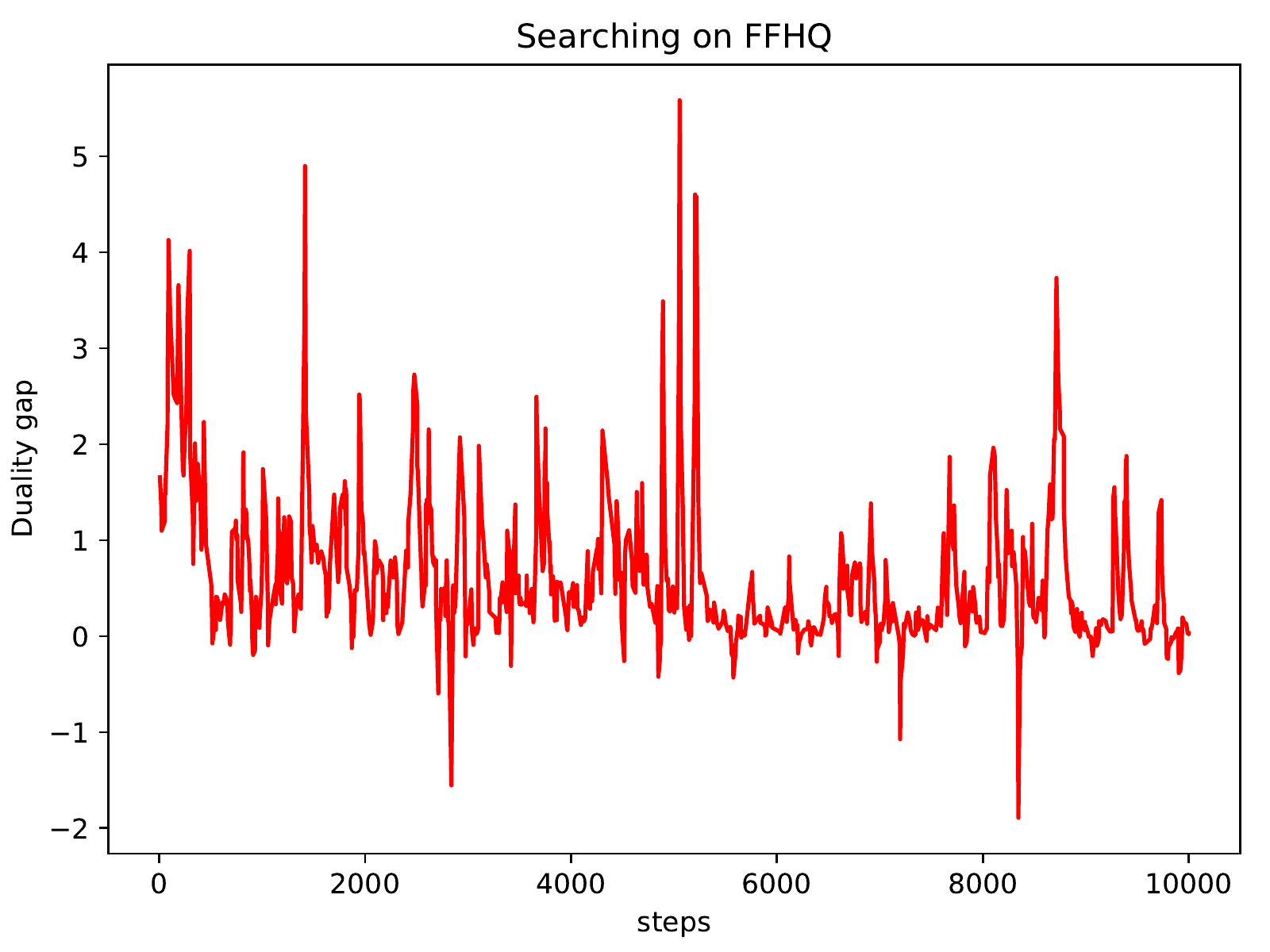}}
  \!\!\!\!\!\!\!
  \vspace{-0.5cm}
  \caption{The curve of duality gap during searching StyleGAN2 on CelebA, LSUN-church, and FFHQ under ``micro + dynamic $\left\{\mathcal{W}_i\right\}$" paradigm, respectively.}
  \label{Duality gap curve}
  \vspace{-0.5cm}
\end{figure*}

We plot the curve of duality gap while searching on CelebA, LSUN-church, and FFHQ, as shown in Fig. \ref{Duality gap curve}. We can see duality gap is decreasing with search proceeding. Interestingly, with the complexity of the datasets increasing, the trend of declining of duality gap becomes less apparent. That is because when the dataset becomes more complex, the approximation of duality gap becomes more difficult and the optimization of alphaGAN becomes more difficult.

In terms of the computational cost of the searching process, due to the efficient solving algorithm of alphaGAN, the cost of directly searching on large datasets is still very low (e.g., ``macro + dynamic $\left\{\mathcal{W}_{i}\right\}$" costs 3 GPU-hours on LSUN-church). Thus, we do not have to search on small datasets (e.g., CIFAR-10) and re-train on large datasets (e.g., FFHQ) like conventional NAS methods, such as DARTS \cite{liu2018darts}. The consistency of the searching dataset and the re-training dataset can enable us to focus on exploiting NAS to promote the architecture of GANs.

From Tab. \ref{Scalability}, we can see that considering the final performance, the architectures obtained via alphaGAN show the superiority over the original architecture of StyleGAN2 on all datasets. In terms of FID, the architectures obtained via alphaGAN are higher than both the released results of StyleGAN2 and the reproduced results on CelebA (1.94 vs 2.17), LSUN-church (2.86 vs 3.86), and FFHQ (2.75 vs 2.84), demonstrating that the improvements can be achieved via introducing NAS to GANs to promote the architectures of GANs, which is our initial motivation. To be highlighted, it is non-trivial to further boost the performance of StyleGAN2 due to its existing excellent performance. Moreover, to the best of our knowledge, the architectures obtained via alphaGAN reach state-of-the-arts on CelebA, LSUN-church, and FFHQ. However, we admit that the model sizes of the searched architectures are much larger than that of the original architecture due to the utilization of convolution with large kernels (e.g. conv\_5x5), which can be a possible reason for the better performance.

Besides searching and re-training on the same dataset, we also validate the transferability of alphaGAN on multiple datasets, shown in Tab. \ref{Scalability}. Unfortunately, only ``micro" paradigm is not confined to certain datasets, as mentioned in Sec. \ref{macro/micro}. Thus we adopt ``micro" paradigm. We search the architecture on CelebA and re-train it on FFHQ. The performance of the architecture searched on CelebA is comparable with the released results of StyleGAN2, with an equal number of parameters. 

Besides the final performance, we also want to highlight the faster convergence speed of the architectures obtained via alphaGAN over original StyleGAN2 in re-training, as depicted in Fig. \ref{Re-training curve}. Measured by FID, the convergence speed of the searched architectures is faster than original StyleGAN2 on all datasets, illustrating the superiority of the architectures obtained via alphaGAN. Moreover, it is interesting that the gain of the convergence speed on LSUN-church and FFHQ is higher than that on CelebA, demonstrating that the gain of convergence speed brought by alphaGAN increases with the complexity of the datasets.

Regarding AdversarialNAS, despite that they do not conduct experiments on large datasets, we re-implement AdversarialNAS on StyleGAN2, and show the results in Tab. \ref{AdversarialNAS on large datasets}. We select two better paradigms in micro paradigms and macro paradigms, i.e., ``micro + dynamic $\left\{\mathcal{W}_i\right\}$" paradigm and ``macro" paradigm. As for ``micro + dynamic $\left\{\mathcal{W}_{i}\right\}$" paradigm, alphaGAN outperforms AdversarialNAS 1.01 on CelebA (1.94 vs 2.95), 1.21 on LSUN-church (2.86 vs 4.07), and 0.72 on FFHQ (2.75 vs 3.47), respectively. As for ``macro" paradigm, alphaGAN outperforms AdversarialNAS 1.12 on CelebA (2.12 vs 3.24), 2.98 on LSUN-church (3.05 vs 6.03), and 1.03 on FFHQ (2.95 vs 3.98), respectively. Moreover, most architectures searched via AdversarialNAS are inferior to that of StyleGAN2 baseline. The architectures obtained via AdversarialNAS are shown in the supplementary material.

\begin{table*}[t]
  \caption{The results of re-implementing AdversarialNAS on CelebA, LSUN-church, FFHQ. $\dagger$ denotes the reproduced results based on the official released code.}
  \label{AdversarialNAS on large datasets}
  \centering
  \vspace{-0.3cm}
  \resizebox{\textwidth}{!}{
  \begin{tabular}{c|ccc|ccc|c}
    \toprule
    \multirow{2}*{Paradigm} & \multicolumn{3}{c|}{Search} & \multicolumn{3}{c|}{Re-train} & \multirow{2}*{FID} \\
    \cline{2-7}
    & Dataset & \tabincell{c}{Cost\\ (GPU-hours)} & Search space & Dataset & kimg & \tabincell{c}{Model size\\ (M)} \\
    \midrule
    \midrule
    \multirow{2}*{StyleGAN2(\cite{karras2020analyzing})} & - & - & - & church & $48000$ & $30.03$ & $3.86$ \\
     & - & - & - & FFHQ & $25000$ & $30.37$ & $2.84$ \\
    \hline
    \multirow{3}*{StyleGAN2$\dagger$} & - & - & - & CelebA & $15000$ & $29.33$ & $2.17$ \\
     & - & - & - & church & $15000$ & $30.03$ & $3.35$ \\
     & - & - & - & FFHQ & $20000$ & $30.37$ & $3.50$ \\
    \midrule
    \midrule
    \multirow{3}*{alphaGAN$_{(s)}$ + micro + dynamic $\left\{\mathcal{W}_{i}\right\}$} & CelebA & $2.5$ & $\sim6.2\times10^{11}$ & CelebA & $15000$ & $29.33$ & $\textbf{1.94}$ \\
     & church & $3$ & $\sim4.0\times10^{13}$ & church & $15000$ & $52.32$ & $\textbf{2.86}$ \\
     & FFHQ & $12$ & $\sim1.6\times10^{17}$ & FFHQ & $20000$ & $52.73$ & $\textbf{2.75}$ \\
    \hline
    \multirow{3}*{AdversarialNAS + micro + dynamic $\left\{\mathcal{W}_{i}\right\}$} & CelebA & $2.8$ & $\sim3.7\times10^{16}$ & CelebA & $15000$ & $18.32$ & $2.95$ \\
     & church & $4.5$ & $\sim2.1\times10^{19}$ & church & $15000$ & $18.89$ & $4.07$ \\
     & FFHQ & $21$ & $\sim6.9\times10^{24}$ & FFHQ & $20000$ & $52.73$ & $3.47$ \\
    \hline
    \multirow{3}*{alphaGAN$_{(s)}$ + macro} & CelebA & $1.5$ & $\sim1.8\times10^5$ & CelebA & $15000$ & $41.39$ & $2.12$ \\
     & church & $1.5$ & $\sim1.6\times10^6$ & church & $15000$ & $41.44$ & $3.05$ \\
     & FFHQ & $6$ & $\sim1.3\times10^9$ & FFHQ & $20000$ & $31.68$ & $2.95$ \\
    \hline
    \multirow{3}*{AdversarialNAS + macro} & CelebA & $1.8$ & $\sim1.1\times10^{10}$ & CelebA & $15000$ & $18.32$ & $3.24$ \\
     & church & $2.5$ & $\sim8.5\times10^{11}$ & church & $15000$ & $23.09$ & $6.03$ \\
     & FFHQ & $6.4$ & $\sim5.6\times10^{16}$ & FFHQ & $20000$ & $21.33$ & $3.98$ \\
    \bottomrule
  \end{tabular}}
  \vspace{-0.2cm}
\end{table*}

Among the four searching paradigms, we can find that ``micro + dynamic $\left\{\mathcal{W}_{i}\right\}$" obtains the architectures with the best performance on LSUN-church and CelebA, further validating our motivation to search the most suitable intermediate latent $\left\{\mathcal{W}_{i}\right\}{i=1,...,n}$ for each convolutional layer $\left\{L_{j}\right\}_{j=1,...,m}$ in the synthesis network. To be highlighted, FID of the architecture obtained via ``micro + dynamic $\left\{\mathcal{W}_{i}\right\}$" is higher than the original architecture (1.94 vs 2.17), with completely equal model size (29.33M) and completely identical operation (i.e., conv\_3x3), which demonstrates that directly adopting the latent $\left\{\mathcal{W}_{n}\right\}$ of the last fully-connected layer in the mapping network is not optimal. The most suitable intermediate latent $\left\{\mathcal{W}_{i}\right\}$ for each convolutional layer $\left\{L_{j}\right\}_{j=1,...,m}$ should be selected carefully, and we exploit alphaGAN to select $\left\{\mathcal{W}_{i}\right\}$ towards pure Nash Equilibrium, proven to be effective.    

\begin{table*}[t!]
  \caption{The searched architectures on FFHQ. In the line of ``Macro", ``c\_1" represents conv\_1x1, ``de" represents ``deconv", ``ne\_c" represents ``nearest\_conv", and ``bi\_c" represents ``bilinear\_conv". $\mathcal{L}_{i}$ in the bracket is the according index of the layer in Fig. \ref{The depiction of StyleGAN2}.}
  \vspace{-0.3cm}
  \label{Searched_arch_FFHQ}
  \centering
  \resizebox{\textwidth}{!}{
  \begin{tabular}{c|c|ccccccccc}
    \toprule
    \multirow{2}*{Paradigm} & \multirow{2}*{Searched operation/$\mathcal{W}_{i}$} & \multicolumn{9}{c}{Input resolution/Layer index} \\
    \cline{3-11}
     &  & 4x4 & 8x8 & 16x16 & 32x32 & 64x64 & 128x128 & 256x256 & 512x512 & 1024x1024 \\
    \midrule
    \midrule
    \multirow{4}*{Micro} & $\mathcal{W}_i$ for the normal operation & \multicolumn{9}{c}{-}\\ 
     & Normal operation & \multicolumn{9}{c}{conv\_5x5} \\
     & $\mathcal{W}_i$ for the up-sampling operation & \multicolumn{9}{c}{-} \\
     & Up-sampling operation & \multicolumn{9}{c}{bilinear\_conv} \\
    \hline
    \multirow{4}*{\tabincell{c}{Micro +\\ dynamic $\left\{\mathcal{W}_{i}\right\}$}} & $\mathcal{W}_i$ for the normal operation & $\mathcal{W}_{3}$ ($\mathcal{L}_{1}$) & $\mathcal{W}_{7}$ ($\mathcal{L}_{3}$) & $\mathcal{W}_{4}$ ($\mathcal{L}_{5}$) & $\mathcal{W}_{3}$ ($\mathcal{L}_{7}$) & $\mathcal{W}_{4}$ ($\mathcal{L}_{9}$) & $\mathcal{W}_{3}$ ($\mathcal{L}_{11}$) & $\mathcal{W}_{2}$ ($\mathcal{L}_{13}$) & $\mathcal{W}_{4}$ ($\mathcal{L}_{15}$) & $\mathcal{W}_{4}$ ($\mathcal{L}_{17}$), ${\mathcal{W}_{2}}^{*}$ \\ 
     & Normal operation & \multicolumn{9}{c}{conv\_5x5} \\
     & $\mathcal{W}_i$ for the up-sampling operation & $\mathcal{W}_{3}$ ($\mathcal{L}_{2}$) & $\mathcal{W}_{6}$ ($\mathcal{L}_{4}$) & $\mathcal{W}_{8}$ ($\mathcal{L}_{6}$) & $\mathcal{W}_{6}$ ($\mathcal{L}_{8}$) & $\mathcal{W}_{6}$ ($\mathcal{L}_{10}$) & $\mathcal{W}_{8}$ ($\mathcal{L}_{12}$) & $\mathcal{W}_{3}$ ($\mathcal{L}_{14}$) & $\mathcal{W}_{6}$ ($\mathcal{L}_{16}$) & - \\
     & Up-sampling operation & \multicolumn{8}{c}{bilinear\_conv} & - \\
    \hline
    \multirow{4}*{Macro} & $\mathcal{W}_i$ for the normal operation & \multicolumn{9}{c}{-} \\  
     & Normal operation & c\_1 ($\mathcal{L}_{1}$) & c\_3 ($\mathcal{L}_{3}$) & c\_5 ($\mathcal{L}_{5}$) & c\_1 ($\mathcal{L}_{7}$) & c\_3 ($\mathcal{L}_{9}$) & c\_5 ($\mathcal{L}_{11}$) & c\_5 ($\mathcal{L}_{13}$) & c\_3 ($\mathcal{L}_{15}$) & c\_3 ($\mathcal{L}_{17}$) \\
     & $\mathcal{W}_i$ for the up-sampling operation & \multicolumn{9}{c}{-}\\
     & Up-sampling operations & de ($\mathcal{L}_{2}$) & bi\_c ($\mathcal{L}_{4}$) & ne\_c ($\mathcal{L}_{6}$) & bi\_c ($\mathcal{L}_{8}$) & de ($\mathcal{L}_{10}$) & bi\_c ($\mathcal{L}_{12}$) & bi\_c ($\mathcal{L}_{14}$) & de ($\mathcal{L}_{16}$) & - \\
    \bottomrule
  \end{tabular}}
  \vspace{-0.3cm}
\end{table*}

To be noted, not all searching paradigms work and some paradigms fail on certain datasets, e.g., ``macro + dynamic $\left\{\mathcal{W}_{i}\right\}$" on church and FFHQ. We conjecture that there are three main reasons for the inferior architecture obtained via ``macro + dynamic $\left\{\mathcal{W}_{i}\right\}$". First, large searching space leads to obtaining inferior architectures. Second, the complexity of datasets can also lead to inferior architectures. All searching paradigms succeed on CelebA, but ``macro + dynamic $\left\{\mathcal{W}_{i}\right\}$" fails on LSUN-church or FFHQ. The internal distribution in LSUN-church or FFHQ is more difficult to fit and explore than that of CelebA, thus resulting in inferior architectures. Third, the combination of ``macro" and ``dynamic $\left\{\mathcal{W}_{i}\right\}$" is not the optimal paradigm for alphaGAN. Both ``macro" and ``micro + dynamic $\left\{\mathcal{W}_{i}\right\}$" work well on LSUN-church and FFHQ, but the combination of them brings inferior architecture. Inferior architectures (i.e., failure cases) are a common problem \cite{zela2019understanding} in the research field of NAS, and we leave the alleviation of failure cases as future work.

\subsection{Analysis of the obtained architectures}
In this section, we analyze the architectures obtained via alphaGAN, providing some insights into the relationship between the performance and the architecture of StyleGAN2.

\subsubsection{``micro + dynamic $\left\{\mathcal{W}_{i}\right\}$"}
Interestingly, the architectures obtained via ``micro + dynamic $\left\{\mathcal{W}_{i}\right\}$" achieve the best performance over the architectures of other searching paradigms and the original architecture of StyleGAN2, indicating that the details in the architectures of ``micro + dynamic $\left\{\mathcal{W}_{i}\right\}$" deserve more attention. Observing and analyzing the architectures shown in Tab. \ref{Searched_arch_style}, some interesting points can be found.

As for selecting the latent $\left\{\mathcal{W}_{i}\right\}_{i=1,...,n}$ for each convolutional layer $\left\{\mathcal{L}_{j}\right\}_{j=1,...,m}$, the obtained architecture on it shows some tendencies. On LSUN-church, the bottom convolutional layers of the synthesis network tend to take $\mathcal{W}_i$ from both the bottom and the top of the mapping network as the input of the style information, whereas the top convolutional layers tend to take $\mathcal{W}_i$ from the bottom of the mapping network as the input of the style information. And $\mathcal{W}_{4}$ dominates in $\left\{\mathcal{W}_i\right\}$ selected for up-sampling operations.

On CelebA, the obtained architecture shows completely opposite tendencies in selecting $\left\{\mathcal{W}_{i}\right\}_{i=1,...,n}$ compared with the architecture on LSUN-church. The bottom convolutional layers of the synthesis network tend to take $\mathcal{W}_i$ from the bottom of the mapping network, whereas the top convolutional layers in the synthesis network tend to take $\mathcal{W}_i$ from both the top and the bottom of the mapping network. We conjecture that the difference of the tendencies between CelebA and LSUN-church results from the intrinsic distribution difference of the two datasets.

On FFHQ, no apparent tendencies in selecting $\left\{\mathcal{W}_{i}\right\}_{i=1,...,n}$ are observed like LSUN-church and CelebA. We observe two phenomena of the architecture on FFHQ. First, $\left\{\mathcal{W}_{i=3,4,6}\right\}$ are selected most frequently. Second, normal operations tend to take $\mathcal{W}_{4}$ frequently, and up-sampling operations tend to take $\mathcal{W}_{6}$ frequently.

As for the operations, the architectures on complex datasets (e.g., LSUN-church and FFHQ) tend to adopt convolution with large kernels (e.g., conv\_5x5), the architecture on relatively small datasets (e.g., CelebA) tends to adopt convolution with common kernels (e.g., conv\_3x3).

\subsubsection{Architectures on FFHQ}
As for FFHQ, we have to admit that even with the architectures obtained via alphaGAN, it is difficult to further promote the performance of StyleGAN2. Among the four searching paradigms, only ``micro" and ``micro + dynamic $\left\{\mathcal{W}_i\right\}$" achieve superior performance over the original StyleGAN2 \cite{karras2020analyzing}. Thus, it is necessary to analyze the architectures with comparable performance, and summarize some directions to further improve. The architectures searched on FFHQ are presented in Tab. \ref{Searched_arch_FFHQ}. 

As for ``micro" and ``micro + dynamic $\left\{\mathcal{W}_{i}\right\}$", the architectures tend to adopt convolution with large kernels (e.g., conv\_5x5), demonstrating that large receptive field is beneficial to the performance of StyleGAN2 on FFHQ. Moreover, the up-sampling operations tend to adopt ``nearest\_conv" or ``bilinear\_conv", rather than ``deconv".

As for ``macro", the architecture of it achieves comparable performance with the released results of StyleGAN2, with a slightly higher number of parameters (31.68M vs 30.37M). The mixture of three normal operations dominates in the architecture of ``macro", illustrating that the mixture of convolution with different kernels is beneficial to the performance of StyleGAN2. Among up-sampling operations, ``nearest\_conv" and ``bilinear\_conv" dominate in the architecture of ``macro", consistent with the phenomenon observed in ``micro" and ``micro + dynamic $\left\{\mathcal{W}_{i}\right\}$".

\section{Discussion}
In this section, we list several limitations of alphaGAN and we hope researchers will improve these limitations in future.

First, can alphaGAN guarantee reaching Nash Equilibrium during search? we have to admit that despite that alphaGAN exploits duality gap to guide the search towards Nash Equilibrium, it cannot be proven that the search process will converge at exactly Nash Equilibrium. As the optimization of GANs is a non-convex con-concave minimax problem and the generator is with two parts of parameters (i.e., the weight parameters $\omega_{G}$ and the architecture parameters $\alpha_{G}$), solving the exactly Nash Equilibrium is highly non-trivial. Empirically, we show that with the descent of duality gap (i.e., near to Nash Equilibrium), the architecture of the generator becomes better, shown in Fig. \ref{Duality gap curve of conventional GANs} and Fig. \ref{performance during search}.

Second, why simultaneously searching both networks leads to sub-optimal architectures? Intuitively, searching both networks should obtain a better generator, because a discriminator with a specified architecture can provide more effective supervision signals. However, in our experiments, searching both leads to sub-optimal architectures. With the architecture of the discriminator added to the search process, the increasing combinatorial complexity makes the problem more difficult to be solved. Thus, alphaGAN only searches the generator can be viewed as a compromise. Solving NAS-GAN problems more precisely is left as the future work.

Third, can we reduce the parameter size of the obtained generator of StyleGAN2? Regarding conventional GANs, we can obtain the generator with the better performance and the smaller parameter size, e.g., alphaGAN$_{(s)}$ on CIFAR-10 and STL-10. However, regarding StyleGAN2, it is not easy to obtain a light-weight generator with better performance, the performance increases with the cost the increase of the parameter size, e.g., ``alphaGAN$_{(s)}$ + micro + dynamic $\left\{\mathcal{W}_{i}\right\}$" on FFHQ. That indicates that reaching a trade-off on the performance and the parameter size for state-of-the-art GANs (e.g., StyleGAN2) is non-trivial and the architecture of state-of-the-art GANs deserves to be further explored.
\section{Conclusions}

alphaGAN, a fully differentiable architecture search framework for GANs, can boost the performance of GANs via exploring the architectures towards pure Nash Equilibrium. To be highlighted, not confined to conventional GANs, alphaGAN can be directly applied to state-of-the-art StyleGAN2, obtaining the superior architectures efficiently and achieving state-of-the-arts on three datasets, CelebA, LSUN-church, and FFHQ, demonstrating that the introduction of AutoML needs to consider the intrinsic property of GANs. There exist several directions to further explore. First, adding the constraint of the number of parameters in search and obtaining the architecture with less parameters and superior performance. Second, trying to accelerate the training speed of GANs from the perspective of AutoML, which we think is the critical point in the current research field of GANs. 



\bibliographystyle{IEEEtran}
\bibliography{references}

\begin{thebibliography}{10}
\providecommand{\url}[1]{#1}
\csname url@samestyle\endcsname
\providecommand{\newblock}{\relax}
\providecommand{\bibinfo}[2]{#2}
\providecommand{\BIBentrySTDinterwordspacing}{\spaceskip=0pt\relax}
\providecommand{\BIBentryALTinterwordstretchfactor}{4}
\providecommand{\BIBentryALTinterwordspacing}{\spaceskip=\fontdimen2\font plus
\BIBentryALTinterwordstretchfactor\fontdimen3\font minus
  \fontdimen4\font\relax}
\providecommand{\BIBforeignlanguage}[2]{{%
\expandafter\ifx\csname l@#1\endcsname\relax
\typeout{** WARNING: IEEEtran.bst: No hyphenation pattern has been}%
\typeout{** loaded for the language `#1'. Using the pattern for}%
\typeout{** the default language instead.}%
\else
\language=\csname l@#1\endcsname
\fi
#2}}
\providecommand{\BIBdecl}{\relax}
\BIBdecl

\bibitem{goodfellow2014generative}
I.~Goodfellow, J.~Pouget-Abadie, M.~Mirza, B.~Xu, D.~Warde-Farley, S.~Ozair,
  A.~Courville, and Y.~Bengio, ``Generative adversarial nets,'' in
  \emph{Advances in neural information processing systems}, 2014, pp.
  2672--2680.

\bibitem{brock2018large}
A.~Brock, J.~Donahue, and K.~Simonyan, ``Large scale gan training for high
  fidelity natural image synthesis,'' \emph{arXiv preprint arXiv:1809.11096},
  2018.

\bibitem{donahue2019large}
J.~Donahue and K.~Simonyan, ``Large scale adversarial representation
  learning,'' in \emph{Advances in Neural Information Processing Systems},
  2019, pp. 10\,542--10\,552.

\bibitem{karras2017progressive}
T.~Karras, T.~Aila, S.~Laine, and J.~Lehtinen, ``Progressive growing of gans
  for improved quality, stability, and variation,'' \emph{arXiv preprint
  arXiv:1710.10196}, 2017.

\bibitem{karras2019style}
T.~Karras, S.~Laine, and T.~Aila, ``A style-based generator architecture for
  generative adversarial networks,'' in \emph{Proceedings of the IEEE
  conference on computer vision and pattern recognition}, 2019, pp. 4401--4410.

\bibitem{karras2020analyzing}
T.~Karras, S.~Laine, M.~Aittala, J.~Hellsten, J.~Lehtinen, and T.~Aila,
  ``Analyzing and improving the image quality of stylegan,'' in
  \emph{Proceedings of the IEEE/CVF Conference on Computer Vision and Pattern
  Recognition}, 2020, pp. 8110--8119.

\bibitem{isola2017image}
P.~Isola, J.-Y. Zhu, T.~Zhou, and A.~A. Efros, ``Image-to-image translation
  with conditional adversarial networks,'' in \emph{Proceedings of the IEEE
  conference on computer vision and pattern recognition}, 2017, pp. 1125--1134.

\bibitem{zhu2017unpaired}
J.-Y. Zhu, T.~Park, P.~Isola, and A.~A. Efros, ``Unpaired image-to-image
  translation using cycle-consistent adversarial networks,'' in
  \emph{Proceedings of the IEEE international conference on computer vision},
  2017, pp. 2223--2232.

\bibitem{wang2018high}
T.-C. Wang, M.-Y. Liu, J.-Y. Zhu, A.~Tao, J.~Kautz, and B.~Catanzaro,
  ``High-resolution image synthesis and semantic manipulation with conditional
  gans,'' in \emph{Proceedings of the IEEE conference on computer vision and
  pattern recognition}, 2018, pp. 8798--8807.

\bibitem{choi2018stargan}
Y.~Choi, M.~Choi, M.~Kim, J.-W. Ha, S.~Kim, and J.~Choo, ``Stargan: Unified
  generative adversarial networks for multi-domain image-to-image
  translation,'' in \emph{Proceedings of the IEEE conference on computer vision
  and pattern recognition}, 2018, pp. 8789--8797.

\bibitem{choi2020stargan}
Y.~Choi, Y.~Uh, J.~Yoo, and J.-W. Ha, ``Stargan v2: Diverse image synthesis for
  multiple domains,'' in \emph{Proceedings of the IEEE/CVF Conference on
  Computer Vision and Pattern Recognition}, 2020, pp. 8188--8197.

\bibitem{park2020contrastive}
T.~Park, A.~A. Efros, R.~Zhang, and J.-Y. Zhu, ``Contrastive learning for
  unpaired image-to-image translation,'' in \emph{European Conference on
  Computer Vision}.\hskip 1em plus 0.5em minus 0.4em\relax Springer, 2020, pp.
  319--345.

\bibitem{li2017adversarial}
J.~Li, W.~Monroe, T.~Shi, S.~Jean, A.~Ritter, and D.~Jurafsky, ``Adversarial
  learning for neural dialogue generation,'' \emph{arXiv preprint
  arXiv:1701.06547}, 2017.

\bibitem{yu2018generative}
J.~Yu, Z.~Lin, J.~Yang, X.~Shen, X.~Lu, and T.~S. Huang, ``Generative image
  inpainting with contextual attention,'' in \emph{Proceedings of the IEEE
  conference on computer vision and pattern recognition}, 2018, pp. 5505--5514.

\bibitem{oliehoek2017gangs}
F.~A. Oliehoek, R.~Savani, J.~Gallego-Posada, E.~Van~der Pol, E.~D. De~Jong,
  and R.~Gro{\ss}, ``Gangs: Generative adversarial network games,'' \emph{arXiv
  preprint arXiv:1712.00679}, 2017.

\bibitem{salimans2016improved}
T.~Salimans, I.~Goodfellow, W.~Zaremba, V.~Cheung, A.~Radford, and X.~Chen,
  ``Improved techniques for training gans,'' in \emph{Advances in neural
  information processing systems}, 2016, pp. 2234--2242.

\bibitem{arjovsky2017wasserstein}
M.~Arjovsky, S.~Chintala, and L.~Bottou, ``Wasserstein gan,'' \emph{arXiv
  preprint arXiv:1701.07875}, 2017.

\bibitem{mao2017least}
X.~Mao, Q.~Li, H.~Xie, R.~Y. Lau, Z.~Wang, and S.~Paul~Smolley, ``Least squares
  generative adversarial networks,'' in \emph{Proceedings of the IEEE
  International Conference on Computer Vision}, 2017, pp. 2794--2802.

\bibitem{miyato2018spectral}
T.~Miyato, T.~Kataoka, M.~Koyama, and Y.~Yoshida, ``Spectral normalization for
  generative adversarial networks,'' \emph{arXiv preprint arXiv:1802.05957},
  2018.

\bibitem{gulrajani2017improved}
I.~Gulrajani, F.~Ahmed, M.~Arjovsky, V.~Dumoulin, and A.~C. Courville,
  ``Improved training of wasserstein gans,'' in \emph{Advances in neural
  information processing systems}, 2017, pp. 5767--5777.

\bibitem{kodali2017convergence}
N.~Kodali, J.~Abernethy, J.~Hays, and Z.~Kira, ``On convergence and stability
  of gans,'' \emph{arXiv preprint arXiv:1705.07215}, 2017.

\bibitem{sinha2019small}
S.~Sinha, H.~Zhang, A.~Goyal, Y.~Bengio, H.~Larochelle, and A.~Odena,
  ``Small-gan: Speeding up gan training using core-sets,'' \emph{arXiv preprint
  arXiv:1910.13540}, 2019.

\bibitem{zhang2019consistency}
H.~Zhang, Z.~Zhang, A.~Odena, and H.~Lee, ``Consistency regularization for
  generative adversarial networks,'' \emph{arXiv preprint arXiv:1910.12027},
  2019.

\bibitem{zhao2020improved}
Z.~Zhao, S.~Singh, H.~Lee, Z.~Zhang, A.~Odena, and H.~Zhang, ``Improved
  consistency regularization for gans,'' \emph{arXiv preprint
  arXiv:2002.04724}, 2020.

\bibitem{zhao2020differentiable}
S.~Zhao, Z.~Liu, J.~Lin, J.-Y. Zhu, and S.~Han, ``Differentiable augmentation
  for data-efficient gan training,'' \emph{arXiv preprint arXiv:2006.10738},
  2020.

\bibitem{radford2015unsupervised}
A.~Radford, L.~Metz, and S.~Chintala, ``Unsupervised representation learning
  with deep convolutional generative adversarial networks,'' \emph{arXiv
  preprint arXiv:1511.06434}, 2015.

\bibitem{ye2019progressive}
S.~Ye, X.~Feng, T.~Zhang, X.~Ma, S.~Lin, Z.~Li, K.~Xu, W.~Wen, S.~Liu, J.~Tang
  \emph{et~al.}, ``Progressive dnn compression: A key to achieve ultra-high
  weight pruning and quantization rates using admm,'' \emph{arXiv preprint
  arXiv:1903.09769}, 2019.

\bibitem{he2016deep}
K.~He, X.~Zhang, S.~Ren, and J.~Sun, ``Deep residual learning for image
  recognition,'' in \emph{Proceedings of the IEEE conference on computer vision
  and pattern recognition}, 2016, pp. 770--778.

\bibitem{liu2015faceattributes}
Z.~Liu, P.~Luo, X.~Wang, and X.~Tang, ``Deep learning face attributes in the
  wild,'' in \emph{Proceedings of International Conference on Computer Vision
  (ICCV)}, December 2015.

\bibitem{yu15lsun}
F.~Yu, Y.~Zhang, S.~Song, A.~Seff, and J.~Xiao, ``Lsun: Construction of a
  large-scale image dataset using deep learning with humans in the loop,''
  \emph{arXiv preprint arXiv:1506.03365}, 2015.

\bibitem{zoph2016neural}
B.~Zoph and Q.~V. Le, ``Neural architecture search with reinforcement
  learning,'' \emph{arXiv preprint arXiv:1611.01578}, 2016.

\bibitem{liu2018darts}
H.~Liu, K.~Simonyan, and Y.~Yang, ``Darts: Differentiable architecture
  search,'' \emph{arXiv preprint arXiv:1806.09055}, 2018.

\bibitem{zoph2018learning}
B.~Zoph, V.~Vasudevan, J.~Shlens, and Q.~V. Le, ``Learning transferable
  architectures for scalable image recognition,'' in \emph{Proceedings of the
  IEEE conference on computer vision and pattern recognition}, 2018, pp.
  8697--8710.

\bibitem{brock2017smash}
A.~Brock, T.~Lim, J.~M. Ritchie, and N.~Weston, ``Smash: one-shot model
  architecture search through hypernetworks,'' \emph{arXiv preprint
  arXiv:1708.05344}, 2017.

\bibitem{nash1950equilibrium}
J.~F. Nash \emph{et~al.}, ``Equilibrium points in n-person games,''
  \emph{Proceedings of the national academy of sciences}, vol.~36, no.~1, pp.
  48--49, 1950.

\bibitem{jin2020local}
C.~Jin, P.~Netrapalli, and M.~Jordan, ``What is local optimality in
  nonconvex-nonconcave minimax optimization?'' in \emph{International
  Conference on Machine Learning}.\hskip 1em plus 0.5em minus 0.4em\relax PMLR,
  2020, pp. 4880--4889.

\bibitem{grnarova2019domain}
P.~Grnarova, K.~Y. Levy, A.~Lucchi, N.~Perraudin, I.~Goodfellow, T.~Hofmann,
  and A.~Krause, ``A domain agnostic measure for monitoring and evaluating
  gans,'' in \emph{Advances in Neural Information Processing Systems}, 2019,
  pp. 12\,069--12\,079.

\bibitem{heusel2017gans}
M.~Heusel, H.~Ramsauer, T.~Unterthiner, B.~Nessler, and S.~Hochreiter, ``Gans
  trained by a two time-scale update rule converge to a local nash
  equilibrium,'' in \emph{Advances in neural information processing systems},
  2017, pp. 6626--6637.

\bibitem{gao2019adversarialnas}
C.~Gao, Y.~Chen, S.~Liu, Z.~Tan, and S.~Yan, ``Adversarialnas: Adversarial
  neural architecture search for gans,'' \emph{arXiv preprint
  arXiv:1912.02037}, 2019.

\bibitem{lin2019gradient}
T.~Lin, C.~Jin, and M.~I. Jordan, ``On gradient descent ascent for
  nonconvex-concave minimax problems,'' \emph{arXiv preprint arXiv:1906.00331},
  2019.

\bibitem{karras2020training}
T.~Karras, M.~Aittala, J.~Hellsten, S.~Laine, J.~Lehtinen, and T.~Aila,
  ``Training generative adversarial networks with limited data,'' \emph{arXiv
  preprint arXiv:2006.06676}, 2020.

\bibitem{zhang2018self}
H.~Zhang, I.~Goodfellow, D.~Metaxas, and A.~Odena, ``Self-attention generative
  adversarial networks,'' \emph{arXiv preprint arXiv:1805.08318}, 2018.

\bibitem{huang2017arbitrary}
X.~Huang and S.~Belongie, ``Arbitrary style transfer in real-time with adaptive
  instance normalization,'' in \emph{Proceedings of the IEEE International
  Conference on Computer Vision}, 2017, pp. 1501--1510.

\bibitem{liu2017hierarchical}
H.~Liu, K.~Simonyan, O.~Vinyals, C.~Fernando, and K.~Kavukcuoglu,
  ``Hierarchical representations for efficient architecture search,''
  \emph{arXiv preprint arXiv:1711.00436}, 2017.

\bibitem{liu2018progressive}
C.~Liu, B.~Zoph, M.~Neumann, J.~Shlens, W.~Hua, L.-J. Li, L.~Fei-Fei,
  A.~Yuille, J.~Huang, and K.~Murphy, ``Progressive neural architecture
  search,'' in \emph{Proceedings of the European Conference on Computer Vision
  (ECCV)}, 2018, pp. 19--34.

\bibitem{real2019regularized}
E.~Real, A.~Aggarwal, Y.~Huang, and Q.~V. Le, ``Regularized evolution for image
  classifier architecture search,'' in \emph{Proceedings of the aaai conference
  on artificial intelligence}, vol.~33, 2019, pp. 4780--4789.

\bibitem{ghiasi2019fpn}
G.~Ghiasi, T.-Y. Lin, and Q.~V. Le, ``Nas-fpn: Learning scalable feature
  pyramid architecture for object detection,'' in \emph{Proceedings of the IEEE
  Conference on Computer Vision and Pattern Recognition}, 2019, pp. 7036--7045.

\bibitem{chen2019detnas}
Y.~Chen, T.~Yang, X.~Zhang, G.~Meng, X.~Xiao, and J.~Sun, ``Detnas: Backbone
  search for object detection,'' in \emph{Advances in Neural Information
  Processing Systems}, 2019, pp. 6638--6648.

\bibitem{peng2019efficient}
J.~Peng, M.~Sun, Z.-X. ZHANG, T.~Tan, and J.~Yan, ``Efficient neural
  architecture transformation search in channel-level for object detection,''
  in \emph{Advances in Neural Information Processing Systems}, 2019, pp.
  14\,290--14\,299.

\bibitem{gong2019autogan}
X.~Gong, S.~Chang, Y.~Jiang, and Z.~Wang, ``Autogan: Neural architecture search
  for generative adversarial networks,'' in \emph{Proceedings of the IEEE
  International Conference on Computer Vision}, 2019, pp. 3224--3234.

\bibitem{wang2019agan}
H.~Wang and J.~Huan, ``Agan: Towards automated design of generative adversarial
  networks,'' \emph{arXiv preprint arXiv:1906.11080}, 2019.

\bibitem{tian2020off}
Y.~Tian, Q.~Wang, Z.~Huang, W.~Li, D.~Dai, M.~Yang, J.~Wang, and O.~Fink,
  ``Off-policy reinforcement learning for efficient and effective gan
  architecture search,'' \emph{arXiv preprint arXiv:2007.09180}, 2020.

\bibitem{lee2020journey}
R.~Lee, {\L}.~Dudziak, M.~Abdelfattah, S.~I. Venieris, H.~Kim, H.~Wen, and
  N.~D. Lane, ``Journey towards tiny perceptual super-resolution,'' \emph{arXiv
  preprint arXiv:2007.04356}, 2020.

\bibitem{osborne1994course}
M.~J. Osborne and A.~Rubinstein, \emph{A course in game theory}.\hskip 1em plus
  0.5em minus 0.4em\relax MIT press, 1994.

\bibitem{du2013minimax}
D.-Z. Du and P.~M. Pardalos, \emph{Minimax and applications}.\hskip 1em plus
  0.5em minus 0.4em\relax Springer Science \& Business Media, 2013, vol.~4.

\bibitem{ho2016generative}
J.~Ho and S.~Ermon, ``Generative adversarial imitation learning,'' in
  \emph{Advances in neural information processing systems}, 2016, pp.
  4565--4573.

\bibitem{pinto2017robust}
L.~Pinto, J.~Davidson, R.~Sukthankar, and A.~Gupta, ``Robust adversarial
  reinforcement learning,'' in \emph{Proceedings of the 34th International
  Conference on Machine Learning-Volume 70}.\hskip 1em plus 0.5em minus
  0.4em\relax JMLR. org, 2017, pp. 2817--2826.

\bibitem{peng2020dggan}
\BIBentryALTinterwordspacing
C.~Peng, H.~Wang, X.~Wang, and Z.~Yang, ``{\{}DG{\}}-{\{}gan{\}}: the
  {\{}gan{\}} with the duality gap,'' 2020. [Online]. Available:
  \url{https://openreview.net/forum?id=ryxMW6EtPB}
\BIBentrySTDinterwordspacing

\bibitem{kingma2014adam}
D.~P. Kingma and J.~Ba, ``Adam: A method for stochastic optimization,''
  \emph{arXiv preprint arXiv:1412.6980}, 2014.

\bibitem{li2019random}
L.~Li and A.~Talwalkar, ``Random search and reproducibility for neural
  architecture search,'' \emph{arXiv preprint arXiv:1902.07638}, 2019.

\bibitem{he2019probgan}
H.~He, H.~Wang, G.-H. Lee, and Y.~Tian, ``Probgan: Towards probabilistic gan
  with theoretical guarantees.''

\bibitem{wang2018improving}
W.~Wang, Y.~Sun, and S.~Halgamuge, ``Improving mmd-gan training with repulsive
  loss function,'' \emph{arXiv preprint arXiv:1812.09916}, 2018.

\bibitem{arber2019understanding}
T.~E. Arber~Zela, T.~Saikia, Y.~Marrakchi, T.~Brox, and F.~Hutter,
  ``Understanding and robustifying differentiable architecture search,''
  \emph{arXiv preprint arXiv:1909.09656}, vol.~2, no.~4, p.~9, 2019.

\bibitem{pham2018efficient}
H.~Pham, M.~Y. Guan, B.~Zoph, Q.~V. Le, and J.~Dean, ``Efficient neural
  architecture search via parameter sharing,'' \emph{arXiv preprint
  arXiv:1802.03268}, 2018.

\bibitem{lin2019coco}
C.~H. Lin, C.-C. Chang, Y.-S. Chen, D.-C. Juan, W.~Wei, and H.-T. Chen,
  ``Coco-gan: generation by parts via conditional coordinating,'' in
  \emph{Proceedings of the IEEE International Conference on Computer Vision},
  2019, pp. 4512--4521.

\bibitem{karnewar2020msg}
A.~Karnewar and O.~Wang, ``Msg-gan: Multi-scale gradients for generative
  adversarial networks,'' in \emph{Proceedings of the IEEE/CVF Conference on
  Computer Vision and Pattern Recognition}, 2020, pp. 7799--7808.

\bibitem{zela2019understanding}
A.~Zela, T.~Elsken, T.~Saikia, Y.~Marrakchi, T.~Brox, and F.~Hutter,
  ``Understanding and robustifying differentiable architecture search,''
  \emph{arXiv preprint arXiv:1909.09656}, 2019.

\bibitem{maddison2014sampling}
C.~J. Maddison, D.~Tarlow, and T.~Minka, ``A* sampling,'' in \emph{Advances in
  Neural Information Processing Systems}, 2014, pp. 3086--3094.

\bibitem{he2020milenas}
C.~He, H.~Ye, L.~Shen, and T.~Zhang, ``Milenas: Efficient neural architecture
  search via mixed-level reformulation,'' in \emph{Proceedings of the IEEE/CVF
  Conference on Computer Vision and Pattern Recognition}, 2020, pp.
  11\,993--12\,002.

\end{thebibliography}
%


\vspace{-0.9cm}
\begin{IEEEbiography}
[{\includegraphics[height=1.16in,clip,keepaspectratio]{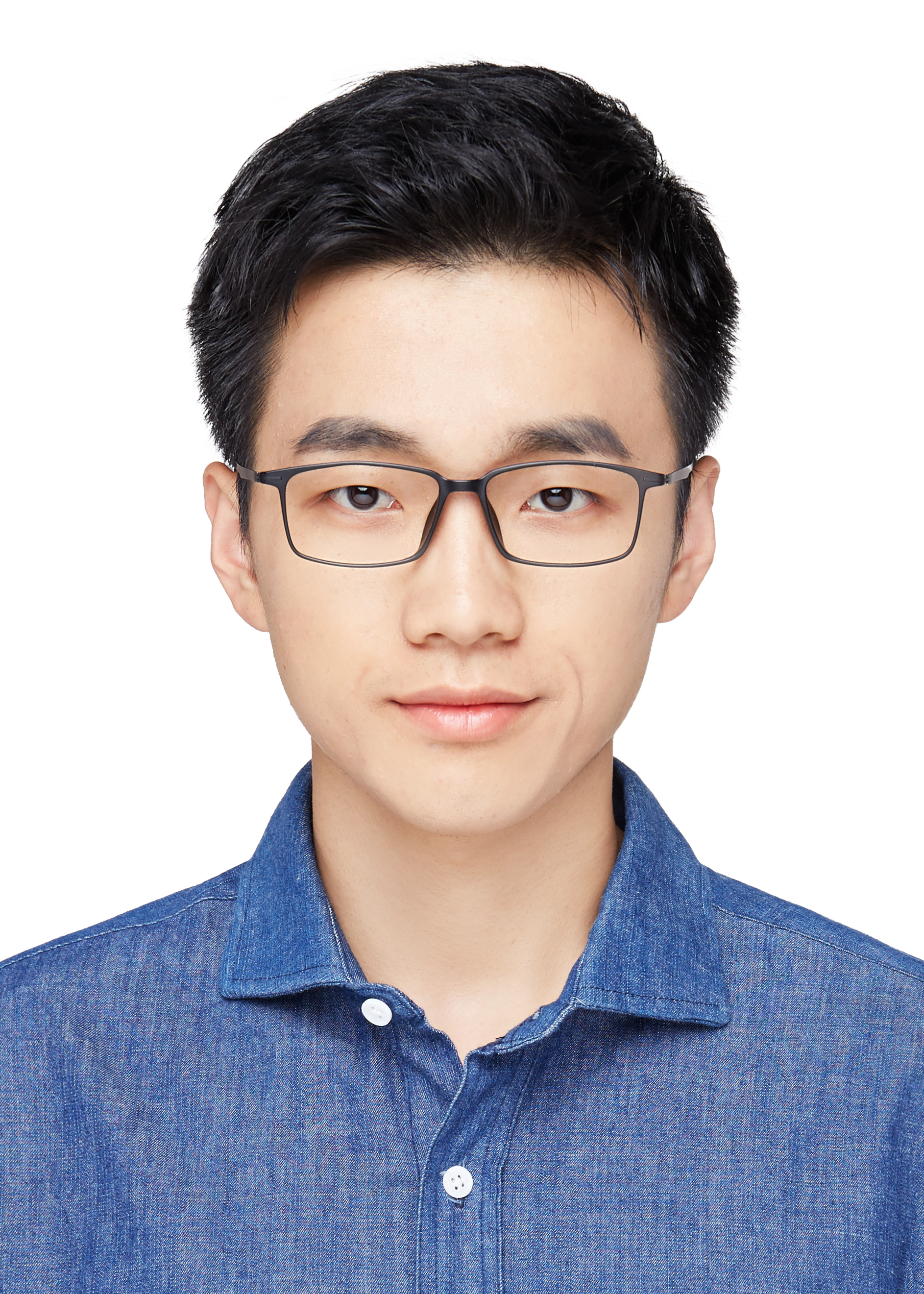}}]
{Yuesong Tian} received his BS degree in electric engineering from Zhejiang University, China, in 2017. He is currently working toward the PhD degree in the College of Biomedical Engineering and Instrument Science, Zhejiang University. He has been with Tencent AI Lab and Tencent Data Platform, Shenzhen, China, as a research intern. His research interests include computer vision and pattern recognition, with a focus on generative models.
\end{IEEEbiography}

\vspace{-1.3cm}
\begin{IEEEbiography}
[{\includegraphics[height=1.16in,clip,keepaspectratio]{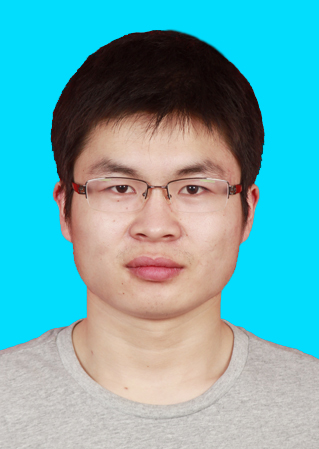}}]
{Li Shen} received his Ph.D. in school of mathematics, South China University of Technology in 2017. He is currently a research scientist at JD Explore Academy, China. Previously, he was a research scientist at Tencent AI Lab, China. His research interests include theory and algorithms for large scale convex/nonconvex/minimax optimization problems, and their applications in statistical machine learning, deep learning, reinforcement learning, and game theory. 
\end{IEEEbiography}

\vspace{-1.2cm}
\begin{IEEEbiography}
[{\includegraphics[height=1.16in,clip,keepaspectratio]{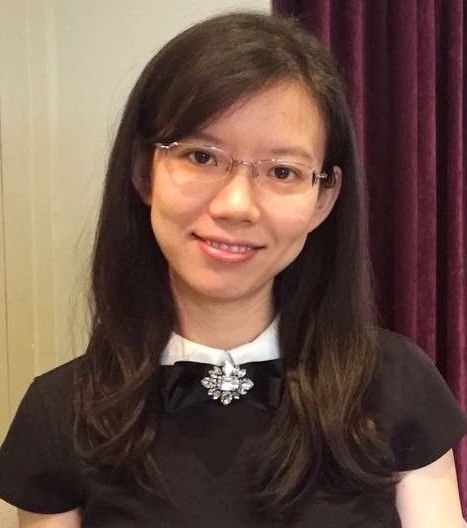}}]{Li Shen} is Senior Researcher at Tencent AI Lab. Her research interests include computer vision and deep learning, particularly in image recognition, and network architectures.
\end{IEEEbiography}

\vspace{-1.2cm}
\begin{IEEEbiography}
[{\includegraphics[height=1.16in,clip,keepaspectratio]{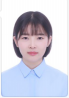}}]{Guinan Su} received her M.S. degree at University of Science and Technology of China. She is currently working in Microsoft Azure Machine learning team. Her research interests include natural language processing, Automatic machine learning, etc.
\end{IEEEbiography}

\vspace{-1.2cm}
\begin{IEEEbiography}
[{\includegraphics[height=1.06in,clip,keepaspectratio]{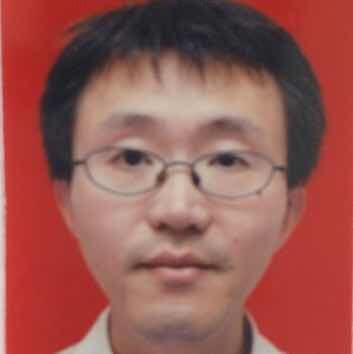}}]{Zhifeng Li (M'06-SM'11)} is currently a top-tier principal researcher with Tencent. He received the Ph. D. degree from the Chinese University of Hong Kong in 2006. After that, He was a postdoctoral fellow at the Chinese University of Hong Kong and Michigan State University for several years. Before joining Tencent, he was a full professor with the Shenzhen Institutes of Advanced Technology, Chinese Academy of Sciences. His research interests include deep learning, computer vision and pattern recognition, and face detection and recognition. He is currently serving on the Editorial Boards of Neurocomputing and IEEE Transactions on Circuits and Systems for Video Technology. He is a fellow of British Computer Society (FBCS).
\end{IEEEbiography}

\vspace{-1.2cm}
\begin{IEEEbiography}
[{\includegraphics[height=1.16in,clip,keepaspectratio]{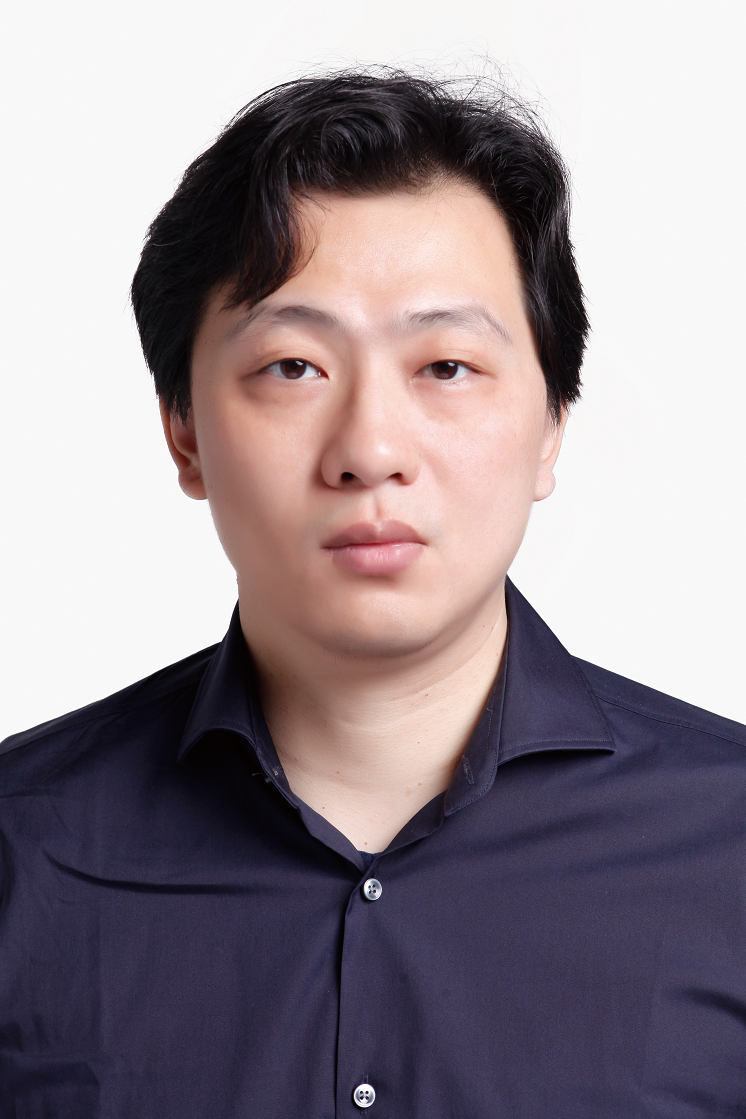}}]{Wei Liu (M'14-SM'19)} is currently a Distinguished Scientist of Tencent and the Director of Ads Multimedia AI Center of Tencent Data Platform, China. Prior to that, he has been a research staff member of IBM T. J. Watson Research Center, Yorktown Heights, NY, USA from 2012 to 2015. Dr. Liu has long been devoted to research and development in the fields of machine learning, computer vision, pattern recognition, information retrieval, big data, etc., and has published extensively in these fields with more than 250 peer-reviewed technical papers, and also held more than 50 granted US and European patents. Dr. Liu currently serves on the editorial boards of IEEE Transactions on Pattern Analysis and Machine Intelligence, IEEE Transactions on Neural Networks and Learning Systems, IEEE Transactions on Circuits and Systems for Video Technology, Pattern Recognition, etc. Dr. Liu was an area chair of well-known computer science and AI conferences, e.g., NeurIPS, IEEE ICCV, IEEE ICDM, IJCAI, ACM Multimedia, etc. He is also a Fellow of the International Association for Pattern Recognition (IAPR), a Fellow of the British Computer Society (BCS), and an Elected Member of the International Statistical Institute (ISI). 
\end{IEEEbiography}




\newpage
\section*{Appendix}

\section{Experiment Details}\label{experiment_details}
\subsection{Conventional GANs}
\subsubsection{Searching on CIFAR-10}
The CIFAR-10 dataset is comprised of $50000$ images for training. The resolution of the images is $32$x$32$. We randomly split the dataset into two sets during searching: one is used as the training set for optimizing network parameters $\omega_{G}$ and $\omega_{D}$ ($25000$ images), and another is used as the validation set for optimizing architecture parameters $\alpha_{G}$ ($25000$ images). The search epochs (i.e., $K$ in Algorithm 1) for alphaGAN$_{(l)}$ and alphaGAN$_{(s)}$ are set to $100$. The dimension of the noise vector is $128$. For a fair comparison, the discriminator adopted in searching is the same as the discriminator in AutoGAN \cite{gong2019autogan}. Batch sizes of both the generator and the discriminator are set to 64. The learning rates of weight parameters $\omega_{G}$ and $\omega_{D}$ are $2{\rm e}-4$ and the learning rate of architecture parameter $\alpha_{G}$ is $3{\rm e}-4$. We use Adam as the optimizer. The hyperparameters for optimizing weight parameters $\omega_{G}$ and $\omega_{D}$ are set as, $0.0$ for $\beta_1$ and $0.999$ for $\beta_2$, and $0$ for the weight decay. The hyperparameters for optimizing architecture parameters $\alpha_{G}$ are set as $0.5$ for $\beta_1$, $0.999$ for $\beta_2$ and $1{\rm e}-3$ for weight decay.

We use the entire training set of CIFAR-10 for retraining the network parameters after obtaining architectures. The dimension of the noise vector is $128$. Discriminator exploited in the re-training phase is identical to that during searching. The batch size of the generator is set to $128$. The batch size of the discriminator is set to $64$. The generator is trained for $100000$ iterations. The learning rates of the generator and discriminator are set to $2{\rm e}-4$.  The hyperparameters for the Adam optimizer are set to $0.0$ for $\beta_1$, $0.9$ for $\beta_2$ and $0$ for weight decay.

\subsubsection{Transferability}\label{STL-10_details}
The STL-10 dataset is comprised of $\sim105$k training images. We resize the images to the size of $48$x$48$ due to the consideration of memory and computational overhead. The dimension of the noise vector is $128$. We train the generator for $80000$ iterations. The batch sizes for optimizing the generators and the discriminator are set to $128$ and $64$, respectively. The channel numbers of the generator and the discriminator are set to $256$ and $128$, respectively. The learning rates for the generator and the discriminator are both set to $2{\rm e}-4$. We also use the Adam as the optimizer, where $\beta_1$ is set to $0.5$, $\beta_2$ is set to $0.9$ and weight decay is set to $0$.

\subsection{Search the architecture of StyleGAN2}\label{StyleGAN2_details}
As for CelebA, it contains $202599$ images, the resolution of which in searching and re-training is 128x128 with center cropping. As for LSUN-church, it contains $126227$ images, the resolution of which in searching and re-training is 256x256 with resizing. As for FFHQ, it contains totally $70000$ images, the resolution of which in searching and re-training is 1024x1024.

In searching, we adopt the searching configuration of alphaGAN$_{(s)}$, with $T=20$, $S=20$, and $R=5$. The datasets are equally split into training set and validation set. The number of searching iterations is $500$ for CelebA, LSUN-church, and FFHQ. The batch size of searching is $2$, and the searching is on a single Tesla V100 GPU. The channels of the generator and the discriminator is $1/8$ of the original channels in StyleGAN2, due to the limited memory.

In re-training, we adopt the config-f configuration in StyleGAN2 except for the loss of the generator, and we abandon the regularization term because the existing of the interpolation layers (``nearest\_conv" and ``bilinear\_conv"). For CelebA and LSUN-church, the mini-batch size is $8$ and the number of GPU is $4$. For FFHQ, the mini-batch size is $4$ and the number GPU is $8$.

\section{Additional Results for conventional GANs}
\subsection{The structure of G searched via alphaGAN}

\begin{figure}[ht!]
    \centering
    \includegraphics[scale=0.3]{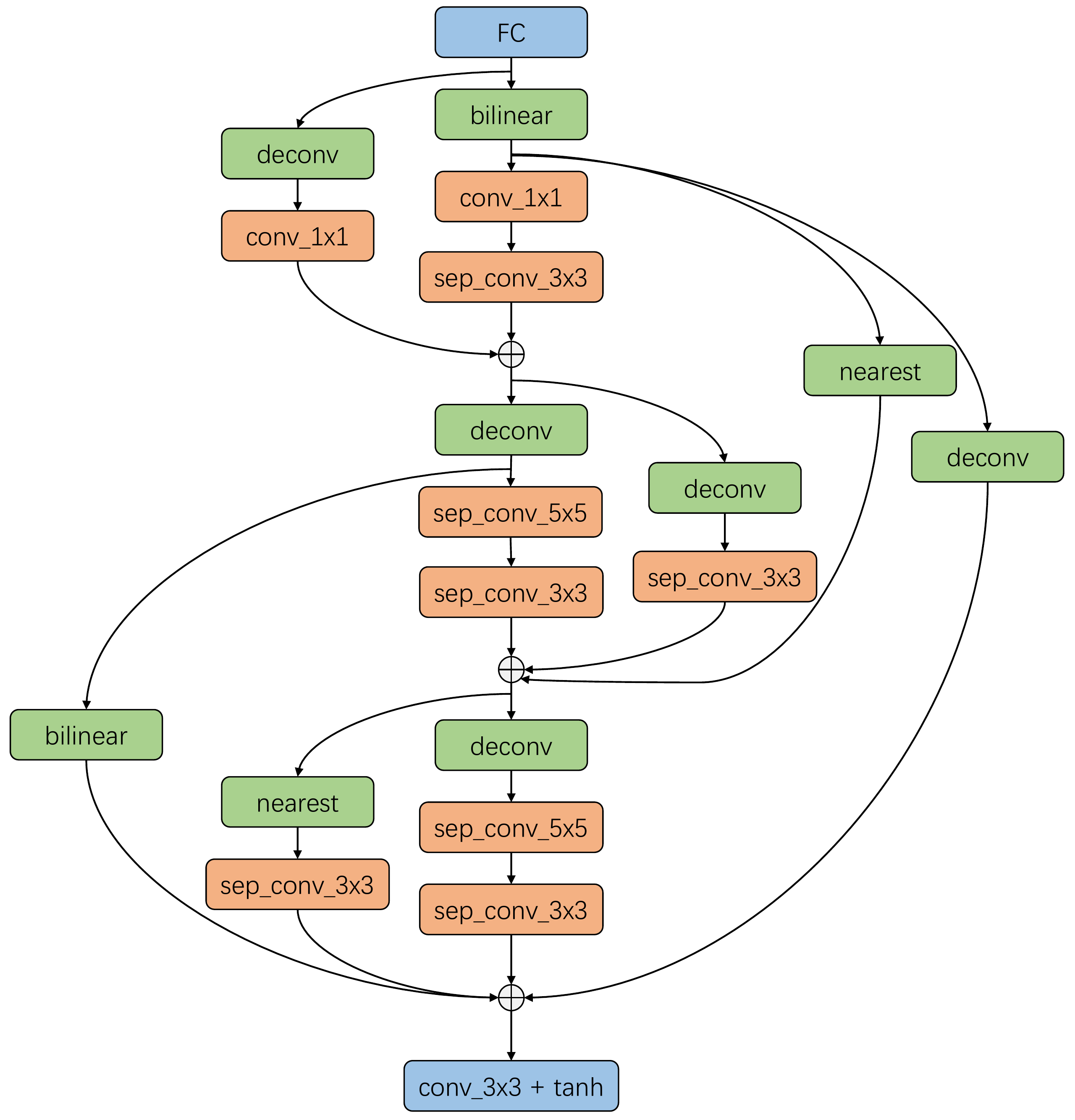}
    \caption{The structure of alphaGAN$_{(s)}$.}
    \label{structure_large_step}
\end{figure}

\begin{figure}[ht!]
    \centering
    \includegraphics[scale=0.3]{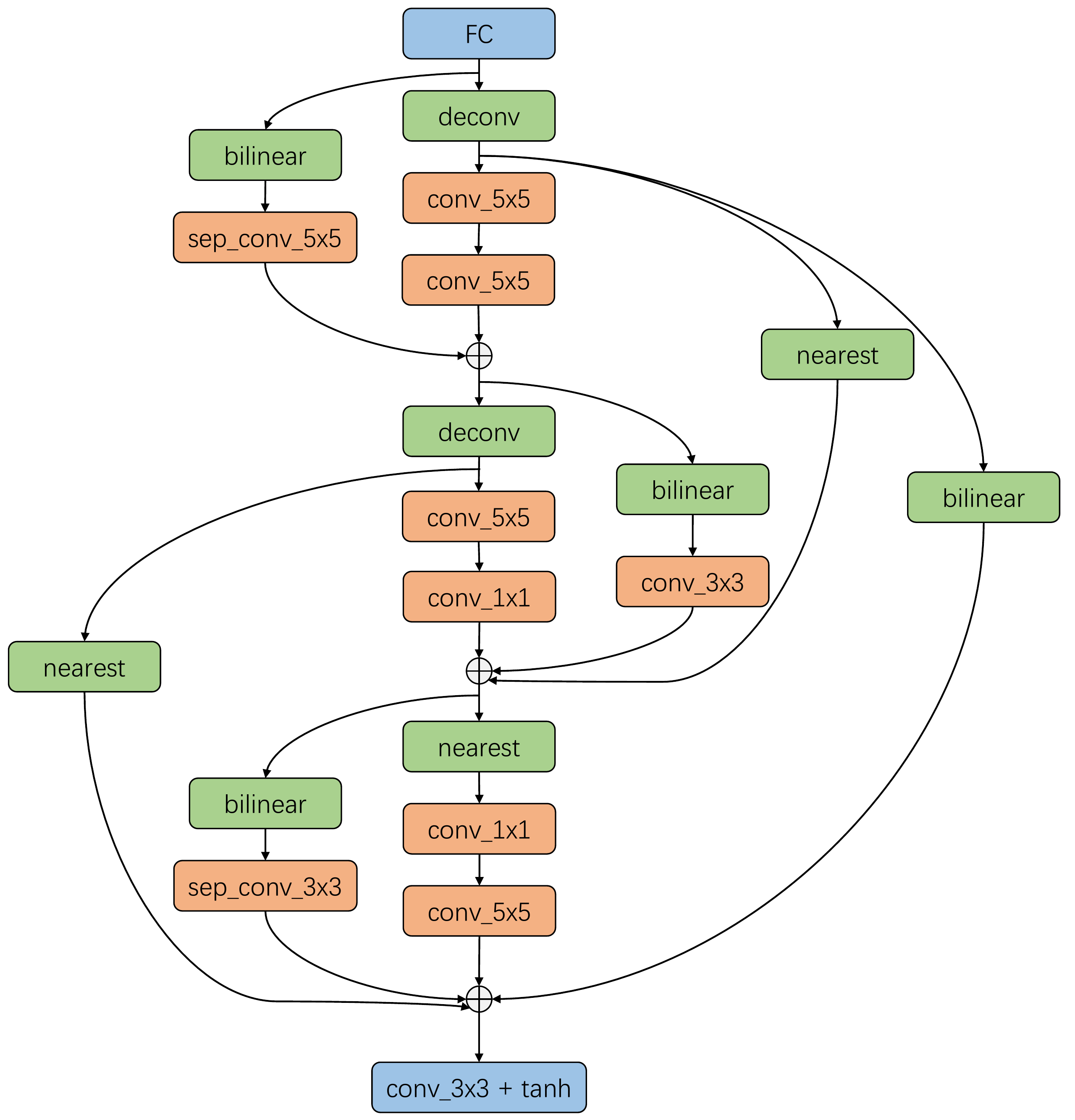}
    \caption{The structure of alphaGAN$_{(l)}$.}
    \label{structure_small_step}
\end{figure}

The structures of alphaGAN$_{(l)}$ and alphaGAN$_{(s)}$ are shown in Fig. \ref{structure_small_step} and Fig. \ref{structure_large_step}.

\subsection{Gumbel-max Trick}
Gumbel-max trick \cite{maddison2014sampling} can be written as,
\begin{align}
    \beta^{o'}=\frac{\exp\Big(\big(\alpha^{o'}+g^{o'}\big)/\tau\Big)}{\sum_{o\in \mathcal{O}_n}\exp\Big(\big(\alpha^{o}+g^{o}\big)/\tau\Big)},
\end{align}
where $\beta^{o'}$ is the probability of selecting operation $o'$ after Gumbel-max, and $\alpha^{o'}$ represents the architecture parameter of operation $o'$, respectively. $\mathcal{O}_{n}$ represents the operation search space. $g^{o}$ denotes samples drawn from the Gumbel (0,1) distribution, and $\tau$ represents the temperature to control the sharpness of the distribution. Instead of continuous relaxation, the trick chooses an operation on each edge, enabling discretization during searching.  
We compare the results by searching with and without Gumbel-max trick. The results in Tab. \ref{Gumbel-max trick and fix alphas} show that searching with Gumbel-max may not be the essential factor for obtaining high-performance generator architectures.

\subsection{Warm-up protocols}
The generator contains two parts of parameters, $(\omega_{G},\alpha_{G})$. The optimization of $\alpha_{G}$ is highly related to network parameters $\omega_{G}$. Intuitively, pretraining the network parameters $\omega_{G}$ can benefit the search of architectures since a better initialization may facilitate the convergence. To investigate the effect, we fix $\alpha_{G}$ and only update $\omega_{G}$ at the initial half of the searching schedule, and then $\alpha_{G}$ and $\omega_{G}$ are optimized alternately. This strategy is denoted as 'Warm-up' in Table \ref{Gumbel-max trick and fix alphas}.

\begin{table}[ht!]
  \caption{Gumbel-max trick and Warm-up. The ``baseline" denotes the structure searched under the default settings of alphaGAN.}
  \label{Gumbel-max trick and fix alphas}
  \centering
  \resizebox{0.5\textwidth}{!}{
  \begin{tabular}{llllll}
    \toprule
    Type & Name & Gumbel-max? & Fix alphas? & IS & FID \\
    \midrule
    \multirow{3}*{alphaGAN$_{(l)}$} & baseline & $\times$ & $\times$ & $8.51\pm0.06$ & $11.38$\\
     & Gumbel-max & $\checkmark$ & $\times$ & $8.48\pm0.10$ & $20.69$\\
     & Warm-up & $\times$ & $\checkmark$ & $8.34\pm0.07$ & $15.49$\\
    \midrule
    \multirow{3}*{alphaGAN$_{(s)}$} & baseline & $\times$ & $\times$ & $8.72\pm0.11$ & $12.86$\\
     & Gumbel-max & $\checkmark$ & $\times$ & $8.56\pm0.06$ & $15.66$\\
     & Warm-up & $\times$ & $\checkmark$ & $8.25\pm0.12$ & $19.07$\\
    \bottomrule
  \end{tabular}}
\end{table}

The results show that the strategy may not help performance, i.e., IS and FID of 'Warm-up' are slightly worse than those of the baseline, while it can benefit the searching efficiency, i.e., it spends $\sim15$ GPU-hours for alphaGAN$_{(l)}$ (compared to $\sim$22 GPU-hours via the baseline) , and $\sim1$ GPU-hour for alphaGAN$_{(s)}$ (compared to $\sim3$ GPU-hours via the baseline). 

\begin{table}[t]
  \caption{The channels in searching on the alphaGAN$_{(s)}$.}
  \label{The channels in search}
  \centering
  \vspace{-0.3cm}
  \resizebox{0.5\textwidth}{!}{
  \begin{tabular}{llllll}
    \toprule
    Search channels & Re-train channels & Params (M) & FLOPs (G) & IS & FID \\
    \midrule
    \multirow{2}*{G\_$32$ D\_$32$} & G\_$32$ D\_$32$ & $0.109$ & $0.02$ & $7.10\pm0.08$ & $36.22$\\
     & G\_$256$ D\_$128$ & $2.481$ & $1.12$ & $8.61\pm0.12$ & $14.98$\\
    \midrule
    \multirow{2}*{G\_$64$ D\_$64$} & G\_$64$ D\_$64$ & $0.403$ & $0.212$ & $7.97\pm0.09$ & $22.49$\\  
     & G\_$256$ D\_$128$ & $4.658$ & $3.26$ & $8.70\pm0.17$ & $14.02$\\
    \midrule
    \multirow{2}*{G\_$128$ D\_$128$} & G\_$128$ D\_$128$ & $1.967$ & $0.91$ & $8.26\pm0.08$ & $16.50$\\
     & G\_$256$ D\_$128$ & $7.309$ & $3.64$ & $8.75\pm0.09$ & $13.02$\\
    \midrule
    \multirow{4}*{G\_$256$ D\_$128$} & G\_$256$ D\_$128$ & $2.953$ & $1.32$ & $8.98\pm0.09$ & $10.35$\\
     & G\_$128$ D\_$128$ & $0.887$ & $0.34$ & $8.36\pm0.08$ & $17.12$\\
     & G\_$64$ D\_$64$ & $0.296$ & $0.09$ & $7.73\pm0.08$ & $24.81$\\
     & G\_$32$ D\_$32$ & $0.111$ & $0.025$ & $6.85\pm0.1$ & $35.6$\\
    \bottomrule
  \end{tabular}}
  \vspace{-0.3cm}
\end{table}

\subsection{The effect of Channels on Searching.}
As the default settings of alphaGAN, we search and re-train the networks with the same channel dimensions (i.e., G\_channels=256 and D\_channels=128), which are predefined. To explore the impact of the channel dimensions during searching on the final performance of the searched architectures, we adjust the channel numbers of the generator and the discriminator during searching based on the searching configuration of alphaGAN$_{(s)}$. The results are shown in Tab. \ref{The channels in search}. We observe that our method can achieve acceptable performance under a wide range of channel numbers (i.e., $32\sim256$). We also observe that using consistent channel dimensions during searching and re-training phases is beneficial to the final performance.

When reducing channels during searching, we observe an increasing trend on the operations of depth-wise convolutions with large kernels (e.g. 7x7), indicating that the operation selection induced by such an automated mechanism is adaptive to the need of preserving the entire information flow (i.e., increasing information extraction on the spatial dimensions to compensate for the channel limits).

\begin{figure}[ht!]
  \centering
  \subfigure[IS]{\includegraphics[width=0.49\textwidth]{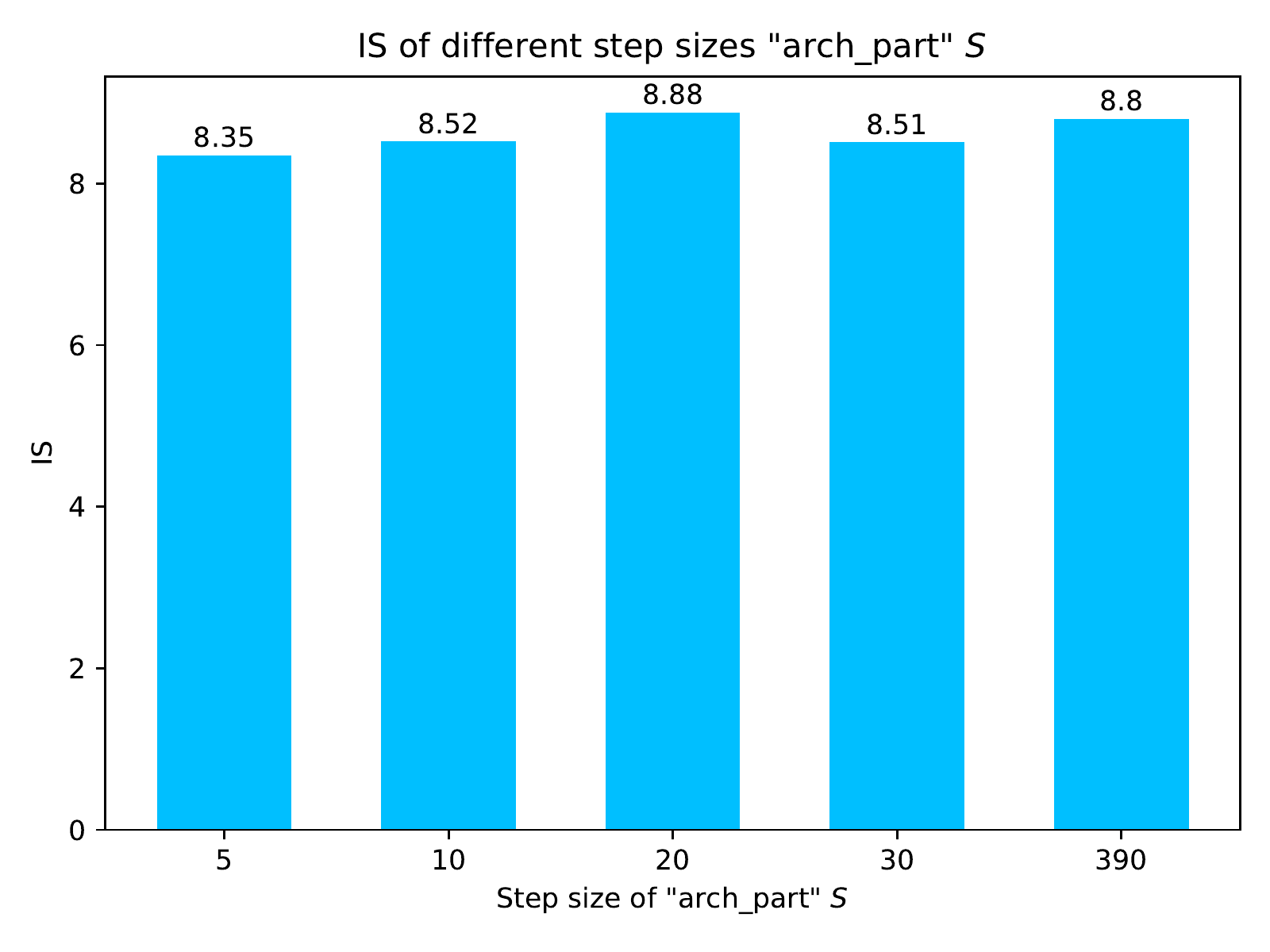}}\hspace{0.1cm}
  \subfigure[FID]{\includegraphics[width=0.49\textwidth]{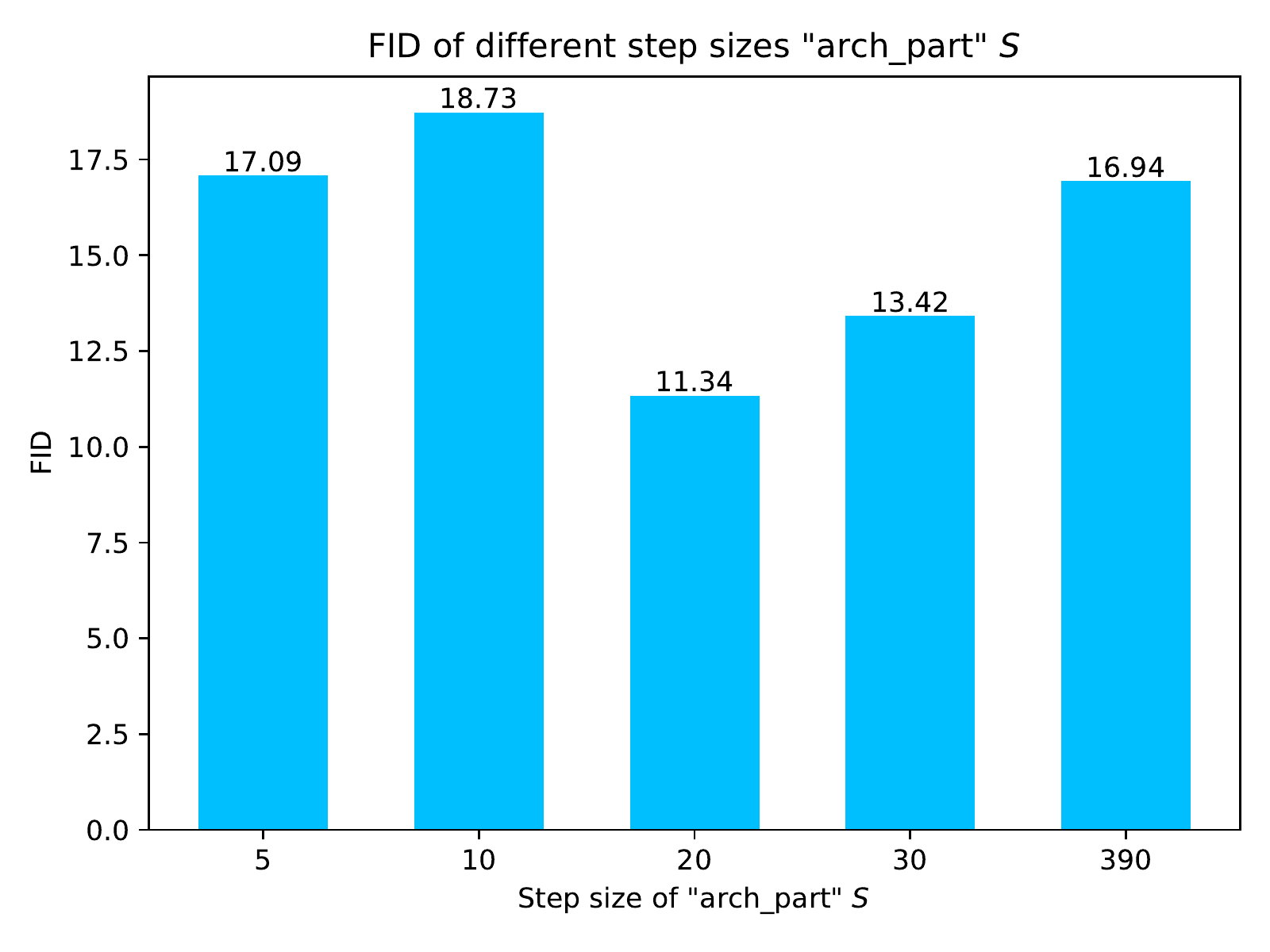}}
  \caption{The effect of different step sizes of 'arch part'.}
  \label{The effect of different step sizes of 'arch part'}
\end{figure}

\subsection{Effect of Step Sizes}
To analyze the effect of different step sizes on the ``arch part", corresponding to the optimization process of the architecture parameters $\alpha_{G}$ in Algorithm 1 (line 10-13). Since alphaGAN$_{(l)}$ has larger step sizes for 'weight part' and 'test-weight part' compared with alphaGAN$_{(s)}$, the step size of 'arch part' can be adjusted in a wider range. We select the alphaGAN$_{(l)}$ to conduct the experiments and the results are shown in Fig. \ref{The effect of different step sizes of 'arch part'}. We can observe that the method perform fair robustness among different step sizes on the IS metric, while network performance based on the FID metric may be hampered with a less proper step.

\begin{table}[t!]
  \caption{Repeated search on CIFAR-10.}
  \label{Repeat experiments of alphaGAN}
  \vspace{-0.3cm}
  \centering
  \resizebox{0.5\textwidth}{!}{
  \begin{tabular}{llllll}
    \toprule
    Name & Description & Params (M) & FLOPs (G) & IS & FID \\
    \midrule
    \multirow{3}*{alphaGAN$_{(s)}$} & \multirow{2}*{normal case} & $4.475$ & $2.36$ & $8.44\pm0.13$ & $13.62$\\
     &  & $2.953$ & $1.32$ & $8.72\pm0.11$ & $12.86$\\
     \cmidrule{2-6}
     & failure case & $2.994$ & $1.08$ & $6.77\pm0.07$ & $45.88$\\
    \midrule
    \multirow{3}*{alphaGAN$_{(l)}$} & \multirow{2}*{normal case} & $8.207$ & $2.41$ & $8.55\pm0.08$ & $15.42$\\
     &  & $8.618$ & $2.78$ & $8.51\pm0.06$ & $11.38$\\
     \cmidrule{2-6}
     & failure case & $4.666$ & $2.36$ & $7.48\pm0.1$ & $52.58$\\
    \bottomrule
  \end{tabular}}
\end{table}

\subsection{Failure Cases}\label{failure case study}
As we pointed out in the paper, the searching of alphaGAN will encounter failure cases, analogous to other NAS methods \cite{zela2019understanding}. To better understand the method, we present the comparison between normal cases and failure cases in Tab. \ref{Repeat experiments of alphaGAN} and the distributions of operations in the supplementary material. We find that deconvolution operations dominate in these failure cases. To validate this, we conduct the experiments on the variant by removing deconvolution operations from the search space under the configuration of alphaGAN$_{(s)}$. The results (with 6 runs) in Tab. \ref{search_wo_deconv} show that the failure cases can be prevented in this scenario.

\begin{table}[h]
  \vspace{-0.3cm}
  \caption{Search w$\backslash$o deconv on alphaGAN$_{(s)}$.}
  \label{search_wo_deconv}
  \vspace{-0.3cm}
  \centering
  \begin{tabular}{lllll}
    \toprule
    Name & Params (M) & FLOPs (G) & IS & FID \\
    \midrule
    Repeat\_1 & $4.594$ & $2.20$ & $8.29\pm0.08$ & $15.12$\\
    Repeat\_2 & $2.035$ & $0.51$ & $8.34\pm0.10$ & $14.92$\\
    Repeat\_3 & $1.586$ & $0.55$ & $8.24\pm0.09$ & $18.07$\\
    Repeat\_4 & $1.631$ & $0.58$ & $8.32\pm0.09$ & $15.85$\\
    Repeat\_5 & $1.631$ & $0.60$ & $8.43\pm0.08$ & $17.15$\\
    Repeat\_6 & $2.064$ & $1.03$ & $8.26\pm0.11$ & $16.00$\\
    \bottomrule
  \end{tabular}
\end{table}

We also test in another setting by integrating conv\_1x1 operation with the interpolation operations (i.e., nearest and bilinear) and making them learnable as deconvolution, denoted as ``learnable interpolation". The results (with 6 runs) under the configuration of alphaGAN$_{(s)}$ are shown in Tab. \ref{Alleviate failure case}, suggesting that the failure cases can also be alleviated by the strategy.

\begin{table}[h]
  \vspace{-0.3cm}
  \caption{The effect of 'learnable interpolation' on alphaGAN$_{(s)}$.}
  \label{Alleviate failure case}
  \centering
  \vspace{-0.3cm}
  \resizebox{0.5\textwidth}{!}{
  \begin{tabular}{llllll}
    \toprule
    Method & Name & Params (M) & FLOPs (G) & IS & FID \\
    \midrule
     \multirow{6}*{Learnable Interpolation} & Repeat\_1 & $2.775$ & $0.99$ & $8.43\pm0.15$ & $14.8$\\
      & Repeat\_2 & $2.243$ & $0.545$ & $8.49\pm0.12$ & $18.82$\\
      & Repeat\_3 & $3.500$ & $0.99$ & $8.35\pm0.1$ & $18.93$\\
      & Repeat\_4 & $3.195$ & $1.53$ & $8.59\pm0.1$ & $13.22$\\
      & Repeat\_5 & $2.968$ & $0.82$ & $8.22\pm0.11$ & $14.76$\\
      & Repeat\_6 & $2.712$ & $0.77$ & $8.41\pm0.11$ & $13.47$\\
    \bottomrule
  \end{tabular}}
\end{table}

\begin{table*}
  \caption{Search on STL-10. We search alphaGAN$_{(s)}$ on STL-10 and re-train the searched structure on STL-10 and CIFAR-10. In our repeated experiments, failure cases are prevented.}
  \label{search_on_STL-10}
  \centering
  \begin{tabular}{lllllll}
    \toprule
    Name & \tabincell{c}{Search time\\ (GPU-hours)} & \tabincell{c}{Dataset of\\ re-training} & Params (M) & FLOPs (G) & IS & FID \\
    \midrule
    Repeat\_1 & \tabincell{c}{$\sim2$} & \multirow{3}*{STL-10} & $4.552$ & $5.55$ & $9.22\pm0.08$ & $25.42$ \\
    Repeat\_2 & \tabincell{c}{$\sim2$} &  & $2.475$ & $2.01$ & $9.66\pm0.10$ & $29.28$ \\
    Repeat\_3 & \tabincell{c}{$\sim2$} &  & $4.013$ & $3.67$ & $9.47\pm0.10$ & $26.61$ \\
    \midrule
    Repeat\_1 & \tabincell{c}{$\sim2$} & \multirow{3}*{CIFAR-10} & $3.891$ & $2.47$ & $8.29\pm0.17$ & $13.94$ \\
    Repeat\_2 & \tabincell{c}{$\sim2$} &  & $1.815$ & $0.90$ & $8.20\pm0.13$ & $16.54$ \\
    Repeat\_3 & \tabincell{c}{$\sim2$} &  & $3.352$ & $1.63$ & $8.62\pm0.11$ & $12.64$ \\
    \bottomrule
  \end{tabular}
\end{table*}

\subsection{Single-level or bi-level?}
The formulation of alphaGAN is a bi-level optimization problem, shown as,

\begin{align}\label{origin formulation of alphaGAN}
    & \min_{\alpha_{G}}\quad \mathcal{V}(G(;\omega_{G}^*,\alpha_{G}),D(;\omega_{D}^*))\\
    & s.t.\quad (\omega_{G}^*,\omega_{D}^*):=\arg\min_{\omega_{G}}\max_{\omega_{D}} {\rm Adv}(G(;\omega_{G},\alpha_{G}),D(;\omega_{D}))\label{minmax constraint}
\end{align}

A natural problem arise from the above formulation. Can alphaGAN be formulated as a single-level optimization problem? Inspired from \cite{he2020milenas}, we reformulate the objective via Lagrangian multiplier method, as an additional experiment. The original objective in Eqn. \ref{origin formulation of alphaGAN} is then reformulated as,

\begin{align}\label{Single-level}
    & \min_{\omega_{G}, \alpha_{G}} \max_{\omega_{D}} \mathcal{V}(G(;\omega_{G},\alpha_{G}),D(;\omega_{D})) +\lambda * {\rm Adv}(G(;\omega_{G},\alpha_{G}),D(;\omega_{D}))
\end{align},

Using the alternative to search, we get the results in Tab. \ref{Single-level bi-level}. We find that the architecture obtained via single-level optimization is inferior to that of bi-level, regarding IS and FID. In single-level optimization, we exploit the combination of adversarial loss ${\rm Adv}(G,D)$ and duality gap $\mathcal{V}(G,D)$ to optimize both the weight parameters $\omega_{G}$ $\omega_{D}$ and the architecture parameters $\alpha_{G}$, whereas in bi-level optimization, we exploit ${\rm Adv}(G,D)$ to optimize the weight parameters $\omega_{G}$ $\omega_{D}$ and $\mathcal{V}(G,D)$ to optimize the architecture parameters $\alpha_{G}$ respectively. We conjecture that the inferior result of single-level optimization is due to the mixture of adversarial loss ${\rm Adv}(G,D)$ and duality gap $\mathcal{V}(G,D)$ in optimizing the weight parameters $\omega_{G}$ $\omega_{D}$ and the architecture parameters $\alpha_{G}$.

To be noted, adding the minimax constraint (i.e., Eqn. \ref{minmax constraint}) as a penalty term into the objective is an open problem, which needs to be further explored. The formulation of Eqn. \ref{Single-level} we propose is heuristic and intuitive. Further exploration on single-level optimization of NAS-GAN methods is left as future work.

\begin{table}[t]
  \caption{The ablation study of single-level optimization and bi-level optimization in alphaGAN. We extend alphaGAN$_{(s)}$ into single-level optimization following Eqn. \ref{Single-level}. Single-level alphaGAN$_{(s)}$ is searched and re-trained on CIFAR-10.}
  \label{Single-level bi-level}
  \centering
  \begin{tabular}{lll}
    \toprule
    Name & IS & FID \\
    \midrule
    baseline (bi-level) & $8.98\pm0.09$ & $10.35$ \\
    single-level & $8.40\pm0.07$ & $23.40$ \\ 
    \bottomrule
  \end{tabular}
\end{table}

\begin{table}[ht!]
  \caption{The results of applying alphaGAN to BigGAN on CIFAR-10.}
  \label{BigGAN on CIFAR-10}
  \centering
  \begin{tabular}{llll}
    \toprule
    Name & IS & FID \\
    \midrule
    BigGAN \cite{brock2018large} & $9.22$ & $14.73$ \\
    BigGAN + alphaGAN & $9.56\pm0.11$ & $10.27$ \\
    \bottomrule
  \end{tabular}
\end{table}

\subsection{Searching on STL-10}
We also search alphaGAN$_{(s)}$ on STL-10. The channel dimensions in the generator and the discriminator are set to 64 (due to the consideration of GPU memory limit). We use the size of 48x48 as the resolution of images. The rest experimental settings are same as the one of searching on CIFAR-10. The settings remain the same as Section \ref{STL-10_details} when retraining the networks. 

The results of three runs are shown in Tab. \ref{search_on_STL-10}. Our method achieves high performance on both STL-10 and CIFAR-10, demonstrating the effectiveness and transferability of alphaGAN are not confined to a certain dataset. alphaGAN$_{(s)}$ remains efficient which can obtain the structure reaching the state-of-the-art on STL-10 with only $2$ GPU-hours. We also find no failure case exists in the three repeated experiments of alphaGAN$_{(s)}$ compared to that on CIFAR-10, which may be related to multiple latent factors that datasets intrinsically possess (e.g., resolution, categories) and we leave as a future work.

\subsection{Relation between performance and structure}
\begin{figure*}[t]
    \centering
    \includegraphics[height=0.95\textwidth,angle=90]{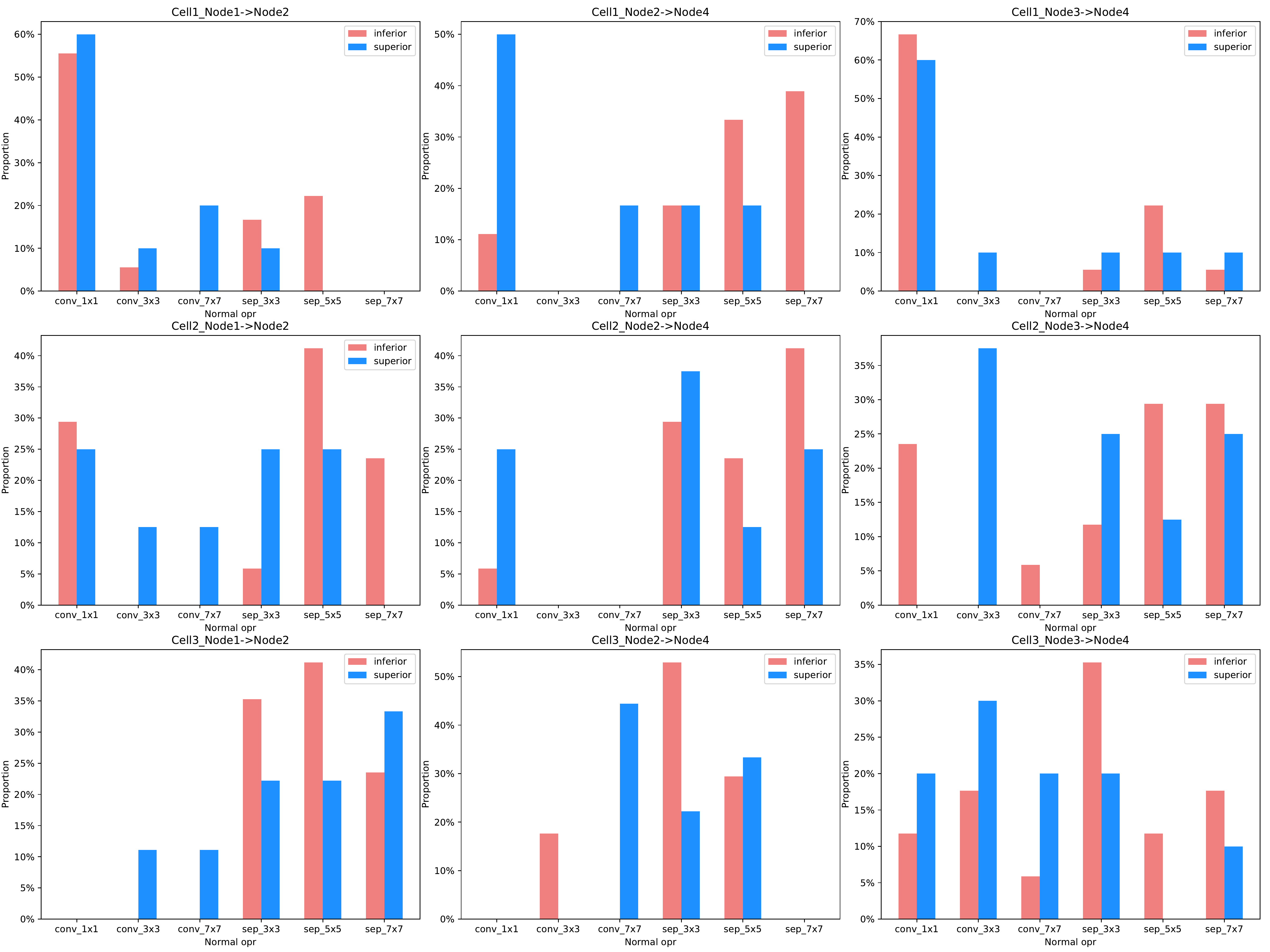}
    \caption{The distributions of normal operations.}
    \label{distribution_normal_operations}
\end{figure*}

\begin{figure*}[t]
    \centering
    \includegraphics[width=0.95\textwidth]{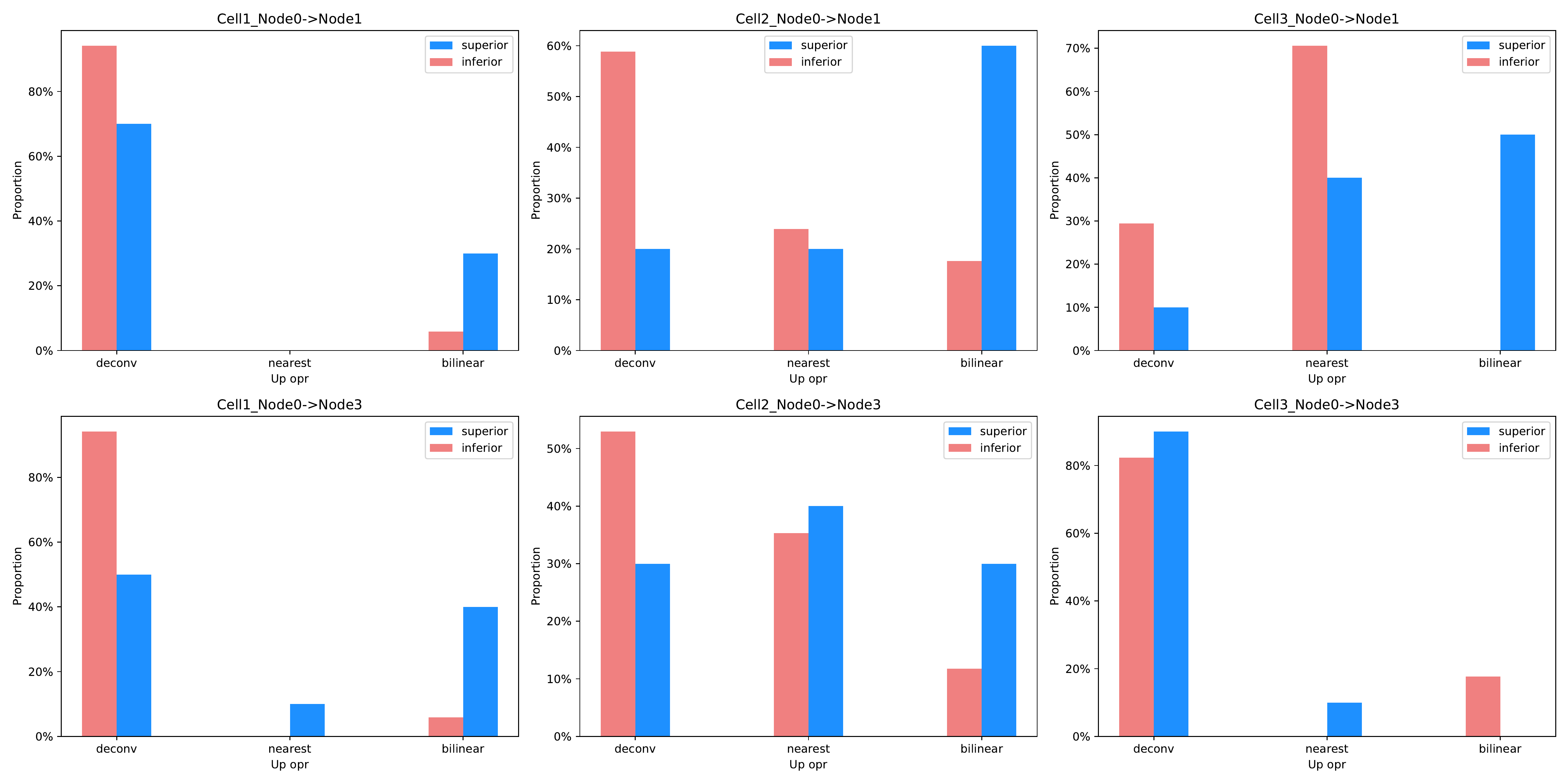}
    \caption{The distributions of up-sampling operations.}
    \label{distribution_up_operations}
\end{figure*}

We investigate the relation between architectures and performances by analyzing the operation distribution of searched architectures, as shown in Fig. \ref{distribution_normal_operations} and Fig. \ref{distribution_up_operations}. For simplicity , we divide the structures into two degrees, 'superior' (achieving IS > 8.0, FID < 15.0) and 'inferior' (achieving IS < 8.0, FID > 15.0). By the comparison between superior and inferior architectures, we have the following observations. First, for up-sampling operations, superior architectures tend to exploit ``nearest" or ``bilinear" rather than ``deconvolution" operations. Second, ``conv\_1x1" operations dominate in the cell\_1 of superior generators, suggesting that convolutions with large kernel sizes may not be optimal when the spatial dimensions of feature maps are relatively small (i.e., 8x8). Finally, convolutions with large kernels (e.g., conv\_5x5, sep\_conv\_3x3, and sep\_conv\_5x5) are preferred on higher resolutions (i.e., cell\_3 of superior generators), indicating the benefit of integrating information from relatively large receptive fields for low-level representations on high resolutions.

\begin{figure*}[t]
  \label{The illustration of StyleGAN2}
  \begin{center}
  \includegraphics[width=\textwidth]{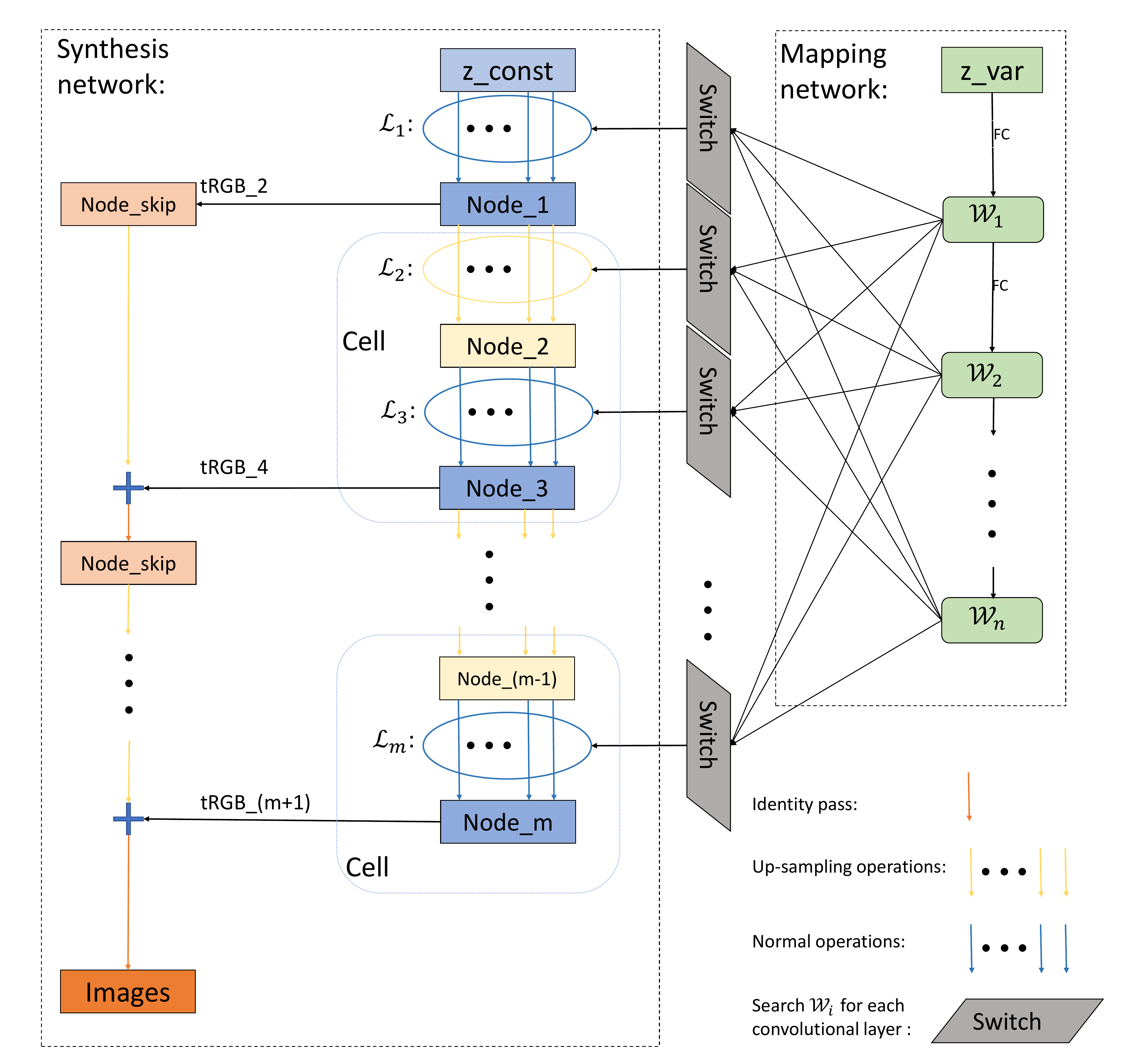}
  \vspace{-0.6cm}
  \caption{Searching the operations and the intermediate latent $\left\{\mathcal{W}_{i}\right\}_{i=1,...,n}$ of StyleGAN2. ``tRGB\_2", ``tRGB\_4", and ``tRGB\_(m+1)" denote the skip connection layers in the original StyleGAN2 \cite{karras2020analyzing}, which are conv\_1x1 operations. The number (e.g., 2) after the underscore in ``tRGB\_2" denotes that ``tRGB\_2" receives the latent $\mathcal{W}_{i}$ identical to the latent received by the convolutional layer $\mathcal{L}_{2}$.}
  \vspace{-0.5cm}
  \end{center}
\end{figure*}

\begin{table*}[t!]
  \caption{The architectures obtained via AdversarialNAS with the ``macro" paradigm, respectively.}
  \label{Searched_arch_macro}
  \vspace{-0.4cm}
  \centering
  \resizebox{\textwidth}{!}{
  \begin{tabular}{c|c|c|ccccccccc}
    \toprule
    \multirow{2}*{Dataset} & \multirow{2}{*}{Network} & \multirow{2}*{Searched operation/$\mathcal{W}_{i}$} & \multicolumn{9}{c}{Input resolution} \\
    \cline{4-12}
     &  &  & 4x4 & 8x8 & 16x16 & 32x32 & 64x64 & 128x128 & 256x256 & 512x512 & 1024x1024 \\
    \midrule
    \midrule
    \multirow{4}*{CelebA} & \multirow{2}{*}{Generator} & Normal operation & conv\_1x1 & conv\_1x1 & conv\_1x1 & conv\_1x1 & conv\_1x1 & conv\_1x1 & - & - & - \\ 
     &  & Up-sampling operation & nearest\_conv & nearest\_conv & nearest\_conv & nearest\_conv & bilinear\_conv & - & - & - & - \\
    \cline{2-12}
     & \multirow{2}*{Discriminator} & Normal operation & - & conv\_5x5 & conv\_1x1 & conv\_3x3 & conv\_3x3 & conv\_1x1 & - & - & - \\
     &  & Down-sampling operation & - & conv\_1x1 & conv\_1x1 & conv\_3x3 & conv\_3x3 & conv\_3x3 & - & - & - \\
    \hline
    \multirow{4}*{church} & \multirow{2}*{Generator} & Normal operation & conv\_1x1 & conv\_3x3 & conv\_1x1 & conv\_3x3 & conv\_1x1 & conv\_1x1 & conv\_1x1 & - & - \\ 
     &  & Up-sampling operation & bilinear\_conv & nearest\_conv & nearest\_conv & nearest\_conv & bilinear\_conv & bilinear\_conv & - & - & - \\
    \cline{2-12}
     & \multirow{2}*{Discriminator} & Normal operation & - & conv\_1x1 & conv\_5x5 & conv\_1x1 & conv\_5x5 & conv\_1x1 & conv\_1x1 & - & - \\
     &  & Down-sampling operation & - & conv\_5x5 & conv\_3x3 & conv\_5x5 & conv\_1x1 & conv\_5x5 & conv\_3x3 & - & - \\
    \hline
    \multirow{4}*{FFHQ} & \multirow{2}*{Generator} & Normal operation & conv\_1x1 & conv\_3x3 & conv\_1x1 & conv\_1x1 & conv\_1x1 & conv\_1x1 & conv\_1x1 & conv\_3x3 & conv\_3x3 \\  
     &  & Up-sampling operation & nearest\_conv & nearest\_conv & nearest\_conv & deconv & nearest\_conv & nearest\_conv & deconv & nearest\_conv & - \\
    \cline{2-12}
    & \multirow{2}*{Discriminator} & Normal operation & - & conv\_1x1 & conv\_3x3 & conv\_5x5 & conv\_1x1 & conv\_1x1 & conv\_5x5 & conv\_1x1 & conv\_1x1 \\
     &  & Down-sampling operation & - & conv\_5x5 & conv\_5x5 & conv\_1x1 & conv\_3x3 & conv\_3x3 & conv\_1x1 & conv\_1x1 & conv\_3x3 \\
    \bottomrule
  \end{tabular}}
  \vspace{-0.3cm}
\end{table*}

\begin{table*}[t!]
  \caption{The architectures obtained via AdversarialNAS with the ``micro + dynamic $\left\{\mathcal{W}_{i}\right\}$" paradigm and the ``macro" paradigm. $*$ denotes that the latent $\mathcal{W}_{i}$ searched for the layer ``tRGB\_(m+1)" in Fig. \ref{The illustration of StyleGAN2}}
  \label{Searched_arch_microStyles}
  \vspace{-0.4cm}
  \centering
  \resizebox{\textwidth}{!}{
  \begin{tabular}{c|c|c|ccccccccc}
    \toprule
    \multirow{2}*{Dataset} & \multirow{2}{*}{Network} & \multirow{2}*{Searched operation/$\mathcal{W}_{i}$} & \multicolumn{9}{c}{Input resolution} \\
    \cline{4-12}
     &  &  & 4x4 & 8x8 & 16x16 & 32x32 & 64x64 & 128x128 & 256x256 & 512x512 & 1024x1024 \\
    \midrule
    \midrule
    \multirow{6}*{CelebA} & \multirow{4}{*}{Generator} & $\mathcal{W}_i$ for the normal operation & $\mathcal{W}_{7}$ & $\mathcal{W}_{7}$ & $\mathcal{W}_{5}$ & $\mathcal{W}_{6}$ & $\mathcal{W}_{7}$ & $\mathcal{W}_{5}$, ${\mathcal{W}_{3}}^{*}$ & - & - & - \\ 
     &  & Normal operation & \multicolumn{6}{c}{conv\_1x1} & - & - & - \\
     &  & $\mathcal{W}_i$ for the up-sampling operation & $\mathcal{W}_{8}$ & $\mathcal{W}_{3}$ & $\mathcal{W}_{5}$ & $\mathcal{W}_{6}$ & $\mathcal{W}_{8}$ & - & - & - & - \\
     &  & Up-sampling operation & \multicolumn{5}{c}{nearest\_conv} & - & - & - & - \\
    \cline{2-12}
     & \multirow{2}*{Discriminator} & Normal operation & - & conv\_5x5 & conv\_5x5 & conv\_1x1 & conv\_3x3 & conv\_3x3 & - & - & - \\
     &  & Down-sampling operation & - & conv\_5x5 & conv\_3x3 & conv\_5x5 & conv\_3x3 & conv\_5x5 & - & - & - \\
    \hline
    \multirow{6}*{church} & \multirow{4}*{Generator} & $\mathcal{W}_i$ for the normal operation & $\mathcal{W}_{2}$ & $\mathcal{W}_{4}$ & $\mathcal{W}_{7}$ & $\mathcal{W}_{2}$ & $\mathcal{W}_{1}$ & $\mathcal{W}_{3}$ & $\mathcal{W}_{4}$, ${\mathcal{W}_{2}}^*$ & - & - \\ 
     &  & Normal operation & \multicolumn{7}{c}{conv\_1x1} & - & - \\
     &  & $\mathcal{W}_i$ for the up-sampling operation & $\mathcal{W}_{8}$ & $\mathcal{W}_{7}$ & $\mathcal{W}_{5}$ & $\mathcal{W}_{8}$ & $\mathcal{W}_{3}$ & $\mathcal{W}_{1}$ & - & - & - \\
     &  & Up-sampling operation & \multicolumn{6}{c}{nearest\_conv} & - & - & - \\
    \cline{2-12}
     & \multirow{2}*{Discriminator} & Normal operation & - & conv\_1x1 & conv\_5x5 & conv\_3x3 & conv\_3x3 & conv\_1x1 & conv\_3x3 & - & - \\
     &  & Down-sampling operation & - & conv\_1x1 & conv\_1x1 & conv\_5x5 & conv\_3x3 & conv\_3x3 & conv\_3x3 & - & - \\
    \hline
    \multirow{6}*{FFHQ} & \multirow{4}*{Generator} & $\mathcal{W}_i$ for the normal operation & $\mathcal{W}_{6}$ & $\mathcal{W}_{6}$ & $\mathcal{W}_{7}$ & $\mathcal{W}_{3}$ & $\mathcal{W}_{3}$ & $\mathcal{W}_{7}$ & $\mathcal{W}_{5}$ & $\mathcal{W}_{5}$ & $\mathcal{W}_{4}$, ${\mathcal{W}_{1}}^{*}$ \\  
     &  & Normal operation & \multicolumn{9}{c}{conv\_5x5} \\
     &  & $\mathcal{W}_i$ for the up-sampling operation & $\mathcal{W}_{3}$ & $\mathcal{W}_{7}$ & $\mathcal{W}_{4}$ & $\mathcal{W}_{7}$ & $\mathcal{W}_{2}$ & $\mathcal{W}_{1}$ & $\mathcal{W}_{2}$ & $\mathcal{W}_{5}$ & - \\
     &  & Up-sampling operations & \multicolumn{8}{c}{deconv} & - \\
    \cline{2-12}
    & \multirow{2}*{Discriminator} & Normal operation & - & conv\_1x1 & conv\_3x3 & conv\_3x3 & conv\_3x3 & conv\_1x1 & conv\_5x5 & conv\_3x3 & conv\_1x1 \\
     &  & Down-sampling operation & - & conv\_1x1 & conv\_5x5 & conv\_5x5 & conv\_3x3 & conv\_1x1 & conv\_5x5 & conv\_3x3 & conv\_1x1 \\
    \bottomrule
  \end{tabular}}
  \vspace{-0.3cm}
\end{table*}

\begin{table*}[t]
  \caption{The results of exploiting alphaGAN to search the architecture of StyleGAN2, with separable depth-wise convolutions in the operation pool. $\dagger$ denotes the reproduced results based on the official released code.}
  \label{Searching StyleGAN2 with depth-wise}
  \vspace{-0.3cm}
  \centering
  \begin{tabular}{c|ccc|ccc|c}
    \toprule
    \multirow{2}*{Paradigm} & \multicolumn{3}{c|}{Search} & \multicolumn{3}{c|}{Re-train} & \multirow{2}*{FID} \\
    \cline{2-7}
    & Dataset & \tabincell{c}{Cost\\ (GPU-hours)} & Search space & Dataset & kimg & \tabincell{c}{Model size\\ (M)} \\
    \midrule
    \midrule
    StyleGAN2(\cite{karras2020analyzing}) & - & - & - & FFHQ & $25000$ & $30.37$ & $2.84$ \\
    \hline
    StyleGAN2$\dagger$ & - & - & - & FFHQ & $20000$ & $30.37$ & $3.50$ \\
    \midrule
    \midrule
    \multirow{2}*{alphaGAN$_{(s)}$ + macro} & FFHQ & $6$ & $\sim1.3\times10^9$ & FFHQ & $20000$ & $31.68$ & $2.95$ \\
     & \tabincell{c}{FFHQ\\ (separable depth-wise convolutions)} & $6$ & $\sim6.6\times10^{10}$ & FFHQ & $20000$ & $21.42$ & $3.88$ \\
    \bottomrule
  \end{tabular}
\end{table*}

\begin{table*}[ht!]
  \caption{The architectures obtained via alphaGAN$_{(s)}$ with the ``macro" paradigm, searched with separable depth-wise convolutions.} 
  \label{Searched_arch_depth}
  \vspace{-0.5cm}
  \centering
  \resizebox{\textwidth}{!}{
  \begin{tabular}{c|c|c|ccccccccc}
    \toprule
    \multirow{2}*{Dataset} & \multirow{2}{*}{Network} & \multirow{2}*{Searched operation/$\mathcal{W}_{i}$} & \multicolumn{9}{c}{Input resolution} \\
    \cline{4-12}
     &  &  & 4x4 & 8x8 & 16x16 & 32x32 & 64x64 & 128x128 & 256x256 & 512x512 & 1024x1024 \\
    \midrule
    \midrule
    \multirow{2}*{FFHQ} & \multirow{2}*{Generator} & Normal operation & conv\_3x3 & sep\_conv\_7x7 & sep\_conv\_5x5 & sep\_conv\_3x3 & sep\_conv\_7x7 & conv\_1x1 & sep\_conv\_7x7 & conv\_3x3 & conv\_5x5 \\  
     &  & Up-sampling operation & deconv & bilinear\_conv & deconv & bilinear\_conv & bilinear\_conv & deconv & deconv & deconv & - \\
    \bottomrule
  \end{tabular}}
  \vspace{-0.3cm}
\end{table*}

\section{Additional results for StyleGAN2}
\subsection{The architectures obtained via AdversarialNAS}
The architectures obtained via AdversarialNAS under ``micro + dynamic $\left\{\mathcal{W}_{i}\right\}$" paradigm and ``macro" paradigm are shown in Tab. \ref{Searched_arch_microStyles} \ref{Searched_arch_macro}.

\begin{table}[t!]
  \caption{The ablation study of separable depth-wise convolutions on StyleGAN2.}
  \label{Depth-wise on StyleGAN2}
  \vspace{-0.4cm}
  \centering
  \begin{tabular}{llll}
    \toprule
    Modulate & Dataset & kimg & FID \\
    \midrule
    Depth-wise & FFHQ & 15000 & $8.7738$ \\
    Point-wise & FFHQ & 15000 & $8.8708$ \\
    \bottomrule
  \end{tabular}
\end{table}

\subsection{Adding separable depth-wise convolutions into the operation pool}
We add separable depth-wise convolutions into the operation pool for searching StyleGAN2.

To show the impact of the separable depth-wise convolutions on StyleGAN2, we replace conv\_3x3 operation in the generator with sep\_conv\_3x3. We modulate depth-wise convolution or point-wise convolution, shown in Tab. \ref{Depth-wise on StyleGAN2}. We find that modulating on depth-wise convolution shows better performance. Thus, we conduct experiments on ``macro" paradigm with the normal operation pool \{conv\_1x1,\quad conv\_3x3,\quad conv\_5x5,\quad sep\_conv\_3x3,\quad sep\_conv\_5x5,\quad sep\_conv\_7x7\}, modulating on the depth-wise convolution, based on the results in Tab. \ref{Depth-wise on StyleGAN2}. We re-train the architecture and show the result in Tab. \ref{Searching StyleGAN2 with depth-wise}. The architecture of the generator is shown in Tab. \ref{Depth-wise on StyleGAN2}. The generator searched with separable depth-wise convolutions reaches FID of 3.88 with the parameter size 21.42M. Despite that the parameter size of the generator with separable depth-wise convolutions is smaller than styleGAN2 baseline (21.42M vs 30.37M), the performance of it is inferior to that of StyleGAN2 baseline (3.88 vs 2.84). Moreover, we find that modulating the depth-wise convolution causes the unstable training behavior during re-training. To improve the performance and stabilize the training process of StyleGAN2 with light-weight separable depth-wise convolution, is left as future work.

\section{Additional results for BigGAN}

\begin{figure*}[t]
  \centering
  \includegraphics[width=0.5\textwidth]{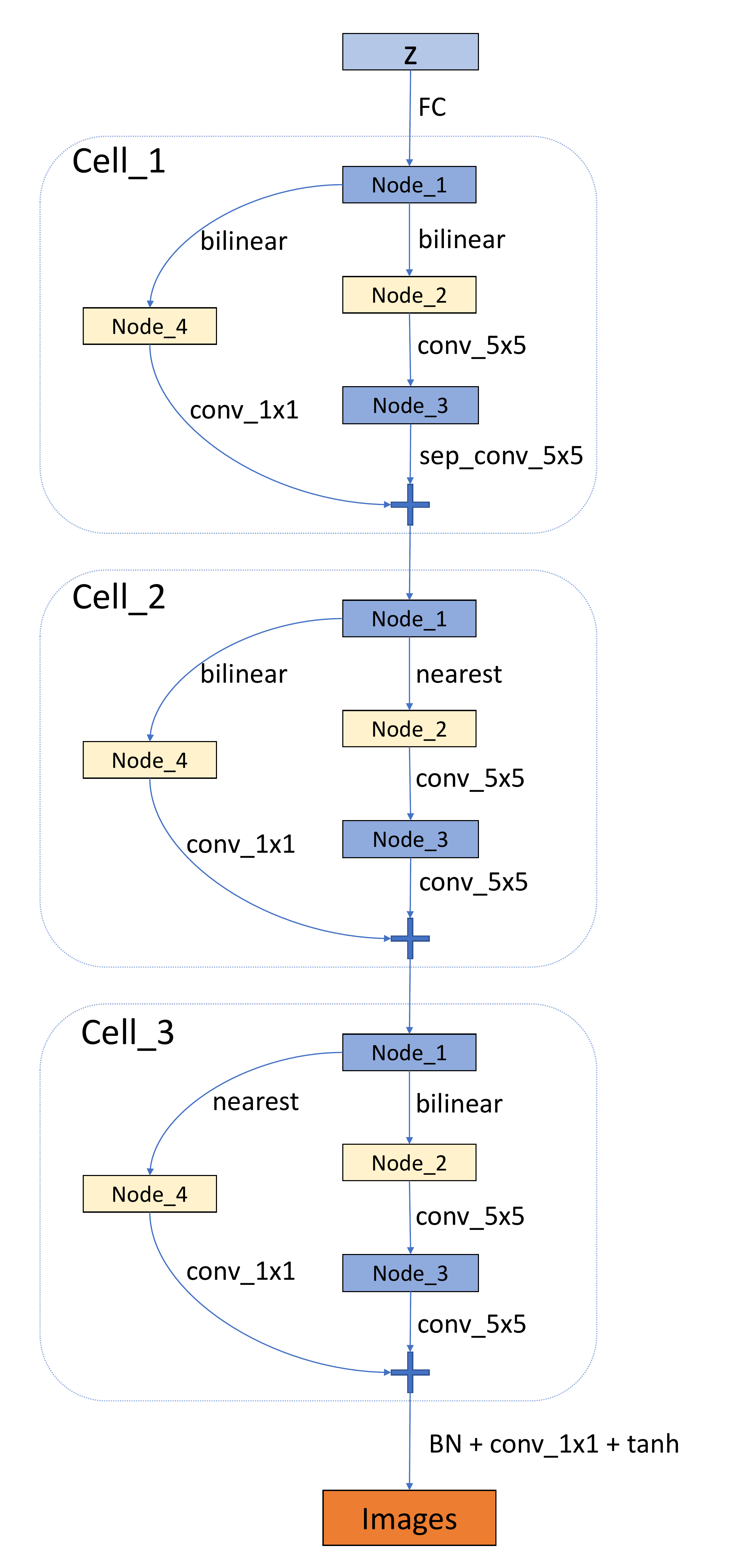}
  \vspace{-0.6cm}
  \caption{The searched architecture of BigGAN. For simplicity, we omit the hierarchical architecture of BigGAN and the label information fed into the class conditional batch normalization layer.}
  \label{Fig: Searched BigGAN}
  \vspace{-0.5cm}
\end{figure*}

As a general NAS-GAN framework, alphaGAN can be applied to BigGAN with minimax loss function. Due to limited time, we search and re-train BigGAN on CIFAR-10 under ``macro" setting with the normal operation pool \{conv\_1x1,\quad conv\_3x3,\quad conv\_5x5,\quad sep\_conv\_3x3,\quad sep\_conv\_5x5,\quad sep\_conv\_7x7\} and the up-sampling operation pool \{nearest,\quad bilinear\}. We remove the ``deconv" operation due to its negative impact on the performance of the generator, as elaborated in Section \ref{failure case study}. The complexity of searching is $2^{6}\times6^{6}=3.0\times10^{6}$. 

The results of searching BigGAN via alphaGAN are shown in Table \ref{BigGAN on CIFAR-10}. The searched architecture of BigGAN consistently outperforms BigGAN baseline on both IS (9.56 vs 9.22) and FID (10.27 vs 14.73), demonstrating the contribution of alphaGAN is orthogonal to BigGAN. The searched architecture of BigGAN is shown in Fig. \ref{Fig: Searched BigGAN}.

\end{document}